\documentclass{article}

\usepackage{PRIMEarxiv}

\usepackage[utf8]{inputenc} % allow utf-8 input
\usepackage[T1]{fontenc}    % use 8-bit T1 fonts
\usepackage{hyperref}       % hyperlinks
\usepackage{url}            % simple URL typesetting
\usepackage{booktabs}       % professional-quality tables
\usepackage{amsfonts}       % blackboard math symbols
\usepackage{nicefrac}       % compact symbols for 1/2, etc.
\usepackage{microtype}      % microtypography
\usepackage{lipsum}
\usepackage{fancyhdr}       % header
\usepackage{graphicx}       % graphics
\usepackage{xcolor}
\graphicspath{{media/}}     % organize your images and other figures
\usepackage{comment}

\usepackage{amsmath}

\usepackage{multirow}
\usepackage[normalem]{ulem}
\useunder{\uline}{\ul}{}

\title{Deep Learning Meets Process-Based Models: A Hybrid Approach to Agricultural Challenges
\thanks{\textit{\underline{Corresponding Author}}: 
\textbf{Liangxiu Han}} 
}

\author{
  Yue Shi,  Liangxiu Han*, Xin Zhang, Tam Sobeih\\
  Department of Computing, and Mathematics,\\
  Faculty of Science and Engineering,  \\
  Manchester Metropolitan University, \\
  Manchester, M1 5GD, UK. \\
  \texttt{\{y.shi, l.han\}@mmu.ac.uk} \\
   \And
  Amit Kumar Srivastava, Krishnagopal Halder, Frank Ewert \\
  Leibniz Centre for Agricultural Landscape Research (ZALF)\\
  Eberswalder Str \\
  Müncheberg, Germany\\
  \texttt{email@email} \\
   \And
  Thomas Gaiser, Nguyen Huu Thuy, Dominik Behrend \\
  Institute of Crop Science and Resource Conservation (INRES)\\
  University of Bonn,\\
  Bonn, Germany.\\
  \texttt{tgaiser@uni-bonn.de} \\
}

\begin{document}
\maketitle

\begin{abstract} 

Process-based models (PBMs) and deep learning (DL) are two key approaches in agricultural modelling, each offering distinct advantages and limitations. PBMs provide mechanistic insights based on physical and biological principles, ensuring interpretability and scientific rigour. However, they often struggle with scalability, parameterisation, and adaptation to heterogeneous environments. In contrast, DL models excel at capturing complex, nonlinear patterns from large datasets but may suffer from limited interpretability, high computational demands, and overfitting in data-scarce scenarios.

This study presents a systematic review of PBMs, DL models, and hybrid PBM-DL frameworks, highlighting their applications in agricultural and environmental modelling. We classify hybrid PBM-DL approaches into DL-informed PBMs, where neural networks refine process-based models, and PBM-informed DL, where physical constraints guide deep learning predictions. Additionally, we conduct a case study on crop dry biomass prediction, comparing hybrid models against standalone PBMs and DL models under varying data quality, sample sizes, and spatial conditions. The results demonstrate that hybrid models consistently outperform traditional PBMs and DL models, offering greater robustness to noisy data and improved generalisation across unseen locations.

Finally, we discuss key challenges, including model interpretability, scalability, and data requirements, alongside actionable recommendations for advancing hybrid modelling in agriculture. By integrating domain knowledge with AI-driven approaches, this study contributes to the development of scalable, interpretable, and reproducible agricultural models that support data-driven decision-making for sustainable agriculture.

\end{abstract}

% keywords can be removed
\keywords{Hybrid PBM-DL Modeling \and Process-based Modeling \and Deep Learning \and Computational Sustainability \and agricultural Modeling}

\section{Introduction}

The world population was projected to reach 9 billion by 2020 \cite{gerland2023s}. Agricultural production must not only meet increasing food demands but also enhance resource use efficiency while minimising negative environmental impacts. These challenges are further exacerbated by the increasing frequency and severity of climate change effects. Therefore, adopting sustainable agricultural practices is both crucial and necessary to ensure long-term productivity and environmental resilience.

Process-based models (PBMs) have long been essential tools \cite{laniak2013integrated} for studying and analyzing the interactions between agricultural production outputs and environmental variables\cite{change2001climate}. Recent advancements in machine learning, deep learning, and artificial intelligence (AI) \cite{shaikh2022towards} have significantly enhanced research capabilities in agriculture. The integration of PBMs with these advanced technologies has opened new possibilities, enabling more accurate, scalable, and actionable analyses of agricultural and environmental dynamics. Such synergies are driving innovation toward more sustainable and resilient farming systems. \par
%Process-based models have been common tools in \cite{laniak2013integrated} studies and investigation on interaction of production output with agricultural and environmental variables \cite{change2001climate}.  

%Recent development of information and communications technology (machine learning/deep learning/artificial intelligence (AI)) \cite{shaikh2022towards} and combination of PBMs model and DL) has also brought great facilitation in the research activities. \par

PBMs encompass a wide range of disciplines with expanding boundaries and increasing societal relevance, particularly in agricultural analysis \cite{nakayama2022impact, macpherson2020linking, couedel2024long}. In agriculture, these models forecast crop productivity and evaluate environmental impacts \cite{galmarini2024assessing, srivastava2023dynamic}. They simulate dynamic system responses to temporal forcings influenced by static landscape attributes. Many PBMs share commonalities across various processes due to their focus on temporal dynamics, often represented by nonlinear equations, including ordinary differential equations (ODEs) and partial differential equations (PDEs). The understanding of vegetation and crop growth processes varies, with some being well understood and others relying on assumptions or empirical representations \cite{sillero2021want, zeng2022optical}. Originally, the spatial scale of process-based cropping system models PBMs used for agro-ecosystem analysis is either a single plant or a point in a field with homogenous soil and canopy conditions. In this case, PBMs use parameterizations to represent processes at much finer scales than the computational grid of climate \cite{geary2020guide, he2023predicting}. However, these representations and parameterizations involve significant uncertainty due to complexity of the model. Within the broader frame of agricultural evaluations, these models are synthesized to deliver critical climatic forecasts and inform decision-making processes for resource custodians and policymakers  \cite{masson2021climate, battiston2021physics}. \par

In addition to PBMs, the rapid growth of Artificial Intelligence (AI), particularly within the realm of deep learning (DL), presents novel prospects for extracting insights from large datasets and bridging the knowledge gaps inherent in process-based models \cite{razavi2021deep, bhusal2022application}. While numerous DL methodologies have been introduced, the intrinsic differentiable advantage of DL has not been fully acknowledged.\par

\begin{figure}[h]  % Placement specifier added  
    \centering  
    \includegraphics[width=5.5in]{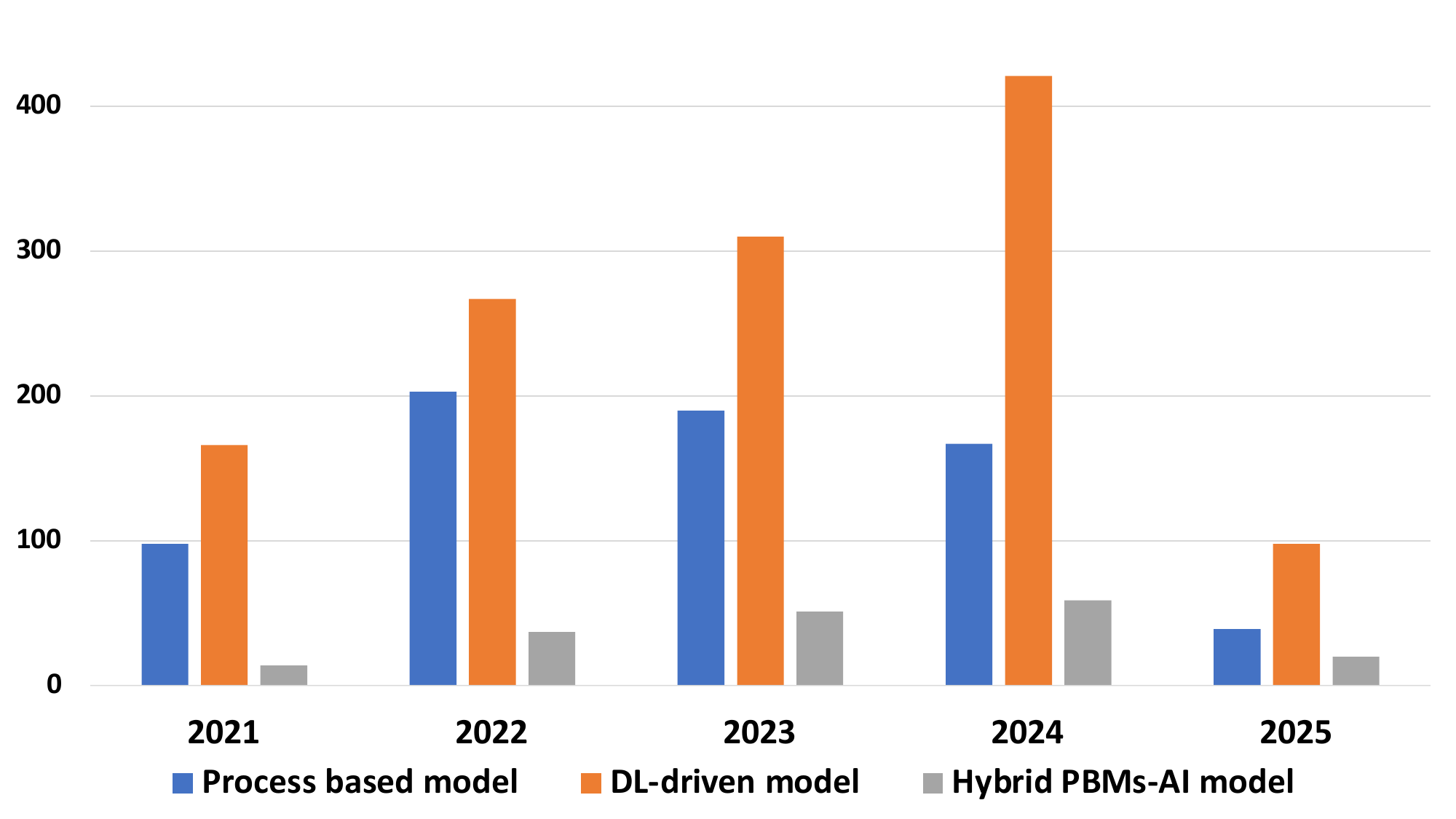}  
    \caption{Number of publications related to the agriculture modeling of hybrid PBM-DL models, comparing purely data-driven DL models and process-based models in the domain of agricultural analysis.}  
    \label{fig:4a1}  
\end{figure}

Recent development with synthesizing elements from PBMs and DL-driven models (so called hybrid PBM-DL models) has brought about a revolutionary advancement in agricultural science. Fig \ref{fig:4a1} illustrates the search results of PBM, DL hybrid PBM-DL models in agriculture modeling on Web of Science platform. It shows that, since 2021, there has been a rapid development in the application of hybrid PBM-DL models in agriculture application.  \par

In this study, we present a systematic overview of modelling approaches in agri-ecosystem research, focusing on process-based models (PBMs), deep learning (DL), and hybrid PBM-DL frameworks. We evaluate their respective strengths and limitations, particularly in the context of agricultural and environmental applications. The paper is structured as follows: Section \ref{sec:1} provides an overview of existing PBM and DL methodologies in agricultural modeling. Section \ref{sec:2} reviews hybrid PBM-DL models and their applications in agricultural systems. Section \ref{sec:4} details an experimental evaluation of the hybrid PBM-DL model's efficiency in agricultural science. Finally, Section \ref{sec:5} presents the conclusions.

\section{Process-Based and Deep Learning Models in Agricultural Systems}
\label{sec:1}

PBMs and DL methodologies represent two important but distinct approaches for modeling agriculturals, each possessing inherent limitations (see Table \ref{tab:1H}). The comparative analysis of these two modeling paradigms reveals both congruities and divergences, suggesting that the benefits they offer can be complementary rather than mutually exclusive.
\par

% Please add the following required packages to your document preamble:
\begin{table}[]
\label{tab:1H}
%\caption{The summary of the existing DL-informed PBM models.}

\caption{A Comparative Perspective: PBM vs. DL models}
\centering
\resizebox{6in}{!}{
\begin{tabular}{lll}
\hline
\textbf{Aspect}                                        & \textbf{Process-Based Models}                                                                                                                                                                                                                                                 & \textbf{Deep Learning-Based Models}                                                                                                                                                                                                                \\ \hline
\multirow{3}{*}{\textbf{Modeling Function}}            & \begin{tabular}[c]{@{}l@{}}\textbf{Mechanistic \& Rule-Based}: Explicitly \\ represent the biological, chemical, \\ and physical processes involved in crop \\ growth (e.g., photosynthesis, transpiration,  \\  nutrient uptake).\end{tabular}                                        & \begin{tabular}[c]{@{}l@{}}\textbf{Data-Driven}: Utilize large volumes \\ of data to learn complex, nonlinear \\ relationships between inputs (such as\\ weather, soil properties, and remote \\ sensing imagery) and outputs (like \\ yield).\end{tabular} \\
                                                       & \begin{tabular}[c]{@{}l@{}}\textbf{Process Representation}: Use differential\\  equations and  empirical relationships derived\\  from experimental and field research to\\  simulate crop development over time.\end{tabular}                                                         & \begin{tabular}[c]{@{}l@{}}\textbf{Pattern Recognition}: Rely on neural\\  network architectures that automatically\\  capture interactions among diverse \\ variables without embedding explicit rules.\end{tabular}                                       \\
                                                       & \begin{tabular}[c]{@{}l@{}}\textbf{Scientific Foundation}: Built upon established\\  agronomic and physiological principles that can\\  be related directly to plant behavior under \\ different environmental conditions.\end{tabular}                                                & \begin{tabular}[c]{@{}l@{}}\textbf{Adaptability}: Can integrate a wide\\  range of heterogeneous data\\  sources, making them suitable for\\  rapidly changing or novel scenarios.\end{tabular}                                                             \\ \hline
\multirow{3}{*}{\textbf{Interpretability}}             & \begin{tabular}[c]{@{}l@{}}\textbf{High Transparency}: Each component of the\\  model is linked to a known physiological or\\  ecological process, making it easier for expert\\  to interpret and validate model behavior.\end{tabular}                                               & \begin{tabular}[c]{@{}l@{}}\textbf{Black-Box Nature}: The internal workings \\ of neural networks are often not intuitive, \\ making it difficult to understand exactly how\\  input variables interact to produce the output.\end{tabular}                 \\
                                                       & \begin{tabular}[c]{@{}l@{}}\textbf{Scientific Insight}: Changes in simulated \\ processes (like reduced photosynthesis\\  under drought) can be directly tied to \\ observed plant behavior.\end{tabular}                                                                              & \begin{tabular}[c]{@{}l@{}}\textbf{Post-hoc Interpretability}:   Techniques such \\ as SHAP values, saliency maps, or feature \\ importance rankings are needed to derive \\ explanations.\end{tabular}                                                     \\
                                                       & \begin{tabular}[c]{@{}l@{}}\textbf{Diagnostic   Capability}: Researchers can\\  tweak individual process parameters \\ (e.g., leaf area index) to study their impacts.\end{tabular}                                                                                                    & \begin{tabular}[c]{@{}l@{}}\textbf{Limited Mechanistic Understanding}: predictions \\ may be accurate, they do not provide \\ insights into the specific biological processes.\end{tabular}                                                                 \\ \hline
\multirow{3}{*}{\textbf{Data Requirements}}            & \begin{tabular}[c]{@{}l@{}}\textbf{Moderate Data Needs}:   Requires detailed, \\ specific input parameters (soil properties, \\ weather data,   crop physiology, management,\\  practices) that are usually gathered from \\ field experiments or literature.\end{tabular}              & \begin{tabular}[c]{@{}l@{}}\textbf{High Data Demands}: Needs large datasets \\ for training to capture the variability in crop \\ responses under different environmental and \\ management scenarios.\end{tabular}                                         \\
                                                       & \begin{tabular}[c]{@{}l@{}}\textbf{Expert Knowledge}: Often developed and\\  parameterized using experimental\\  observations and historical data from\\  controlled studies.\end{tabular}                                                                                             & \begin{tabular}[c]{@{}l@{}}\textbf{Data Variety}: Can incorporate diverse datasets, \\ including high-resolution remote sensing,   \\ weather stations, and field sensor data.\end{tabular}                                                                 \\
                                                 \\ \hline
\multirow{3}{*}{\textbf{Calibration and   Validation}} & \begin{tabular}[c]{@{}l@{}}\textbf{Process-Based Calibration}:   Involves tuning \\ model parameters based on field experiments, \\ controlled   studies.\end{tabular}                                                                                                                 & \begin{tabular}[c]{@{}l@{}}\textbf{Performance Metric}: Validation focuses\\  on overall prediction accuracy and \\ might not shed light on individual \\ process dynamics.\end{tabular}   \\
                                                       & \begin{tabular}[c]{@{}l@{}}\textbf{Validation with Observed State Variables} \\
                                                       \textbf{of Relevant Processes}: Model outputs are\\ compared with measured state variables, enabling \\ adjustments to specific components \\ (e.g., canopy development under water stress).\end{tabular} & \begin{tabular}[c]{@{}l@{}}\textbf{Overfitting Risk}: Without careful \\ cross-validation and regularization,\\  the model might perform well on \\ training data but poorly on unseen data.\end{tabular}                                                   \\
                                                       &                            \\ \hline
\textbf{Generalizability and   Flexibility}            & \begin{tabular}[c]{@{}l@{}}\textbf{Robust within Known Conditions}: Excels\\  in scenarios that are well understood \\ within the bounds of the underlying \\ scientific processes.\end{tabular}                                                                                       & \begin{tabular}[c]{@{}l@{}}\textbf{Adaptive Learning}:   Capable of capturing \\ unexpected or complex interactions\\  that are not easily modeled by traditional \\ equations.\end{tabular}                                                                \\
\textbf{}                                              & \begin{tabular}[c]{@{}l@{}}\textbf{Limitation in Novel Conditions}: May struggle\\  when applied to conditions or crops that differ \\ significantly from those used to develop and   \\ validate the model\end{tabular}                                                             & \begin{tabular}[c]{@{}l@{}}\textbf{Flexibility to New Patterns}: Can adapt\\  to changing conditions when retrained with\\  new data.\end{tabular}                                                                                                          \\ \hline
\textbf{Computational   Complexity}                    & \begin{tabular}[c]{@{}l@{}}\textbf{Simulation Complexity}: Once calibrated, \\ simulations are relatively fast; however, the\\  initial calibration and parameter tuning can \\ be computationally intensive.\end{tabular}                                                        & \begin{tabular}[c]{@{}l@{}}\textbf{Training Overhead}:   Requires significant \\ computational resources during the training\\  phase,   especially when using deep \\ architectures on high-resolution or large-volume  \\  datasets.\end{tabular}         \\
\textbf{}                                              & \begin{tabular}[c]{@{}l@{}}\textbf{Resource Use}: Depending on number of grid \\ cells, very high at larger spatial scales \end{tabular}                                                                                                                                                                          & \begin{tabular}[c]{@{}l@{}}\textbf{Fast Inference}: Once the   model is trained, \\ predictions are typically made very quickly,\\  enabling   real-time applications.\end{tabular}                                                                         \\ \hline
\end{tabular}
}
\end{table}

\subsection{Process-based models}

Process-based models (PBMs) rely on mathematical formulations to represent physical phenomena, drawing deductively on well-established physical principles and empirical correlations. These models serve critical functions, including elucidating system behaviors, testing theoretical constructs, and predicting responses to changes in external forces or internal properties \cite{jeong2020process}. PBMs are highly versatile, capable of simulating both observable variables (e.g., volumetric streamflow \cite{wagena2020comparison}, leaf area index \cite{sinha2020estimation}) and latent variables (e.g., groundwater recharge \cite{chen2023evaluating}, fine root distributions \cite{kang2021semantic}). This adaptability is essential for advancing scientific understanding, communicating complex ideas to stakeholders, and supporting informed decision-making processes.

In this section, we provide a synthesis of some typical PBMs, including SIMPLACE \cite{enders2023simplace}, WOFOST \cite{van1989wofost}, APSIM \cite{mccown1995apsim}, DSSAT  \cite{hoogenboom2004decision}, STICS \cite{fraga2015modeling}, and AquaCrop \cite{steduto2009aquacrop}, along with their applications.  \par

\subsubsection{Typical PBM models}

\textbf{SIMPLACE} \par

\begin{figure}[]   
    \centering  
    \includegraphics[width=5.5in]{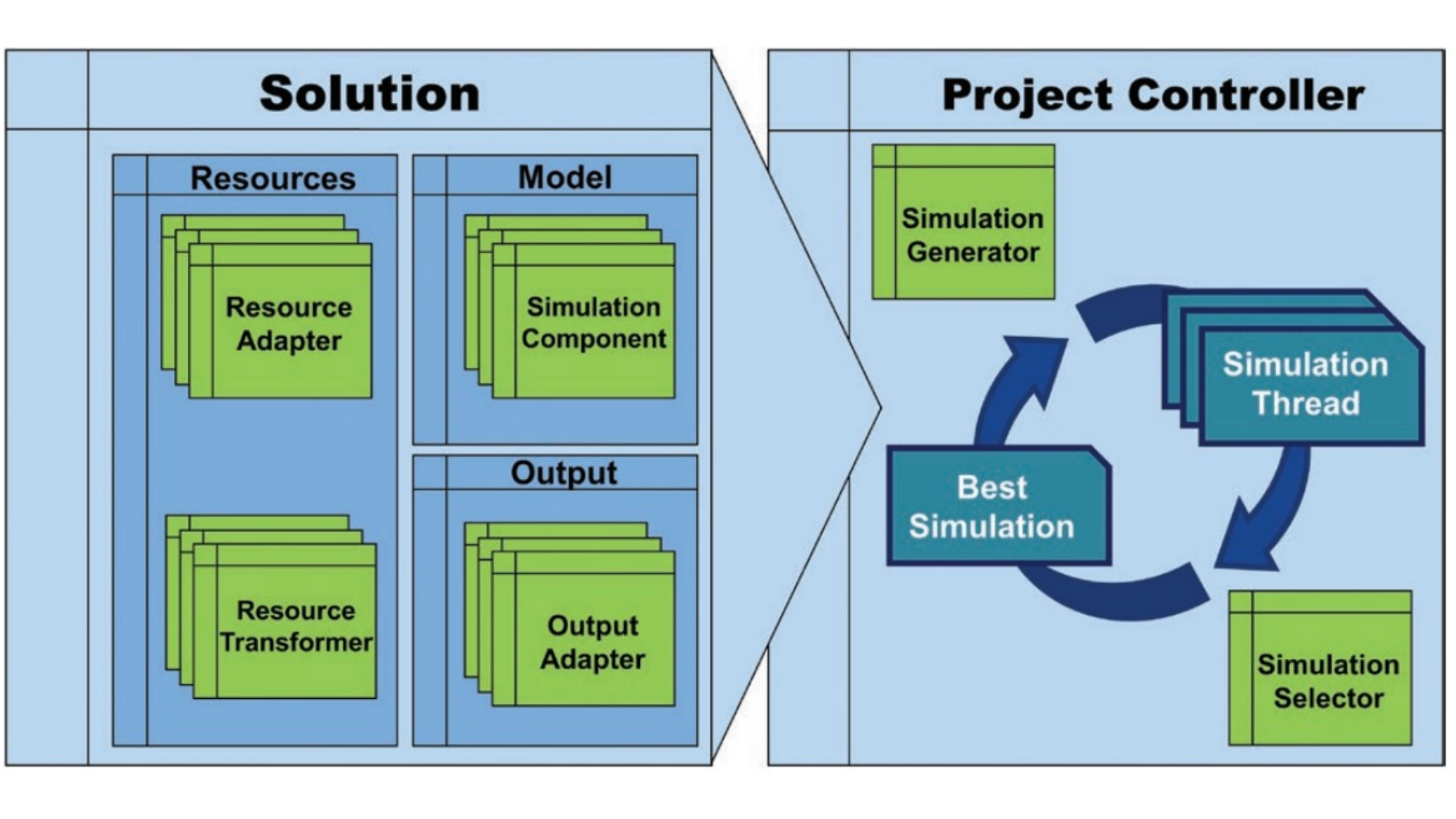}  
    \caption{Simplified overview of a sub-model solution within the SIMPLACE framework \cite{enders2023simplace}.}  
    \label{fig:9_1}  
\end{figure}

SIMPLACE (Scientific Impact assessment and Modelling Platform for Advanced Crop and Ecosystem) \cite{enders2023simplace} is a modular simulation framework, as shown in Fig. \ref{fig:9_1}, designed to capture plant-environment interactions with high precision. SIMPLACE integrate various process-based sub-models and data for simulating agricultural systems. The framework's architecture is modular, allowing for the combination of different sub-model components to address specific research questions \cite{enders2023simplace}. This modularity is achieved through a component-based structure, where each component represents a distinct process or data source within the agricultural system. \par

Mathematically, SIMPLACE employs differential equations to model dynamic processes such as crop growth, soil water balance, and nutrient cycling. These equations are numerically solved within the framework to simulate system behavior over time. Additionally, statistical models are integrated to handle variability and uncertainty in input data and model parameters. The workflow within SIMPLACE involves the selection and configuration of model components relevant to the research objective. Users define input data sources, parameter settings, and simulation scenarios. The framework then executes the simulations, processes the output data, and provides tools for analysis and visualization of the results. This workflow facilitates iterative model development and refinement, supporting comprehensive studies of agricultural systems under various conditions. \par

The versatility of SIMPLACE has been demonstrated across a range of applications. Researchers have employed it to assess the impacts of microclimatic variability on canopy development \cite{webber2018physical}, investigate the partitioning of assimilates under varying environmental conditions \cite{chen2021no}, and evaluate the combined effects of climate change and agronomic practices on field-scale yield predictions \cite{kouadio2021performance}. These studies underscore the model’s capacity to provide robust insights into complex crop-environment interactions and to inform strategies for optimizing resource use and management practices. In addition to its research applications, SIMPLACE offers a transparent and extensible framework that supports collaborative, interdisciplinary efforts in precision agriculture. Its capability to model complex feedback loops within agroecosystems and its customizable architecture position SIMPLACE as a critical tool for advancing both theoretical understanding and practical applications in agricultural science. \par

\textbf{WOFOST} \par

WOFOST (WOrld FOod STudies), implemented in the FORTRAN-77 programming language and is accompanied by a graphical user interface known as the WOFOST Control Centre (WCC), is one of the pioneering process-based crop simulation models, developed at Wageningen University in the late 20th century \cite{van1989wofost}. Renowned for its comprehensive treatment of crop physiology and interactions with environmental factors, WOFOST employs a daily time-step framework to simulate crop growth and development with precision. The model’s core philosophy is to simulate crop behavior under different environmental conditions, by considering factors such as weather, soil, and crop management practices \cite{de201925}, as shown in as shown in Fig. \ref{fig:9_wf},. This approach allows WOFOST to simulate key physiological processes such as photosynthesis, respiration, and carbon allocation in crops, linking them to environmental conditions. \par

\begin{figure}[]   
    \centering  
    \includegraphics[width=5.5in]{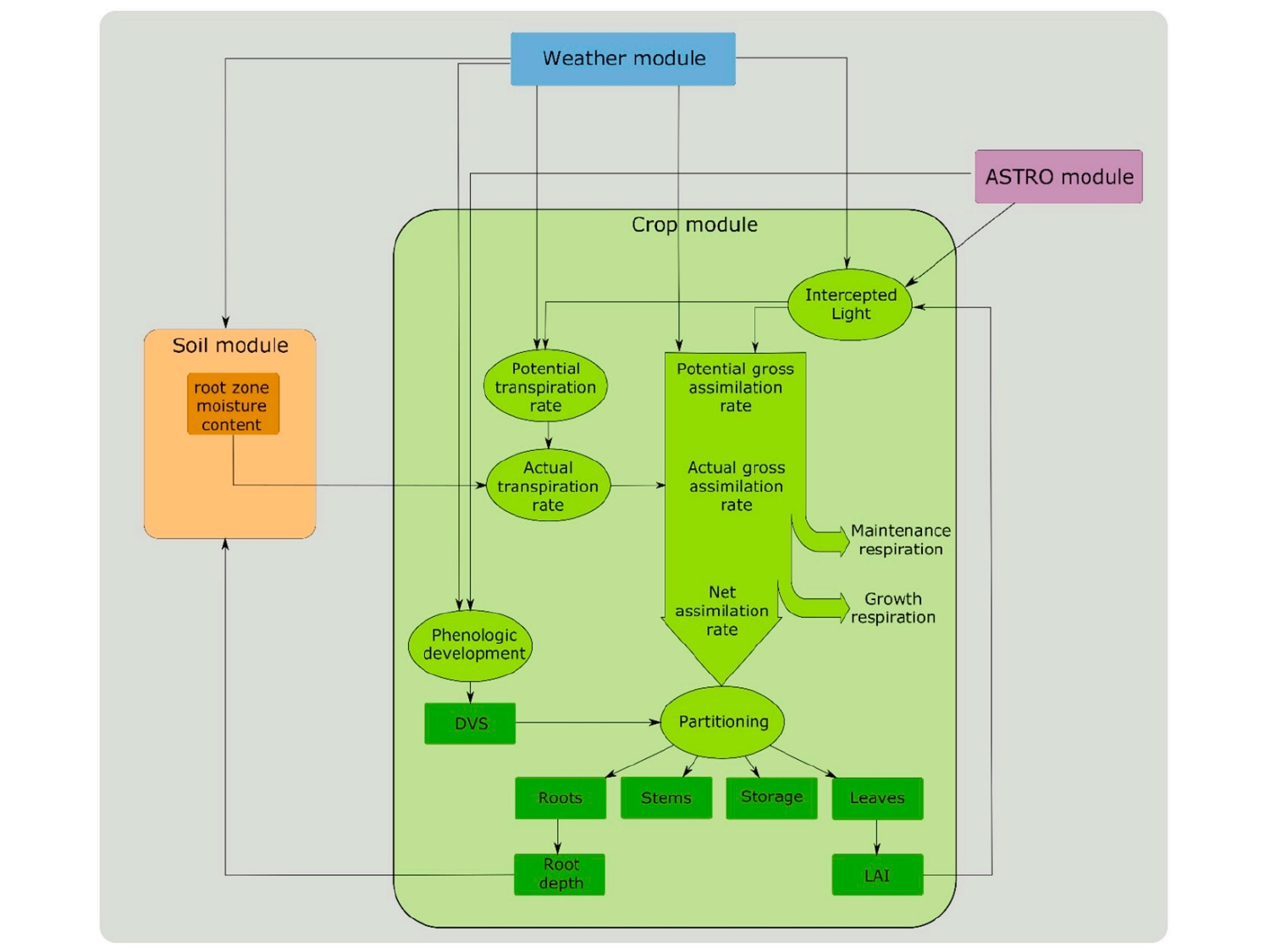}  
    \caption{Schematic overview of the major processes implemented in WOFOST \cite{de201925}.}  
    \label{fig:9_wf}  
\end{figure}

Mathematically, WOFOST employs a range of mathematical equations to represent physiological processes, for example, Gross Carbon dioxide assimilation is calculated based on intercepted photosynthetically active radiation (PAR). The model uses Lambert-Beer's law \cite{su2013lambert} to estimate light distribution within the canopy, accounting for both direct and diffuse PAR. The extinction coefficient is adjusted according to the crop's canopy architecture. The model runs simulations on a daily time step, computing variables like leaf area index (LAI), biomass accumulation, soil moisture content, and yield. Results are analyzed using the WCC's visualization tools, which offer graphical and tabular representations of simulated data \cite{de201925}. Outputs can also be exported for further analysis, allows the model to capture complex interactions between water availability and crop productivity, making it particularly valuable in diverse agricultural settings. WOFOST's modular design and comprehensive representation of crop growth processes make it a valuable tool for researchers and agronomists aiming to assess crop performance under varying environmental and management conditions. \par

\textbf{APSIM} \par

APSIM (Agricultural Production Systems Simulator) is a comprehensive modeling framework renowned for its holistic approach to simulating entire farming systems. Developed through collaborative efforts involving Australian and international researchers \cite{mccown1995apsim}, APSIM integrates a wide range of interrelated processes, including crop growth, soil biogeochemistry, water dynamics, and livestock management, within a unified and flexible platform. At its core, APSIM features a highly modular architecture, allowing users to combine and customize components such as soil water movement, nitrogen and carbon cycling, crop phenology, and economic decision-making \cite{ojeda2017evaluation}. This modularity enables APSIM to adapt to diverse farming systems and environmental conditions, ensuring its applicability across a wide array of research and practical contexts. \par

\begin{figure}[]   
    \centering  
    \includegraphics[width=5.5in]{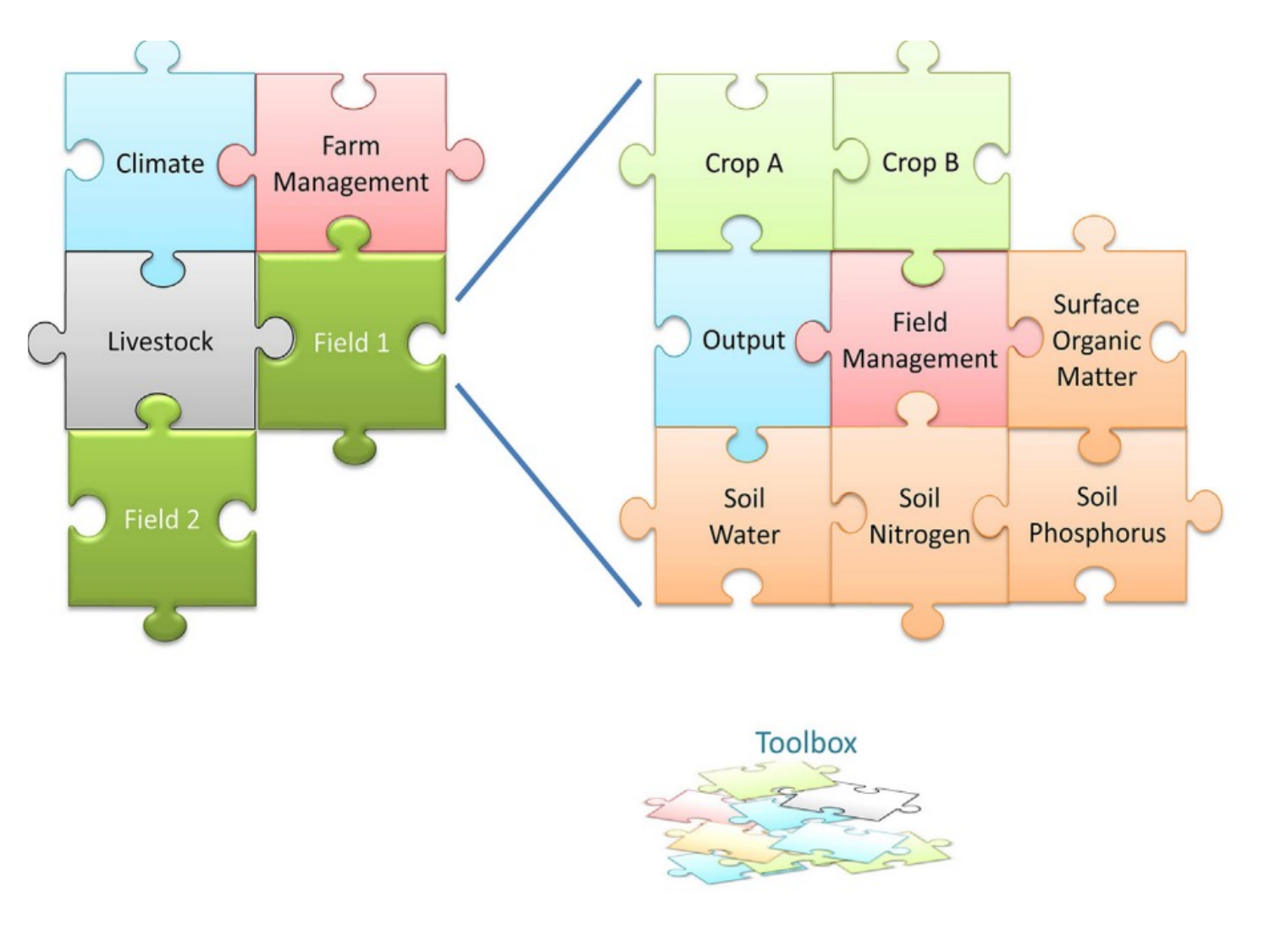}  
    \caption{Schematic overview of an APSIM simulation \cite{holzworth2014apsim}.}  
    \label{fig:9_wf}  
\end{figure}

The model’s systems-level approach provides a robust framework, as shown in Fig. \ref{fig:9_wf}, for evaluating both the immediate effects of agronomic practices on crop yields and the long-term sustainability of farming systems. By simulating processes over multiple growing seasons, APSIM facilitates in-depth analyses of crop rotations, soil health, and nutrient dynamics, as well as assessments of farm-level resilience to risks such as climate variability and market fluctuations. It is particularly adept at capturing the interactions between environmental factors and management decisions, making it an indispensable tool for scenario analyses and policy development \cite{dury2012models}.\par

\textbf{DSSAT} \par

The Decision Support System for Agrotechnology Transfer (DSSAT) is a comprehensive suite of process-based crop simulation models that has been continuously refined over several decades, establishing itself as a cornerstone in agricultural systems modeling \cite{hoogenboom2004decision}. Developed through extensive international collaboration, DSSAT incorporates a robust foundation of crop physiology research, innovative modeling techniques, and empirical calibrations \cite{jones2003dssat}. The suite comprises over 40 crop-specific modules, each meticulously designed to account for the physiological and developmental characteristics of major crop species. These modules enable detailed simulations of key processes, including germination, phenological development, photosynthetic carbon assimilation, biomass partitioning, and yield formation \cite{yan2020simulating}. \par

\begin{figure}[]   
    \centering  
    \includegraphics[width=5.5in]{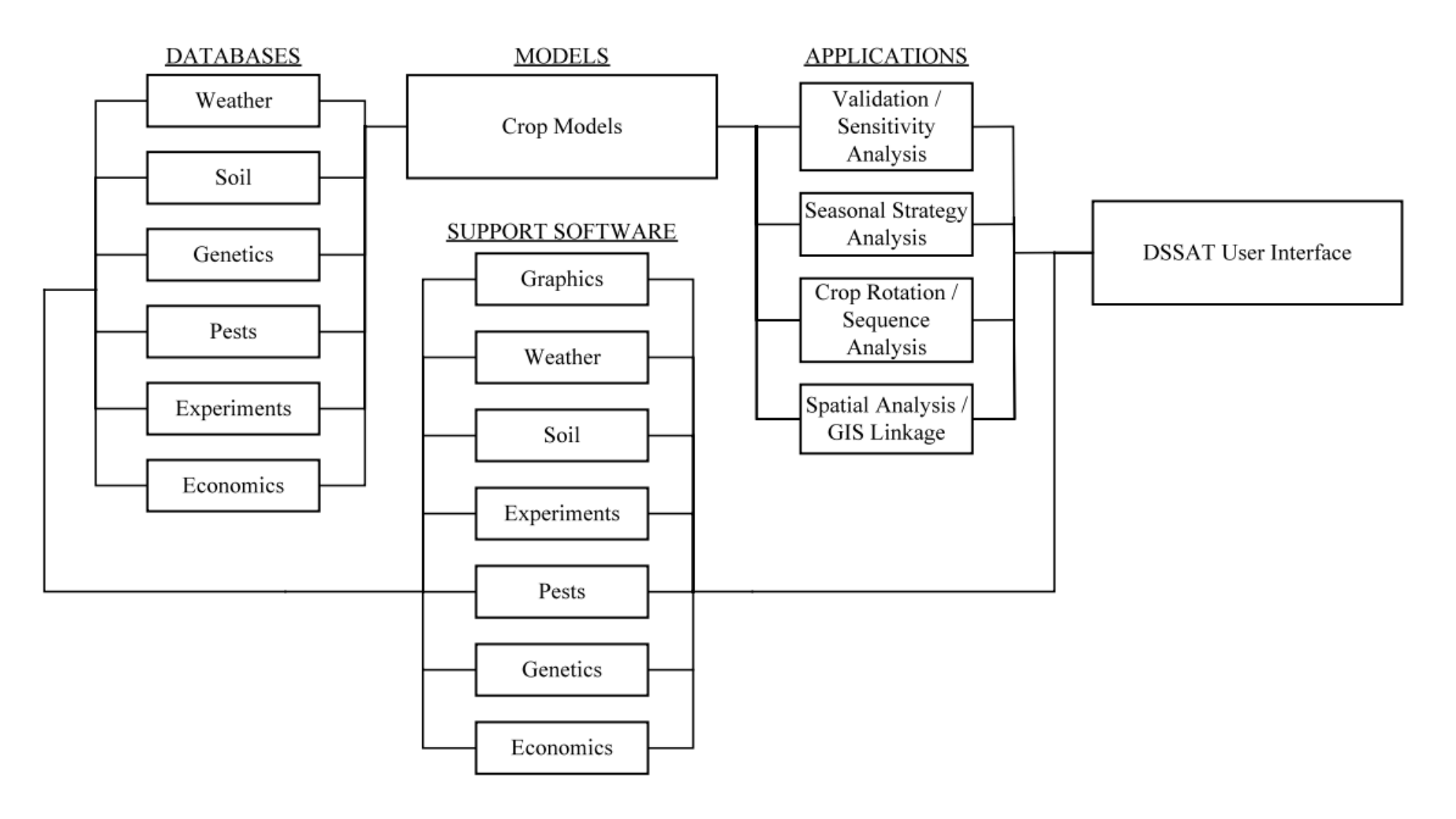}  
    \caption{Schematic overview of an DSSAT simulation \cite{jones2003dssat}.}  
    \label{fig:9_wf}  
\end{figure}

A distinctive feature of DSSAT is its ability to integrate diverse datasets, including daily weather parameters, soil profiles, and management practices such as tillage, irrigation, fertilization, and planting schedules \cite{jones2003dssat}. This integration allows DSSAT to simulate the complex interactions among crops, environmental conditions, and management strategies with high precision. Its soil module, for example, incorporates water and nutrient balance calculations, capturing critical processes like evaporation, transpiration, nutrient uptake, and leaching, which are vital for understanding crop responses under varying conditions \cite{hoogenboom2004decision}.\par

DSSAT has been extensively applied across a wide range of research and operational contexts. These include forecasting seasonal crop yields, evaluating the impacts of climate change, optimizing resource allocation, and developing best management practices for resource-constrained environments \cite{pathak2017application}. Its role in assessing future scenarios—such as the effects of altered climate regimes, shifting planting dates, or new cultivars, makes it an indispensable tool for both strategic planning and policy development \cite{jones2003dssat}. \par

Over the years, DSSAT has undergone rigorous validation using experimental data from diverse agroecological zones and cropping systems worldwide \cite{hoogenboom2004decision}. Its ability to reproduce observed trends and outcomes with high accuracy has solidified its reputation as a reliable and versatile modeling framework \cite{pathak2017application}. In addition to academic research, DSSAT is widely used as a decision-support tool, aiding policymakers, extension agents, and farmers in making informed decisions regarding sustainable agricultural practices. \par

The adaptability and scalability of DSSAT ensure its relevance in tackling global challenges such as food security, resource management, and climate adaptation \cite{jones2003dssat}. By providing a scientifically robust platform for simulating crop-environment-management interactions, DSSAT continues to bridge the gap between empirical research and practical agricultural applications, underscoring its dual role as an investigative model and a pragmatic decision-support system.

%The Decision Support System for Agrotechnology Transfer (DSSAT) is a comprehensive suite of crop simulation models that has evolved over several decades to become a cornerstone in agricultural modeling \cite{hoogenboom2004decision}. Developed through extensive collaborative efforts, DSSAT incorporates a rich legacy of crop physiology research, modeling innovations, and empirical calibrations. Each crop module within the DSSAT suite is tailored to the physiological and developmental idiosyncrasies of major crop species, allowing for detailed simulation of processes such as germination, phenological development, photosynthetic assimilation, and biomass partitioning \cite{yan2020simulating}. The model rigorously integrates daily weather data, soil characteristics, and management practices—including tillage, irrigation, fertilization, and planting dates—to reproduce the complex interactions that govern crop growth under a variety of environmental conditions. Over its long history, DSSAT has been employed in a multitude of research projects that range from forecasting seasonal yields and assessing the potential impacts of climate change to guiding best management practices in resource-limited settings \cite{pathak2017application}. Its extensive validation against experimental data across continents has cemented its status as a reliable and versatile tool for both research and policy formulation, underscoring its dual role as an investigative model and a pragmatic decision-support system.

\textbf{STICS} \par

\begin{figure}[]   
    \centering  
    \includegraphics[width=5.5in]{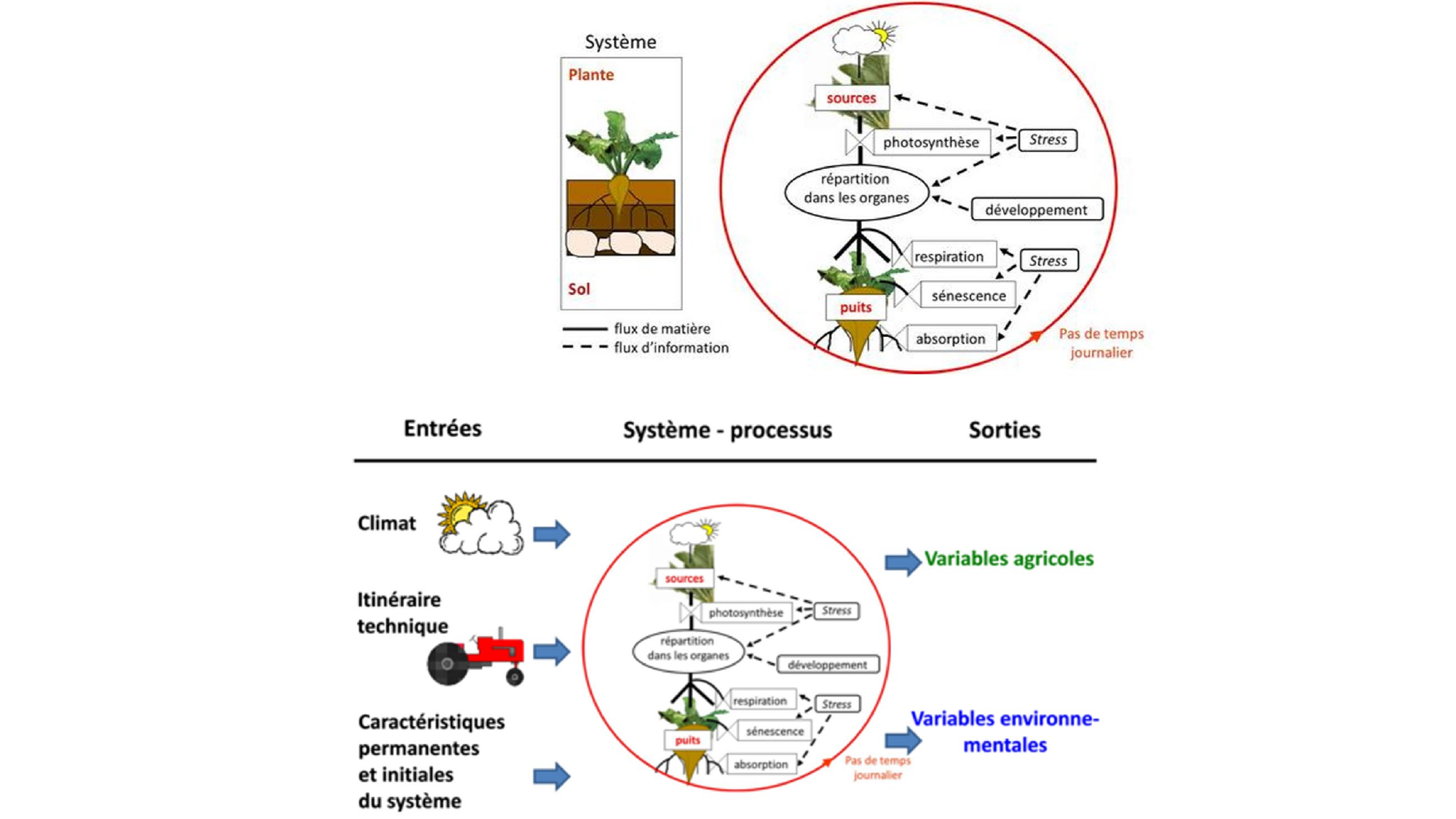}  
    \caption{Schematic overview of an STICS simulation \cite{brisson2004crop}.}  
    \label{fig:9_st}  
\end{figure}

STICS (Simulateur mulTIdisciplinaire pour les Cultures Standard) is a process-based model developed in France that emphasizes the intricate interdependencies between crop growth, soil water dynamics, and nutrient cycling, with a notable focus on nitrogen processes \cite{fraga2015modeling}. Designed to simulate the interactions within the soil-plant-atmosphere continuum, STICS provides detailed and dynamic representations of crop physiological processes and their responses to environmental and management factors, as shown in Fig. \ref{fig:9_st}. Its formulations include finely tuned simulations of photosynthesis, respiration, and assimilate partitioning, enabling the accurate modeling of plant growth under diverse agricultural conditions \cite{brisson2004crop}. \par

A key strength of STICS lies in its advanced water balance module, which captures the temporal variability of soil moisture by integrating processes such as infiltration, evaporation, transpiration, and deep percolation. This allows the model to assess water availability and its effects on crop performance under varying climatic and irrigation scenarios \cite{brisson2003overview}. Additionally, the model incorporates mechanistic submodules for nitrogen cycling, simulating critical processes such as nitrogen uptake, mineralization, nitrification, and leaching. These features enable a nuanced understanding of how nutrient availability and soil dynamics influence crop development, yield formation, and environmental impacts \cite{brisson2004crop}. \par

STICS is particularly well-suited for research on sustainable agricultural practices, offering robust capabilities to evaluate trade-offs between productivity and environmental sustainability. For instance, it has been used to investigate the effects of optimized nitrogen fertilization on crop yields and nitrate leaching, providing valuable insights into reducing environmental contamination without compromising productivity \cite{queyrel2016pesticide}. Its ability to simulate the long-term impacts of soil conservation strategies, such as cover cropping and reduced tillage, further enhances its relevance in agroecological studies.\par

The model’s flexibility and adaptability have allowed its application across diverse cropping systems and environments. It has been extensively validated and calibrated for a wide range of crops, including cereals, legumes, and vegetables, as well as for various climatic zones \cite{jego2015impact}. Moreover, STICS supports scenario analyses for evaluating the impacts of climate change, altered management practices, and policy interventions on crop performance and environmental quality \cite{brisson2004crop}. \par

%STICS (Simulateur mulTIdisciplinaire pour les Cultures Standard) is a process-based model developed in France that emphasizes the intricate interdependencies between crop growth, soil water dynamics, and nutrient cycling, with a notable focus on nitrogen processes \cite{fraga2015modeling}. At the heart of STICS lies a commitment to simulating the detailed interactions between the soil-plant-atmosphere continuum. The model’s formulation includes finely tuned representations of photosynthesis, respiration, and assimilate partitioning, integrated with a dynamic water balance module that captures the temporal variability of soil moisture under varying climatic conditions \cite{brisson2004crop}. Moreover, STICS incorporates mechanistic submodels to simulate nitrogen uptake, mineralization, and losses, providing a nuanced understanding of how nutrient availability modulates crop development and yield formation \cite{brisson2004crop}. This level of detail renders STICS particularly well-suited to studies investigating sustainable fertilization practices, soil conservation strategies, and the trade-offs inherent in water and nutrient management. In practical applications, STICS has been utilized to assess the impacts of different agricultural practices on both crop performance and environmental quality, making it an essential tool for researchers aiming to unravel the complex feedbacks within agroecosystems.

\textbf{AquaCrop} \par

AquaCrop \cite{steduto2009aquacrop}, developed by the Food and Agriculture Organization (FAO), is a process-based crop simulation model specifically designed to address the challenge of water scarcity by focusing on the quantitative relationship between water availability and crop yield \cite{steduto2009aquacrop}. Its development stemmed from the need for a user-friendly, scientifically robust tool capable of evaluating water productivity in a wide range of cropping systems, with a particular emphasis on regions facing water limitations. AquaCrop achieves this by simplifying some of the more complex processes found in other crop models, centering its simulations on transpiration-driven biomass formation while maintaining an adequate balance between simplicity and scientific rigor \cite{abedinpour2012performance}. \par

\begin{figure}[]   
    \centering  
    \includegraphics[width=5.5in]{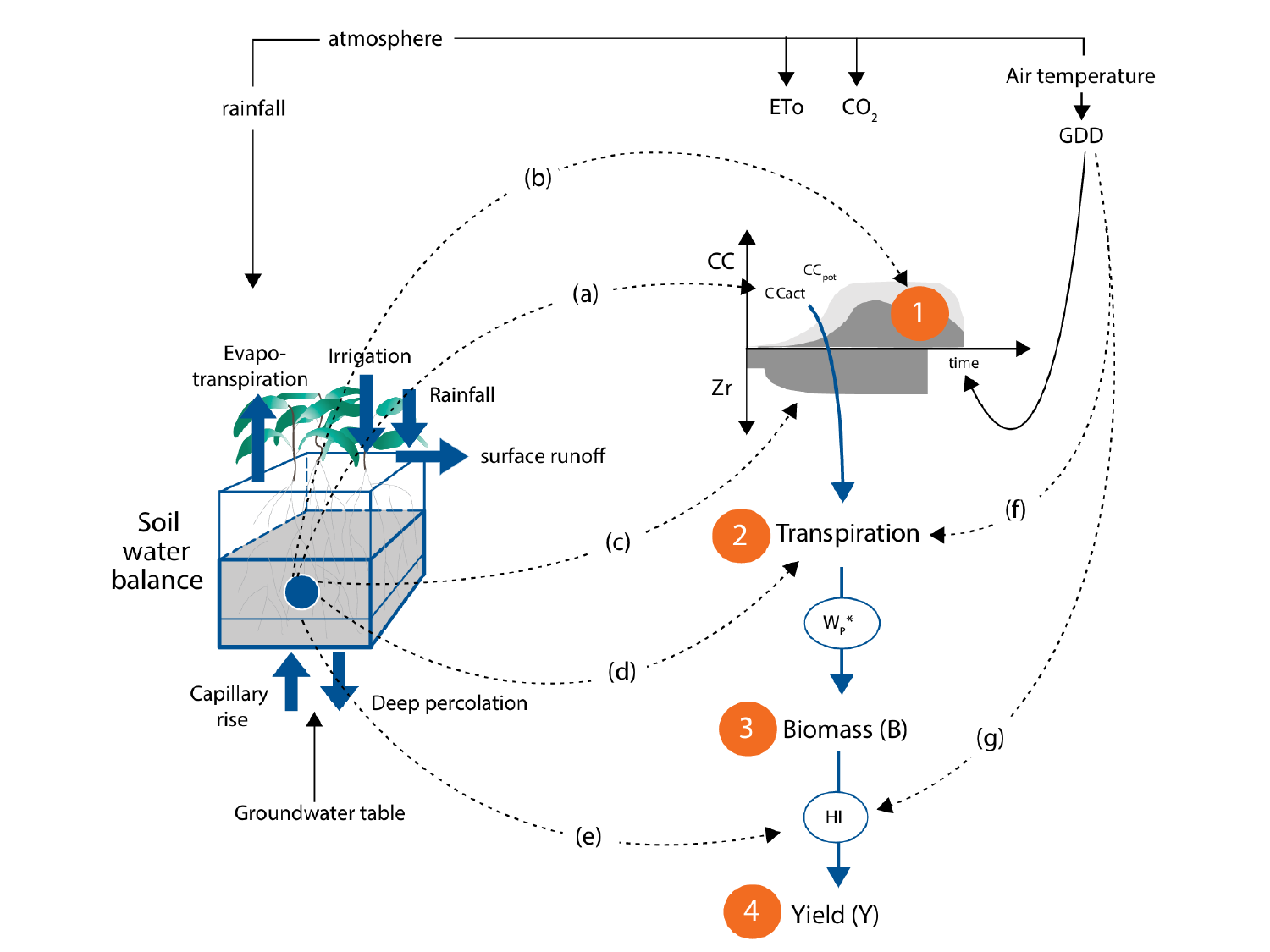}  
    \caption{Calculation scheme of AquaCrop \cite{steduto2009aquacrop}.}  
    \label{fig:9_aquc}  
\end{figure}

At the core of AquaCrop is a streamlined representation of canopy development, soil water dynamics, and water stress effects. As shown in Fig.\ref{fig:9_aquc}, the model simulates critical processes such as evapotranspiration, infiltration, runoff, root water uptake, and soil moisture depletion. These features enable AquaCrop to assess crop performance under varying water availability with high precision, making it applicable to both irrigated and rainfed systems \cite{abedinpour2012performance}. Additionally, AquaCrop explicitly separates evaporation and transpiration in its calculations, allowing for a more nuanced analysis of water-use efficiency. This capability ensures the model’s suitability for scenarios where water resources are a limiting factor. \par

AquaCrop’s simplicity does not compromise its reliability. The model has been rigorously tested and calibrated against experimental datasets from diverse geographic and climatic conditions, demonstrating its accuracy in simulating crop growth and yield under water-limited environments \cite{steduto2009aquacrop}. Its adaptability to a wide range of crops, including cereals, legumes, vegetables, and fruit trees, enhances its utility across varying agricultural systems \cite{steduto2009aquacrop}. \par

The model has been widely employed to evaluate the impacts of different irrigation strategies, forecast yield losses due to drought, and guide policy decisions related to sustainable water resource management \cite{steduto2009concepts}. AquaCrop’s applications extend to scenario analyses for climate change adaptation, enabling researchers to explore the potential effects of altered rainfall patterns, rising temperatures, and extreme weather events on crop productivity and water requirements. Its capacity to model diverse soil types, climatic zones, and management practices further underscores its relevance in addressing global challenges related to water scarcity and food security. \par

\subsubsection{Limitations of process-based model}
Despite the effectiveness of process-based models (PBMs) in simulating complex agricultural systems, they face several significant challenges that limit their applicability and accuracy. These include the inability to rapidly adapt to new data sources, reliance on extensive input data, challenges with calibration, and limited to larger spatial scales or heterogeneous landscapes. Furthermore, uncertainties in model inputs, oversimplifications of system dynamics, and the static nature of many models impede their predictive accuracy and adaptability. Below, we explore these limitations in detail.

\textbf{Slow adaptation to emerging data sources} 

One of the major limitations of PBMs is their inability to swiftly capitalize on burgeoning big data resources, such as those generated by IoT devices, remote sensing, and high-frequency monitoring systems. Developing and validating new process representations and parameterizations is time-consuming, making PBMs less dynamic in integrating novel data streams. Yin \textit{et al.} \cite{yin2022observational} pointed out that discrepancies between model forecasts and empirical data are often addressed through parameter calibration, which can introduce substantial uncertainty and complexity. When model errors exceed the scope of parameter adjustments, omitted processes within governing equations must be hypothesized, followed by structural modifications and iterative validation (an intricate and resource-intensive cycle).

\textbf{High input data requirements and calibration complexity}

PBMs demand a large number of input parameters and variables, including detailed soil properties, crop traits, weather data, and management practices \cite{roukh2020big}. Calibrating these parameters is challenging due to the equifinality problem, where different parameter combinations yield similar outputs, making it difficult to identify the true underlying processes. Additionally, uncertainties in input data, such as soil maps or fertilizer application rates, further complicate calibration and can lead to unreliable predictions \cite{kouadio2021performance}.

\textbf{Limited scalability to large spatial scales}

While PBMs perform well at field or farm scales, their application to larger spatial scales introduces significant challenges. Heterogeneity in soil properties, land management practices, and climatic conditions creates uncertainties that are difficult to address within existing model structures \cite{de201925}. High-resolution soil data and detailed spatial information on crop types and management practices, such as cultivar maps or planting dates, are often unavailable, compounding the issue.

\textbf{Static representations and lack of flexibility}

Many PBMs rely on static assumptions about system processes, such as fixed crop coefficients or simplified representations of biological dynamics \cite{yin2022observational}. This rigidity limits their ability to simulate novel scenarios, such as extreme weather events or non-standard management practices. Furthermore, process representations within PBMs may remain unchanged for extended periods, hindering scientific advancements and reducing model accuracy in evolving agricultural contexts \cite{kouadio2021performance}.

\subsection{Deep Learning-based models in the context of agriculture}

Deep learning (DL) models, particularly deep neural networks (DNNs), have revolutionized agricultural modeling by offering data-driven insights that address many limitations of traditional process-based models. These models excel at learning complex, nonlinear patterns in diverse datasets, making them invaluable for applications such as crop yield prediction and environmental monitoring \cite{mathew2021deep}. For instance, DL models have been used to forecast crop yields \cite{muruganantham2022systematic}, precipitation patterns \cite{li2022using}, and nutrient concentrations such as dissolved oxygen, phosphorus, and nitrogen levels \cite{uddin2021review}. Additionally, they have been applied to various aspects of the hydrological cycle, including soil moisture content \cite{alabdrabalnabi2022machine}, streamflow \cite{lin2021hybrid}, evapotranspiration rates \cite{yuan2020deep}, and groundwater levels \cite{khan2023comprehensive}.

The versatility of DL models lies in their ability to adapt to diverse types of data, including time-series information, geospatial datasets, and image-based inputs. This adaptability is made possible by specialized architectures such as multi-layer perceptron networks (MLPs) \cite{taud2018multilayer}, convolutional neural networks (CNNs) \cite{venkatraman2024channel}, recurrent neural networks (RNNs) \cite{park2023development}, and generative adversarial networks (GANs)\cite{ur2024generative}. These architectures enable DL models to tackle a wide range of agricultural challenges, from plant phenotyping to irrigation management and disease detection.

By leveraging large datasets, DL models consistently outperform traditional methodologies, achieving state-of-the-art results in precision and scalability- even when the underlying biological or physical processes are not fully understood \cite{shi2021biologically}. This capability highlights the power of DL to extract latent patterns and uncover complex relationships from data that may be overlooked by traditional approaches.
The following sections analyze key deep learning architectures and their agricultural applications, highlighting their transformative impact and effectiveness in addressing diverse modeling challenges.

\subsubsection{Key DL architectures}

\textbf{Multi-layer Perceptron (MLP) Network}\par

\begin{figure}[]   
    \centering  
    \includegraphics[width=3.5in]{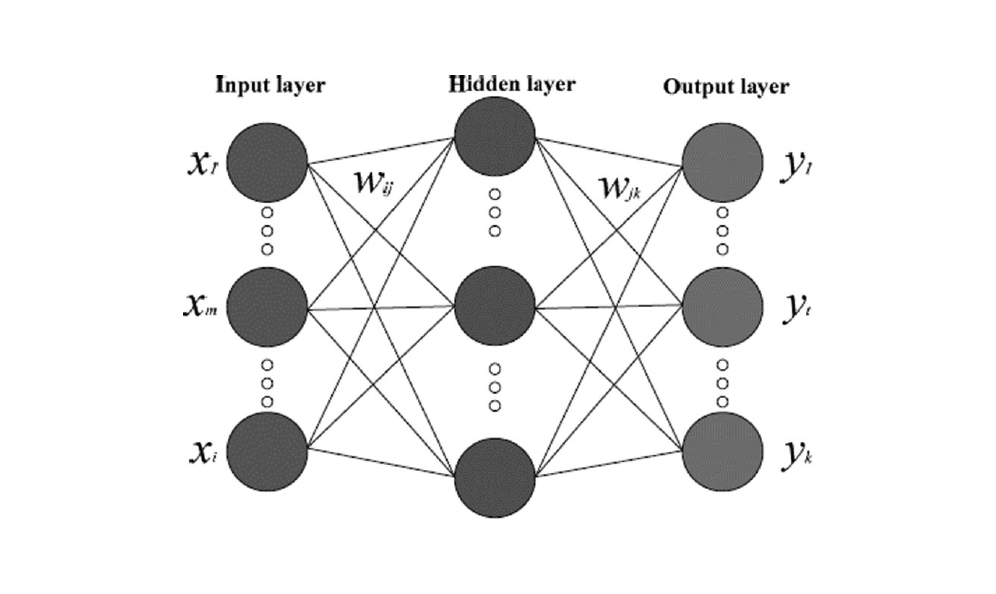}  
    \caption{Schematic overview of an MLP architecture for agriculture study \cite{bazrafshan2022predicting}.}  
    \label{fig:9_mlp}  
\end{figure}

Multi-layer perceptrons (MLPs) \cite{taud2018multilayer}, one of the simplest forms of deep neural networks, consist of neurons organized across multiple layers, enabling computations to flow sequentially from the input layer to the output layer. These networks, equipped with multiple hidden layers, only permit forward data flow and are foundational in deep learning architectures(as shown in Fig. \ref{fig:9_mlp}). While single, fully connected networks are common, recent research has emphasized the integration of multiple networks to address more complex processes or approximate specific equations within larger mathematical frameworks. For instance, Haghighat \textit{et al.} \cite{haghighat2022physics} proposed an architecture consisting of five feed-forward networks to solve specific physics-related problems. Similarly, Cai \textit{et al.} \cite{cai2021physics} utilized deep neural networks to model the interface between different material phases in a two-phase Stefan problem, demonstrating their applicability to multidimensional agricultural challenges.

In agricultural applications, innovative approaches to segment complex systems have emerged. For example, Moseley \textit{et al.} \cite{moseley2023finite} advocated for deploying multiple MLPs, each targeting a specific agricultural subdomain, rather than relying on a single network for the entire system. Further advancements include the development of architectures like DeepONet, introduced by Lu \textit{et al.} \cite{lu2021learning}, which employs two distinct DNNs - one for encoding the input space and the other for the output function domain. This flexibility makes MLPs highly adaptive to diverse agricultural challenges, such as nutrient management and soil-water interactions.

The choice of activation functions plays a critical role in MLP performance, significantly influencing model efficiency and accuracy. Common activation functions like ReLU, Sigmoid, and Tanh have been widely employed, but newer functions, such as Swish, have demonstrated superior performance. Swish, defined as $x \cdot  Sigmoid(\beta x)$ and $\beta$ is a trainable parameter, enhancing the flexibility and potential performance of neural networks \cite{venkatraman2024channel}. The Swish activation function, leveraging a trainable parameter within a sigmoid function has been shown to improve neural networks' convergence rate and accuracy in agriculture application, as evidenced by research conducted by Padhi \cite{padhi2024paddy} and Zhang \cite{zhang2022robustness}. They demonstrated that Swish activation can surpass traditional activation in crop disease classification and agriculture growth monitoring by incorporating it within a ResNet block to bolster the stability of fully-connected neural networks.\par

\textbf{Convolutional Neural Networks (CNNs)}

Convolutional neural networks (CNNs) are specialized architectures for processing grid-like data, making them ideal for analyzing agricultural images such as satellite imagery, UAV data, and plant phenotypes \cite{shahhosseini2021corn}. CNNs extract spatial features through convolutional layers and reduce data dimensionality via pooling layers, achieving high efficiency in tasks like disease detection, weed classification, and crop monitoring \cite{latif2022deep}.

\begin{figure}[]   
    \centering  
    \includegraphics[width=5.5in]{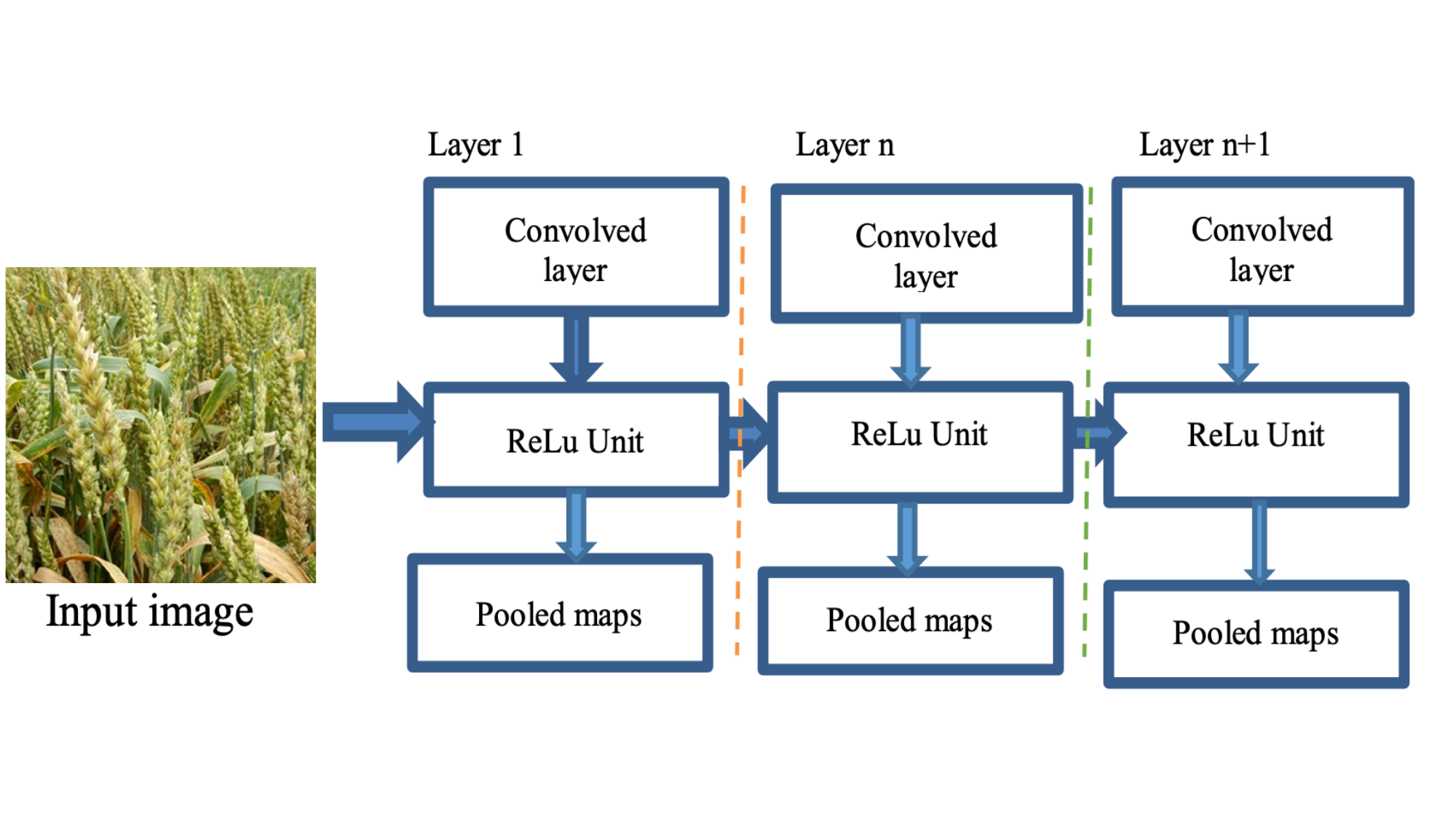}  
    \caption{An example of CNN model architecture in agriculture study \cite{abdullahi2017convolution}.}  
    \label{fig:9_cnn}  
\end{figure}

CNNs in smart agriculture leverage specialized architectures to process grid-like data such as satellite imagery, UAV captures, and plant phenotype images. Fig. \ref{fig:9_cnn} shows an example of CNN model architecture in agriculture stress detection \cite{abdullahi2017convolution}. At the core of these networks are convolutional layers that extract spatial features through operations defined by equations such as:

\begin{equation}
\label{fuc:c1}
S(i,j) = (I * K)(i,j) = \sum_m \sum_n I(i-m, j-n) \, K(m,n),
\end{equation}

where $I$ represents the input data and $K$ is the convolutional kernel. Pooling layers as follow:
\begin{equation}
\label{fuc:c2}
P(i,j) = \max_{(m,n) \in \text{window}} S(m,n)
\end{equation}

which reduce data dimensionality, enhancing computational efficiency. Activation functions like ReLU introduce non-linearity, enabling the modeling of complex patterns essential for tasks such as disease detection, weed classification, and crop monitoring.

Traditional CNNs can struggle with irregular or non-uniform agricultural data. To address this, Zhao \textit{et al.} \cite{zhao2023physics} developed a physics-constrained CNN that transforms irregular agricultural domains into standardized rectangular grids, ensuring compatibility with boundary conditions. Similarly, Faroughi \textit{et al.} \cite{faroughi2024physics} extended CNN adaptability by integrating finite-volume numerical methods, enabling convolution operations over graphs rather than Cartesian grids. These innovations allow CNNs to handle complex agricultural processes, such as soil moisture mapping and hydrological modeling, with enhanced robustness and accuracy.

\textbf{Recurrent Neural Networks (RNNs)}

Recurrent neural networks (RNNs) are designed to capture temporal dependencies in sequential data, making them particularly effective for agricultural time-series applications such as weather forecasting, crop yield predictions, and soil moisture analysis \cite{ndikumana2018deep}. Unlike feed-forward networks, RNNs incorporate memory states, allowing them to account for historical data in their predictions. \par

\begin{figure}[]   
    \centering  
    \includegraphics[width=5.5in]{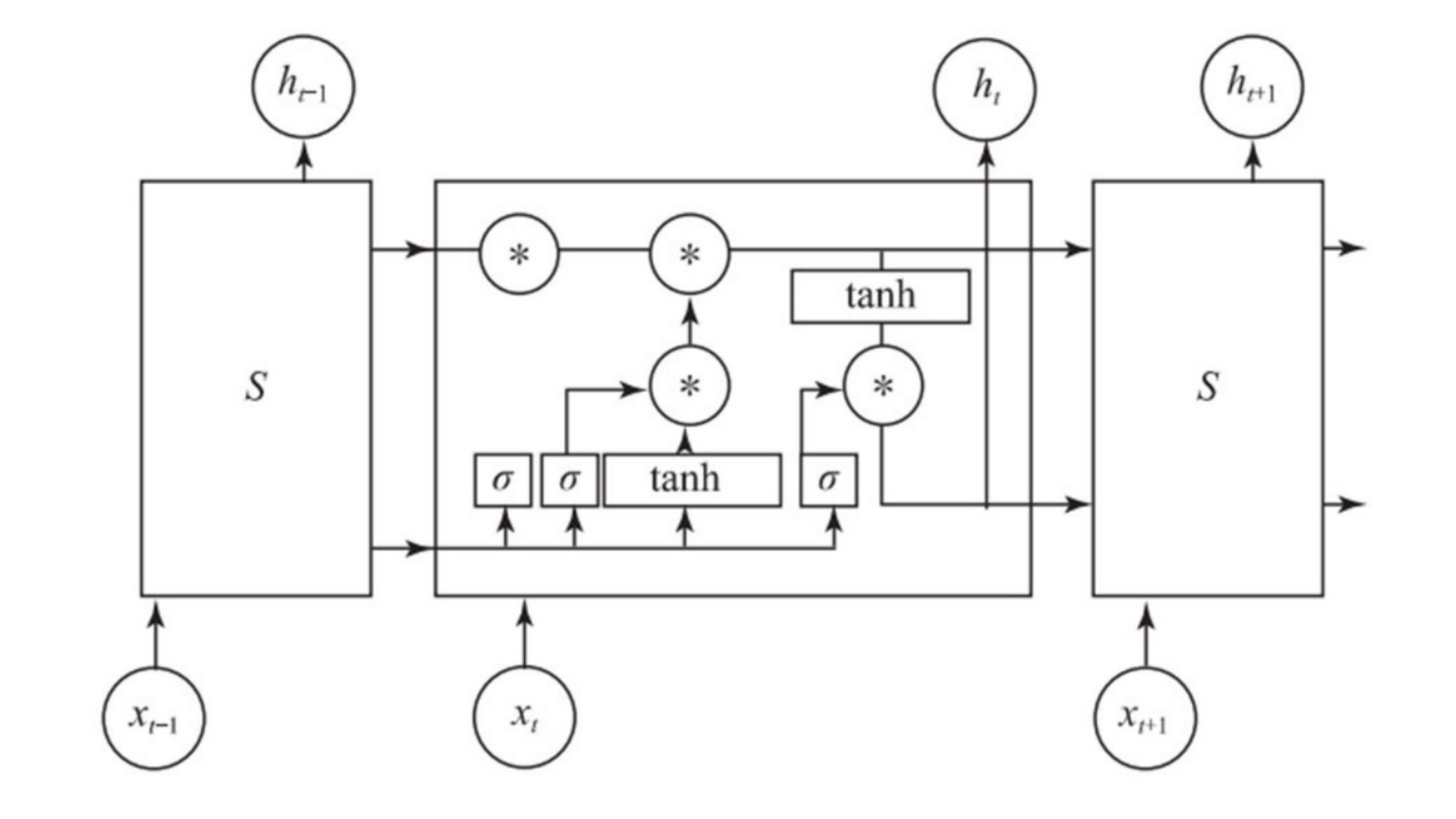}  
    \caption{An example of RNN-LSTM model architecture. \cite{ren2021research}.}  
    \label{fig:9_rnn1}  
\end{figure}

Advanced variants like long short-term memory (LSTM) networks and gated recurrent units (GRUs) address key challenges such as vanishing gradients, enabling accurate modeling over extended time horizons. Fig. \ref{fig:9_rnn1} provide an example of RNN-LSTM model architecture. RNN-LSTM units are pivotal in smart agriculture for handling temporal data from diverse sources such as weather stations, soil moisture sensors, and crop growth records. The LSTM architecture addresses the vanishing and exploding gradient problems typical of standard RNNs, allowing the model to capture long-term dependencies critical for forecasting and sequential decision-making. Key to the LSTM's success are its gating mechanisms, mathematically described by the following equations:
\begin{equation}
\label{fuc:r1}
f_t = \sigma(W_f \cdot [h_{t-1}, x_t] + b_f),
\end{equation}

\begin{equation}
\label{fuc:r2}
i_t = \sigma(W_i \cdot [h_{t-1}, x_t] + b_i),
\end{equation}

\begin{equation}
\label{fuc:r3}
\tilde{C}_t = \tanh(W_C \cdot [h_{t-1}, x_t] + b_C),
\end{equation}

\begin{equation}
\label{fuc:r4}
C_t = f_t \odot C_{t-1} + i_t \odot \tilde{C}_t,
\end{equation}

\begin{equation}
\label{fuc:r5}
o_t = \sigma(W_o \cdot [h_{t-1}, x_t] + b_o),
\end{equation}

\begin{equation}
\label{fuc:r6}
h_t = o_t \odot \tanh(C_t).
\end{equation}

Here, $x_t$ represents the input at time $t$, $h_{t-1}$ is the previous hidden state, $C_t$ is the cell state, $\sigma$ denotes the sigmoid activation function, and $\odot$ indicates element-wise multiplication. These equations facilitate dynamic control over information flow, enabling the model to remember or forget specific aspects of the input sequence based on their relevance to predicting future outcomes. \par

The workflow of an RNN-LSTM model in smart agriculture begins with the acquisition and preprocessing of sequential data, such as time-series sensor readings and meteorological data, which are then normalized and structured for analysis \cite{ren2021research}. The preprocessed data feeds into one or more LSTM layers where the model learns temporal dependencies and trends inherent in the data. This learning process is guided by the aforementioned mathematical formulations and is often supplemented by additional layers—such as dropout layers to prevent overfitting and fully connected layers for output generation. Ultimately, the trained LSTM model is deployed to perform a range of predictive tasks, including yield forecasting, irrigation scheduling, and anomaly detection in crop growth patterns, thereby enabling precision agriculture practices that optimize resource use and enhance crop management. For instance, Bhimavarapu \textit{et al.} \cite{bhimavarapu2023improved} demonstrated that an improved LSTM model could effectively predict crop yields while mitigating overfitting and underfitting issues. Similarly, Wang \textit{et al.} \cite{wang2022winter} used LSTMs to analyze leaf area index (LAI) time-series data, finding them superior to traditional machine learning models in estimating crop properties under varying conditions. These enhancements make RNNs indispensable tools for managing agricultural time-series data and improving predictive accuracy.

\textbf{Generative Adversarial Networks (GANs)}

Generative adversarial networks (GANs) employ a unique adversarial framework, where two neural networks—the generator and the discriminator—compete to create realistic synthetic data while distinguishing it from real data \cite{shi2023physics}. This capability makes GANs highly valuable for addressing data scarcity in agriculture by augmenting datasets with high-quality synthetic samples.

\begin{figure}[]   
    \centering  
    \includegraphics[width=5.5in]{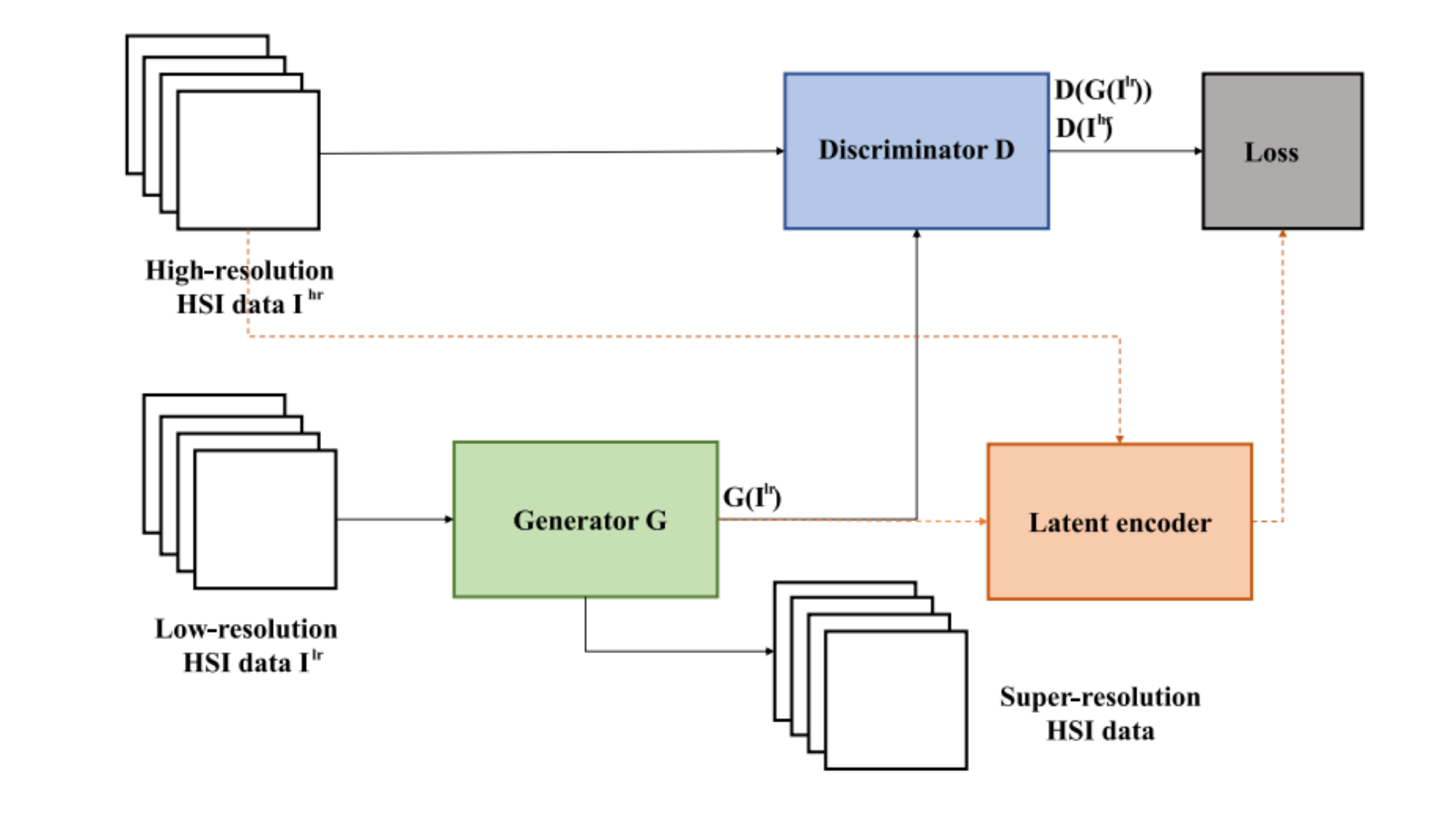}  
    \caption{An example of GAN model architecture for agriculture data super-resolution \cite{shi2022latent}.}  
    \label{fig:9_rnn1}  
\end{figure}

GANs have been increasingly applied in smart agriculture for tasks such as crop monitoring, disease detection, and yield prediction, especially when data availability is limited. The core architecture of a GAN consists of two neural networks: a generator $G$ and a discriminator $D$. The generator learns to synthesize realistic agricultural data (e.g., UAV images, meteorological readings) from a random noise input $z$ drawn from a prior distribution $p_z(z)$, while the discriminator aims to distinguish between real data $x \sim p_{\text{data}}(x)$ and the synthetic data generated by $G$. This adversarial process is mathematically modeled by the following minimax objective function:

\begin{equation}
\label{fuc:g1}
\min_G \max_D \; V(D, G) = \mathbb{E}_{x \sim p_{\text{data}}(x)}[\log D(x)] + \mathbb{E}_{z \sim p_z(z)}[\log (1-D(G(z)))].
\end{equation}

Innovative implementations have been reported in literature, such as Prasad \textit{et al.} \cite{prasad2022two} who used GANs to balance class distributions in UAV image datasets, thereby enhancing the performance of disease classification models. In addition, Zhang \textit{et al.} \cite{zhang2022improving} combined GANs with CNNs to improve winter wheat yield prediction by augmenting training datasets with synthetic meteorological and remote sensing data.\par

The typical workflow in agricultural applications involves first preprocessing limited datasets to standardize inputs for both the generator and discriminator. During training, the generator creates synthetic samples, while the discriminator iteratively learns to differentiate these samples from real data, refining both networks through backpropagation \cite{zhang2022improving}. The adversarial training process continues until a Nash equilibrium is reached, where the generator produces data that the discriminator can no longer reliably distinguish from actual observations \cite{shi2022latent}. This capability to generate high-quality synthetic data addresses data scarcity issues, improves class balance, and enhances model robustness. Consequently, GANs not only facilitate crop monitoring and disease detection but also significantly boost yield prediction accuracy by providing augmented datasets that capture the complex variability inherent in agricultural environments.

\subsubsection{Limitations of deep learning models}

Despite their transformative potential, deep learning (DL) models face several limitations when applied in agricultural contexts, primarily stemming from their dependence on data quality, computational demands, and challenges in generalization and interpretability.

\textbf{Dependence on high-quality data}

DL models rely heavily on large volumes of high-quality, labeled data for effective training. However, agricultural datasets are often sparse, noisy, or geographically constrained, limiting the models' effectiveness. For instance, high-resolution data on soil properties, crop management, or climatic variability is rarely available in many regions, particularly in developing countries \cite{zhang2022improving}. This data scarcity poses significant challenges in accurately predicting crop yields, disease outbreaks, and other critical variables.

\textbf{Lack of causality and interpretability}

While DL models excel at capturing complex patterns, they often operate as "black boxes," providing little insight into how predictions are derived. This lack of interpretability makes it difficult for stakeholders, such as farmers and policymakers, to trust their recommendations. Furthermore, DL models rely on statistical correlations rather than explicit causal relationships, which can limit their robustness in novel scenarios \cite{bazrafshan2022predicting}.

\textbf{Challenges in generalization and handling rare events}

DL models trained on specific datasets often struggle to generalize to new regions, climatic conditions, or crop types. For example, the ability of DL architectures like Long Short-Term Memory (LSTM) networks to model temporal dependencies complements the structured parameterization of PBMs. LSTM models have been successfully employed in hydrological studies, leveraging their capability to handle datasets with hundreds of thousands of parameters \cite{bartz2014evolutionary}. However, these models trained in temperate regions may perform poorly in tropical environments which are inadequate at predicting rare or extreme events, such as severe droughts or pest infestations \cite{shi2022latent}. This lack of generalization and event coverage can significantly hinder their reliability in diverse agricultural settings.

\textbf{Computational and resource demands}

Training and deploying DL models require significant computational resources, such as GPUs or TPUs, which may not be accessible to small-scale farmers or researchers in resource-limited settings \cite{holland1992genetic, wang2018particle}. Moreover, the energy consumption associated with training large models raises concerns about sustainability. These computational demands also necessitate access to robust infrastructure, which is often unavailable in many agricultural contexts.

\textbf{Overfitting and bias}

In scenarios with limited or imbalanced datasets, DL models are prone to overfitting, capturing noise instead of meaningful patterns. This issue is particularly pronounced in agriculture, where datasets often lack diversity or are biased toward specific crops, farming practices, or regions \cite{bazrafshan2022predicting}. Such biases can lead to suboptimal recommendations and predictions that fail to account for real-world variability.

\textbf{Limited domain knowledge integration}

Purely data-driven DL models often fail to incorporate essential domain knowledge, such as biological, ecological, or environmental principles. This lack of explicit knowledge integration can result in models that produce biologically or physically unrealistic predictions, particularly in complex agroecosystems \cite{prasad2022two}. Integrating domain knowledge with data-driven techniques remains an ongoing challenge to improve the reliability and robustness of these models.

\section{Hybrid PBM-DL Modeling}
\label{sec:2}

Hybrid models have emerged as a transformative solution to address the limitations of standalone process-based models (PBMs) and deep learning (DL) models. By seamlessly integrating mechanistic principles with data-driven learning, these models leverage the strengths of both paradigms, enabling more robust and accurate predictions in agricultural systems. This synthesis allows hybrid models to capture the non-linear, multi-scale interactions in agriculturals, such as genotype-environment-management (G×E×M) dynamics, which are often challenging for PBMs or DL models alone \cite{sahoo2019long}. For instance, hybrid approaches have demonstrated superior capabilities in predicting crop performance across diverse climate scenarios by incorporating unique genetic and management factors \cite{liu2024hybrid}. In this section, we first examine the existing categories of hybrid model architectures, followed by an exploration of their applications in agricultural research.

\subsection{Existing Hybrid Modelling Architectures}
\label{subsec:hybridarchitecture}
Hybrid PBM-DL architectures leverage the complementary strengths of both paradigms. The foundation of the PBM-DL hybrid model lies in differentiable modeling (DM), which represents a continuum of hybrid approaches integrating process-based and DL techniques. DM leverages various methods to efficiently and accurately compute gradients, enabling large-scale optimization of the combined system. A key characteristic of DM is its emphasis on end-to-end differentiability, ensuring that the entire modeling framework remains conducive to gradient-based learning. This allows pre-trained DL components to dynamically adapt and evolve in response to new data. \par

%In this study, depending on how the DL-driven DM interacts with the PBMs, the hybrid model can be categorized into two primary frameworks: 1) DL-informed PBMs:   PBMs are enhanced with DL components to model complex, computationally intensive, or poorly understood processes. and 2) PBM-informed DL models.   
Depending on the interaction between PBMs and DL components, hybrid models can be categorized into two primary frameworks: \par 
1) DL-informed PBMs: DL components are embedded within PBMs to enhance their ability to model intricate, poorly understood, or computationally intensive processes. By learning nonlinear patterns from historical data, DL augments traditional PBMs, improving their predictive accuracy and adaptability \cite{abdel2024proposed}. This bidirectional integration of PBM and DL enables hybrid models to balance physical interpretability with data-driven flexibility, making them powerful tools for tackling complex environmental and biological systems. \par
2) PBM-informed DL models: PBMs serve as structural constraints that guide DL models, ensuring biologically meaningful predictions while reducing the search space for optimization \cite{brown2022designing}. This approach leverages mechanistic components, such as soil moisture dynamics or crop growth processes, to steer the learning process of DL models, allowing them to focus on complex high-dimensional patterns. A more details investigation of the hybrid architectures and approaches are described below.   \par

Each approach offers distinct advantages, balancing physical interpretability with the flexibility of data-driven learning.

%In DL-informed PBMs, DL components are embedded within PBMs to enhance their ability to model intricate, poorly understood, or computationally intensive processes. By learning nonlinear patterns from historical data, DL augments traditional PBMs, improving their predictive accuracy and adaptability \cite{abdel2024proposed}. This bidirectional integration of PBM and DL enables hybrid models to balance physical interpretability with data-driven flexibility, making them powerful tools for tackling complex environmental and biological systems. Conversely, in PBM-informed DL models, PBMs serve as structural constraints that guide DL models, ensuring biologically meaningful predictions while reducing the search space for optimization \cite{brown2022designing}. This approach leverages mechanistic components, such as soil moisture dynamics or crop growth processes, to steer the learning process of DL models, allowing them to focus on complex high-dimensional patterns. A more details investigation of the hybrid architectures and approaches are described below.

\subsubsection{DL-informed PBMs}
\label{subsub:DL-Informed}

DL-informed PBMs integrate domain-specific mechanistic models with deep learning (DL) to enhance predictive performance and adaptability \cite{zhou2023application}. While traditional PBMs rely solely on predefined equations and fixed structural assumptions, they often struggle to capture the dynamic and nonlinear interactions inherent in agricultural systems. In contrast, DL-informed PBMs incorporate data-driven learning to refine parameters, improve process simulation, and optimize model behavior under changing conditions \cite{muruganantham2022systematic}.
A core enabler of this hybrid approach is differentiable modeling, which provides two key advantages:

\begin{itemize}
    \item Constrained Learning: It restricts DL training to a structured search space informed by process-based priors, ensuring that the PBM's mechanistic structure remains intact and scientifically grounded \cite{wang2021modelling}.
    \item Model Augmentation: It allows the seamless integration of learnable components, extending the PBM’s capabilities beyond traditional parameterizations. This enables hybrid models to perform robustly even under data-scarce conditions, while capturing complex and emergent system behaviors.
\end{itemize}

Mathematically, the integration can be framed as an optimization problem \cite{baydin2018automatic}, where the goal is to minimize the discrepancy between the model's predictions and observed data. This is achieved by adjusting the parameters or functions within the model that are identified as uncertain or poorly understood. This idea can be explained in concise mathematical terms using a PBM:

\begin{equation}
\label{fuc:1}
y=g(u, x, \theta)
\end{equation}

where $y$ is the environmental variable to be predicted, and $u$, $x$, and $\theta$ represent state variables, dynamic forcings, and physical parameters, respectively. This representation of a PBMs is generic and encompasses differential equations, for example:
\begin{equation}
\label{fuc:2}
\frac{\partial u}{\partial t} = g(u, x, \theta)
\end{equation}

\begin{figure}[]   
    \centering  
    \includegraphics[width=5.5in]{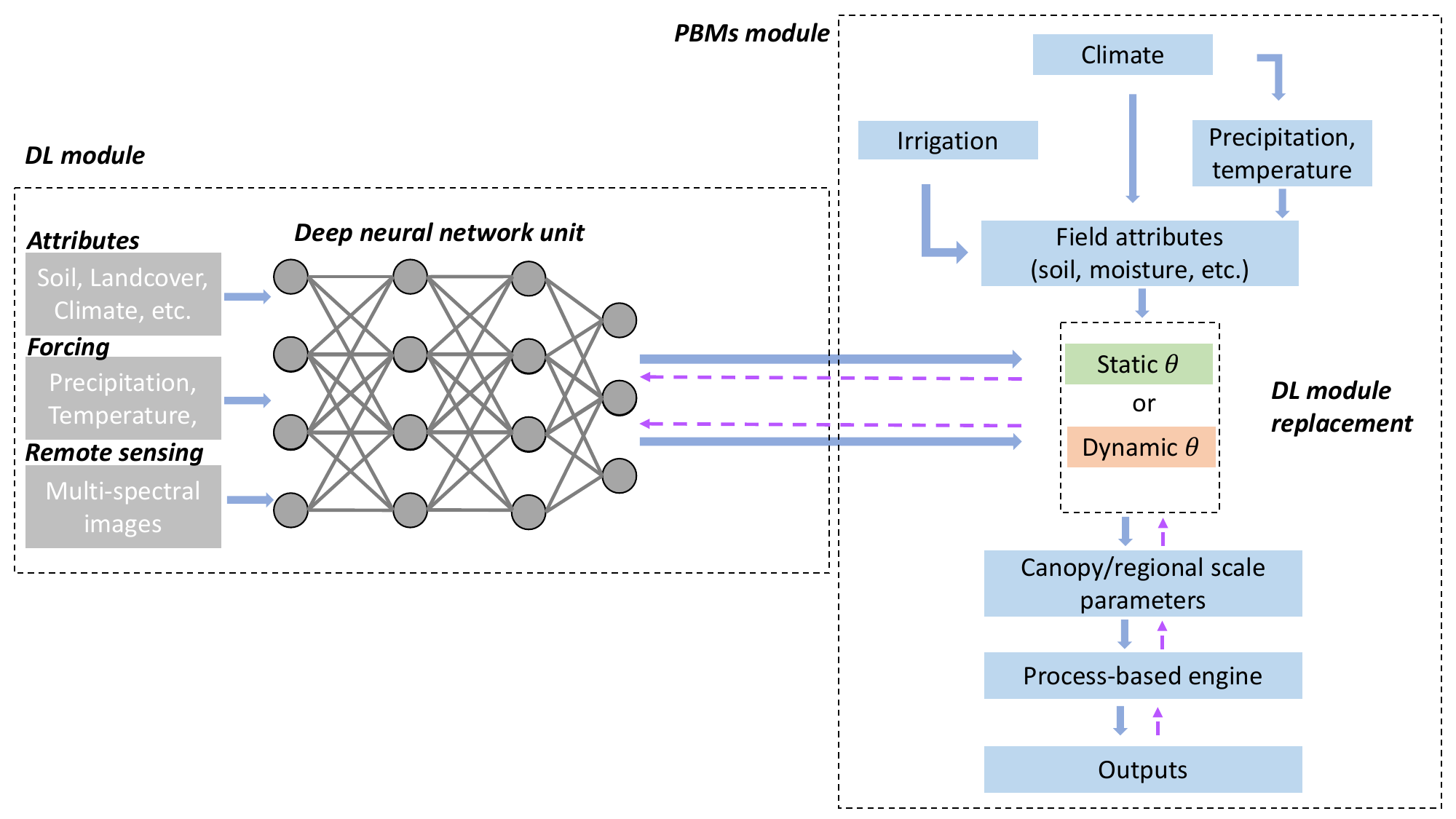}  
    \caption{The general workflow of DL-informed PBMs approach.}  
    \label{fig:3a}  
\end{figure}

%Based on this mathematic basis, we survey preliminary ventures into DL-informed PBMs methods, classified by gradient computation and domain-knowledge application. The examples provided, while not exhaustive, aim to clarify the concept and stimulate further innovative efforts. The general workflow of DL-informed PBMs is shown in Fig. \ref{fig:3a}.  The representative DL-informed PBMs techniques are summarized in Table \ref{tab:1bb} and described below. 
Building on this mathematical foundation, we examine the existing approaches in DL-informed PBMs, classifying them based on their method of gradient computation and domain knowledge integration. These methods generally fall into two broad categories: 1) Differentiating through numerical models and 2) DL-dominant hybrid models with limited physics. While the following examples are not exhaustive, they illustrate how deep learning (DL) can enhance process-based modeling, and they encourage further research in this area. The general workflow of DL-informed PBMs is depicted in Fig. \ref{fig:3a}, with representative techniques summarized in Table 2.

\begin{table}[]
\label{tab:DLP}
\caption{Some typical existing DL-informed PBM models.}
\centering
\resizebox{6in}{!}{
\begin{tabular}{ccc}
\toprule
\textbf{Model   Name} & \textbf{DL Involved}                     & \textbf{Application Area}                                       \\ \midrule
APSIM                 & Random Forest, Deep Neural Network (DNN) & Crop growth modeling \cite{feng2019incorporating}      \\
Tumaini               & Convolutional Neural Network (CNN)       & Disease and pest detection \cite{choudhary2023non}             \\
Melisa                & Long Short-Term Memory (LSTM)            & Yield predictions \cite{perdomo2022literature}              \\
CO2FIX                & Convolutional Neural Network (CNN)       & Carbon monitoring and biomass prediction  \cite{terasaki2023priority} \\
Artemis               & CNN               & Crop phenotyping  \cite{baker2023artemis}       \\ \bottomrule       
\end{tabular}
}
\end{table}

\textbf{1) Differentiating through numerical models}\par

The most straightforward DL-informed PBMs approach is differentiating through numerical models. For this kind of method, DM facilitate tracking gradients at fundamental operational levels. Wherein, DM provides a readily accessible avenue for transforming existing models into the PDM-DL framework. For models devoid of iterative solvers, DL platforms such as PyTorch, Julia, or JAX can retrieve existing physical models, originally developed in languages like Fortran or C/C++ into differentiable versions via AD, ensuring reproducibility. These DM approach can seamlessly integrate with DL by inputting data into the network and incorporating its outputs into the broader PBM framework, thereby efficiently simulating physical laws under the given conditions. This approach's gleaned relationships can readily augment existing models, enhancing operational tasks like flood forecasting or agricultural productivity projections.\par

A case study in agricultural modeling involves adapting the Hydrologiska Byråns Vattenbalansavdelning (HBV) hydrological model, restructured on PyTorch and linked with NNs for regionalized parameterization, this reimagined HBV model demonstrates streamflow simulation capabilities comparable to LSTM models \cite{seibert2021retrospective}. In Signh \textit{et al.} \cite{singh2021soil}'s study, by substituting the soil moisture-runoff relationship with an NN, the developed model learned the interplay between soil moisture, precipitation, and runoff for watershed systems exhibiting threshold behaviors, this model outputs variables like evapotranspiration and baseflow, aligning with alternative estimates and showcasing enhanced numerical precision and parameter robustness through adjoint backward functions for implicit time-stepping. Similarly, the EXP-HYDRO hydrological model, proposed by Li \textit{et al.} \cite{li2023enhancing}, recorded as a recurrent NN and coupled with fully connected NNs, exhibits robust transferability across basins. Grigorian \textit{et al.} \cite{grigorian2024hybrid} proposed the Hybrid neural ODE methods, where NNs substitute hydrological differential equations, to yield more precise predictions than basin-specific LSTMs, retaining the interpretability of mechanistic models. Differentiable models have also facilitated biogeophysical and ecosystem modeling improvements, particularly in photosynthesis parameterization at larger scales.\par

% Beyond ODE-based models, direct differentiation applies to graph-based natural system representations, such as river networks, where an advective dispersion equation modeling streamwater temperature outperforms LSTM under data-scarcity conditions. An innovative differentiable river routing model learning Manning’s roughness coefficient parameterization from daily discharge data exemplifies the potential for nuanced catchment area relationships. Moreover, an adjoint-centric approach, employing NNs to approximate unknown PDE functions or operators before discretizing via finite element methods and leveraging adjoint methods for gradient calculation, has elucidated nonlinear coefficients for equations like Poisson and heat equations. Overcoming Newton iteration convergence challenges, an operator-splitting strategy discretizes the PDE into differential operator and NN subproblems, facilitating precise integration through Gaussian quadrature. This methodology extends to agricultural equations like subsurface reactive transport, illustrating the breadth and adaptability of the PDM-DL model in addressing complex environmental and agricultural challenges.\par
\textbf{2) DL-dominant hybrid models with limited physics}\par

An alternative category of DL-informed PBMs primarily employs deep learning for modeling while incorporating physical operators to enforce constraints derived from fundamental physical laws. A representative example of this approach is the use of Long Short-Term Memory (LSTM) networks to estimate physical surface fluxes such as evaporation, runoff, and recharge, guided solely by mass balance equations \cite{zhou2023application}. Initially, these flux estimations were constrained only by discharge observations, raising concerns about whether the predicted flux terms retained their physical significance. For instance, Huang \textit{et al.} \cite{huang2022novel} demonstrated how incorporating dual data sources significantly enhanced model performance, improving the correlation with observations from 0.8 to 0.91 in soil moisture prediction tasks. The model’s effectiveness was further validated by evaluating loss functions at multiple resolutions.\par

To refine this approach, additional observational data were integrated to further strengthen PBM constraints, enhancing the physical consistency of the system. For example, Reichstein \textit{et al.} \cite{reichstein2019deep} highlighted the importance of integrating observational data into process-based models to improve their predictive accuracy and physical consistency. The authors argue that observational data can help constrain model parameters, reducing uncertainties and enhancing the model's ability to simulate real-world processes. Schuur \textit{et al.} \cite{schuur2015climate} discussed how integrating observational data from permafrost regions into PBM models helps refine predictions of carbon feedbacks. The additional data provide stronger constraints on model parameters, enhancing the physical consistency of the system. While DL-dominant models prove to be powerful predictive tools within the PBMs framework, their interpretability and the physical validity of intermediate diagnostic variables must be rigorously assessed. Thus, this approach highlights the delicate balance between leveraging the predictive power of deep learning and maintaining adherence to established physical principles. \par

\subsubsection{PBMs-informed DL}
\label{subsub:PBMS-InformedDL}

PBM-informed DL represents a transformative approach in artificial intelligence, where PBMs enhance data-driven DL by embedding structured domain knowledge into the learning process. This approach enables DL systems to operate more efficiently and with improved scientific validity, especially in complex agricultural fields. Unlike their purely data-driven counterparts, PBMs-informed DL exchange a degree of generality for enhanced interpretability and the capacity to probe specific phenomena, without necessarily compromising on accuracy. For example, Brown \textit{et al} \cite{brown2022designing} presented a hybrid modeling approach that combines mechanistic models of pest population dynamics with machine learning techniques to predict pest outbreaks. The hybrid model is shown to outperform both purely mechanistic and purely data-driven models in terms of accuracy and predictive power. Abdel \textit{et al} \cite{abdel2024proposed} proposed a hybrid model for crop yield prediction that integrates a mechanistic crop growth model with machine learning techniques. The hybrid model is able to capture both the underlying biological processes and the complex patterns in historical data, leading to improved prediction accuracy.\par 

%\textcolor{blue}{\textbf{( what do you mean here see below.  should the below" the DL-driven DM approach" be PB, informed DL?)}}

Mathematically, the approach introduces a paradigm shift, offering the capability to directly interrogate and potentially revise the functional form of the model itself based on observed data \cite{van2024forward}. This approach diverges significantly from traditional methodologies by not merely adjusting parameters within a fixed framework but allowing the model's structural assumptions to evolve in response to empirical evidence as follows:

\begin{equation}
\label{fuc:3}
y = NN^{w}(u, x, \theta)
\end{equation}

where $W$ represents the high-dimensional weights. The function that is estimated with this approach could also be a parameterization scheme, as in differentiable parameter learning, for example:
\begin{equation}
\label{fuc:4}
y=g(u,x,\theta=NN^{w}(A))
\end{equation}

where $NN$ is a network $A$ is some raw information relevant to the physical parameters $\theta$. DM makes it possible to place questions precisely in the model, to extract fine-grained relationships from data. For example, for a model written simply as:
\begin{equation}
\label{fuc:5}
y=g(g_1,g_2,g_3(u,x,\theta))
\end{equation}

where $(g_1,g_2,g_3)$ are process equations as subcomponents of the model, $g_3$ can be replaced with an Network:
\begin{equation}
\label{fuc:6}
\frac{\partial u}{\partial t}=g(g1,g2,NN^{W}(u,x,\theta))
\end{equation}

The differential equation terms will be integrated in time using numerical approaches. $NN^{W}$ could represent rainfall–runoff relationships, or a constitutive relationship producing effective hydraulic conductivities in a subsurface reactive transport model. \par

Within the aforementioned equations, the equations delineating the physical processes act as a foundational structure (or inductive bias) for the composite model: in equation \ref{fuc:4}, the fundamental structure is denoted by $g(\cdot)$; in equation \ref{fuc:5} and \ref{fuc:6}, it is represented by $g$, $g_1$, and $g_2$. These invariant elements (structural priors) such as $g_1$, and $g_2$ function as physical constraints within the model framework. Enhanced understanding of the model can be achieved either through the visualization of the relationships elucidated by the DL. Furthermore, this approach facilitates the refinement of process representations for certain components of the model, such as $g_3$, thereby enabling inquiries to be conducted with greater specificity and adaptability. Additionally, while a purely data-driven DL is tasked with learning a direct mapping from $x$ to $y$, encapsulating multiple processes simultaneously, this can obscure the interpretability of the outcomes. By segmenting this mapping into discrete components and incorporating prior knowledge, both the scope of learning and the complexity of the relationships being modeled are effectively diminished. This not only enhances the interpretability but also bolsters the robustness of the model's deductions.\par

In this section, we categorize the PBM-informed DL approach into two sub-categories: 1) Surrogate Models, 2) Physics-Informed Models. Surrogate Models leverage PBMs to create faster, computationally efficient alternatives, replacing complex simulations with simplified DL versions that retain essential characteristics \cite{ma2023multisource}. Physics-Informed Models integrate the fundamental laws governing the system (like conservation laws) directly into the DL's architecture, ensuring that predictions stay physically consistent, even in data-scarce environments \cite{raissi2017physics}. This classification reflects the versatility and scientific rigor of PBM-informed DL, making it a robust framework for creating reliable, interpretable, and highly accurate DL models across diverse scientific and engineering applications. To date, several PBM-informed DL techniques have been proposed to address various types of research questions in different scenarios. These techniques are summarized in Table 3 and described below\par

\begin{table}[]
\label{tab:1a}
\caption{Some typical existing PBM-informed DL models.}
\centering
\resizebox{6in}{!}{
\begin{tabular}{cccc}
\hline
\textbf{Model   Name} & \textbf{PBM Involved}                             & \textbf{Network Architecture}           & \textbf{Application Area}                                                            \\ \hline
APSIM-DNN             & APSIM          & Deep Neural Network (DNN)               & Crop yield prediction \cite{shahhosseini2021coupling, kheir2023integrating}       \\
CO2FIX Hybrid         & CO2FIX         & Convolutional Neural Network (CNN)      & Biomass estimation \cite{kotwal2024applying}  \\
Yield-SAFE DL         & Yield-SAFE     & Fully Connected Neural Network (FCNN)   & Yield prediction \cite{airaldi2023learning}\\
LINTUL3-LSTM          & LINTUL3        & Long Short-Term Memory (LSTM)           & Crop growth \cite{wu2022optimizing}  \\
Hi-sAFe Model         & Hi-sAFe        & 3D CNN                                  & Yield prediction \cite{malezieux2009mixing} \\
WaNuLCAS-ML           & WaNuLCAS       & Multi-layer Perceptron (MLP)            & Water and nutrient balance \cite{czembor2022simulating}\\ \hline
\end{tabular}
}
\end{table}

\textbf{1) Surrogate models}\par
\begin{figure}[]   
    \centering  
    \includegraphics[width=5.5in]{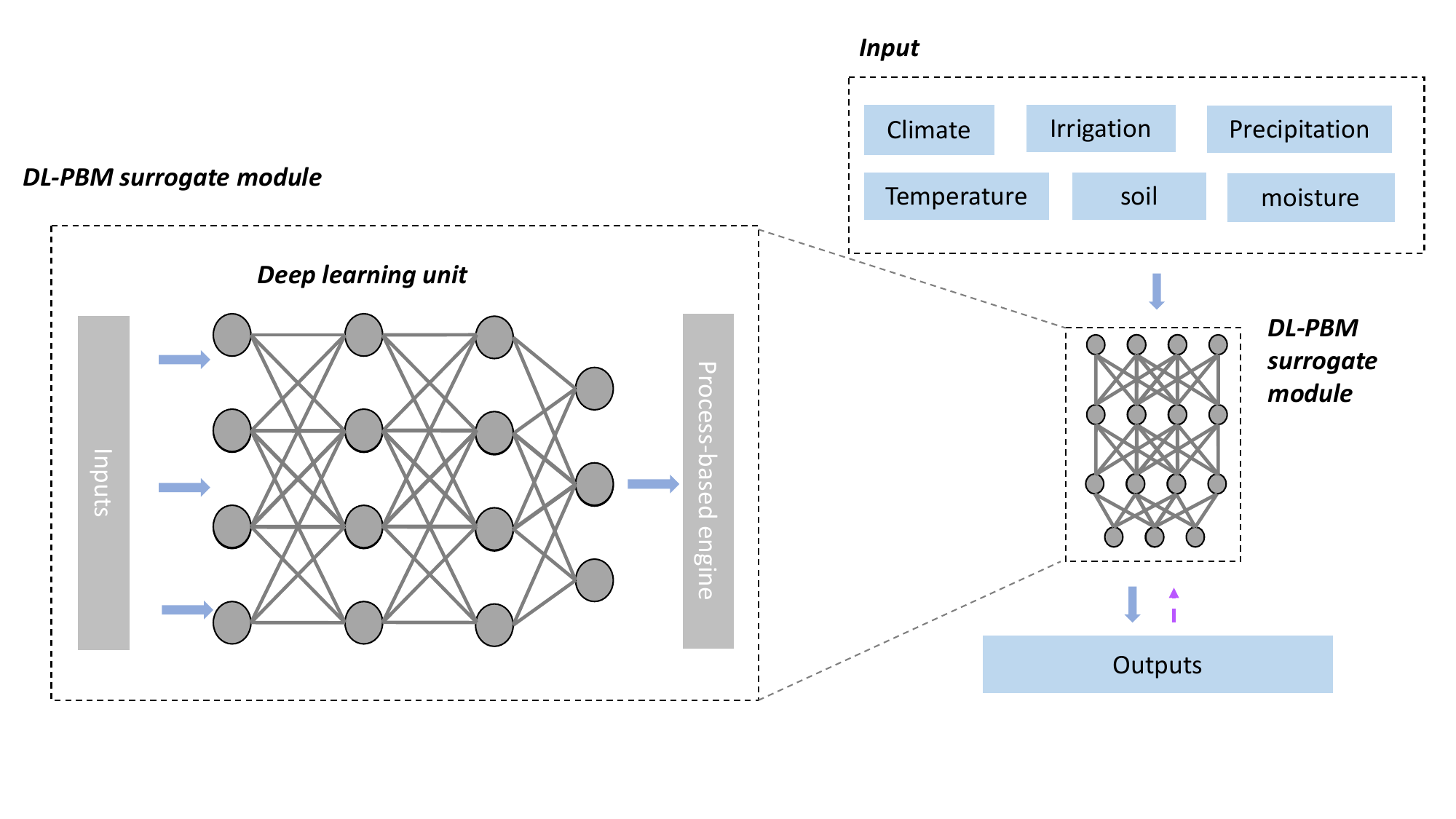}  
    \caption{The general workflow of connecting DL with PBMs through surrogate models.}  
    \label{fig:3b}  
\end{figure}

When a DL network is trained to emulate the input - output behavior of a PBM, the resulting DL-based system is referred to as a surrogate model. Rather than running the PBM directly, the surrogate model approximates its outputs for given inputs, thus greatly reducing computational costs when performing tasks like model inversion or parameter optimization. This strategy has been widely employed in various PBM settings, including hydrological, hydraulic, and reactive transport models. In 2021, Tsai \textit{et al.} \cite{tsai2021calibration} introduced a novel methodology termed differentiable parameter learning (dPL), leveraging the differentiable nature of such surrogate models for training purposes. They established a surrogate model of the Variable Infiltration Capacity (VIC) hydrological process model, which was integrated with an additional neural network (denoted as g) designed to estimate the physical parameters ($\theta$) of VIC based on accessible attributes (A): $\theta$ = g(A). This integration facilitates an 'end-to-end' workflow wherein $\theta$ is fed into the VIC model, and its outputs are then compared against empirical observations. This process effectively transforms the challenge of parameter calibration into a machine learning problem, which is concurrently trained across multiple sites through back-propagation and gradient descent. The workflow of connecting NNs with PBMs through surrogate models is shown in Fig. \ref{fig:3b} \par

While not constituting the core focus or philosophical underpinning of PBMs, surrogate models undeniably contribute to accelerating and enhancing the objectives of PBM by offering a viable means to streamline complex modeling processes. The exploration of NNs for the numerical resolution of the PBMs underscores this potential, with applications increasingly spanning various scientific disciplines. For instance, in Xue \textit{et al.} study \cite{xue2022ensemble}'s study, NNs have been deployed to approximate the solutions to PDEs like the Richards equation, which delineates water movement within soil matrices. Similarly, Innes \textit{et al.} \cite{innes2019differentiable} propose a differentiable surrogate models which find application in the inversion of bathymetry measurements for two-dimensional hydraulic simulations, illustrating the breadth of their utility. Finally, while surrogate models present a pragmatic entry point into the realm of PBMs for intricate and computationally intensive models, their utility is balanced by limitations in adaptability and the necessity for periodic retraining. \par

\textbf{2) Physics-informed models}\par
\begin{figure}[]   
    \centering  
    \includegraphics[width=5.5in]{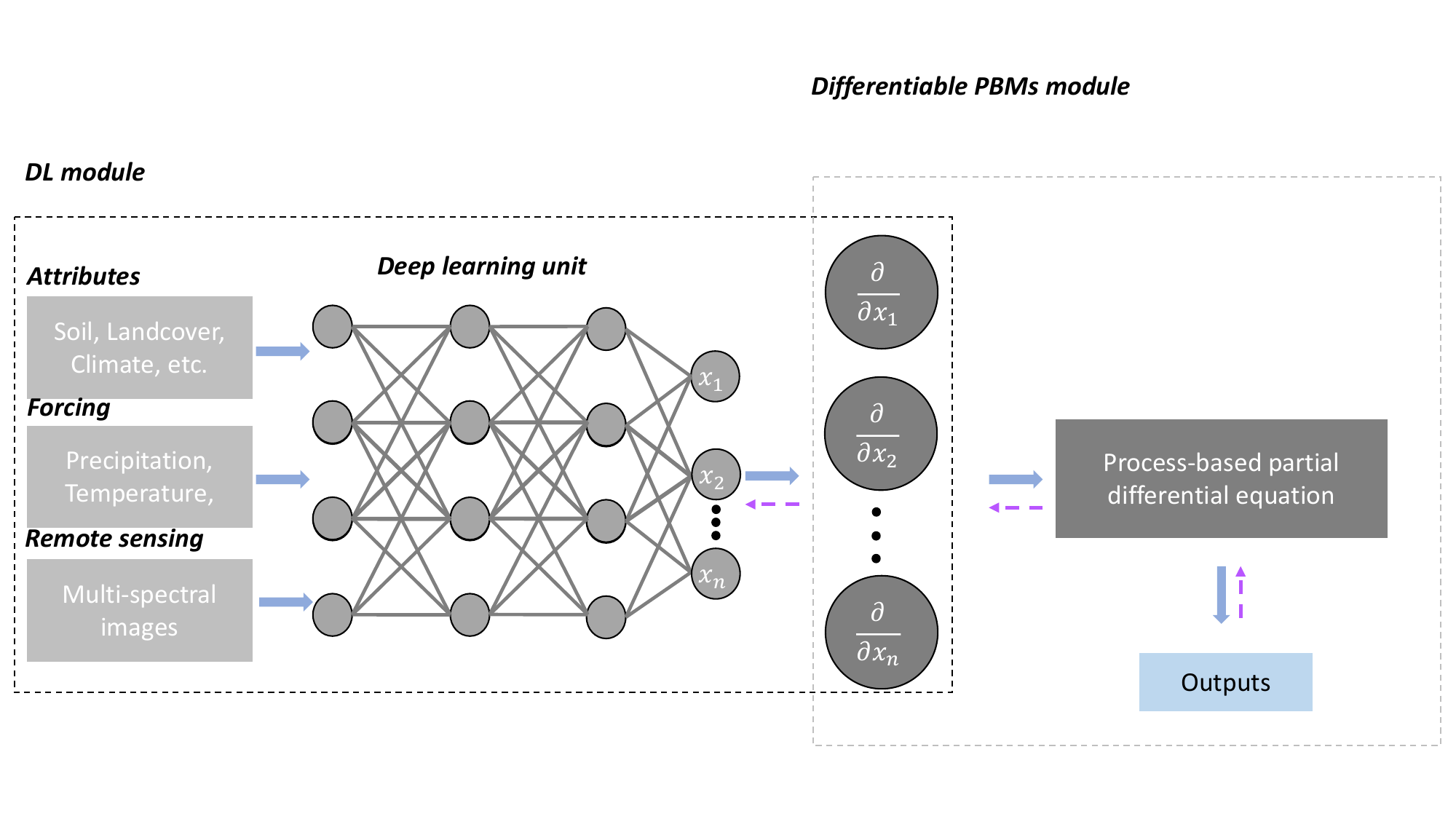}  
    \caption{The general workflow of physics-informed neural networks.}  
    \label{fig:3c}  
\end{figure}

Since their introduction in 2017 \cite{raissi2017physics}, Physics-Informed Neural Networks (PINNs) have been recognized as a distinct subset of DM, given their integral use of gradient information during the training process \cite{raissi2017physics}. PINNs adopt an innovative approach by training a neural network, denoted as $h(t,x)$, where t is time and x represents spatial coordinates such that $h(t,x)$ agrees with known data points at $(t,x)$. Crucially, the derivatives of h with respect to t and x, such as $dh/dx$ and $dh/dt$, are calculated via automatic differentiation and are constrained to adhere to the underlying PDEs governing the system \cite{shi2023physics}. Physical parameters may also be integrated as inputs into the network, enhancing the model's comprehensiveness. The workflow of physics-informed neural networks is shown in Fig. \ref{fig:3c} \par

PINNs have been applied across a spectrum of disciplines, evidencing their versatility and effectiveness in addressing complex problems. These applications include data assimilation \cite{zhao2023physics} and the derivation of governing equations from empirical data \cite{faroughi2024physics}. However, like any modeling approach, PINNs are not without their limitations. The formulation of $h(t,x)$ is inherently dependent on the specific initial and boundary conditions of the system being modeled, necessitating individual training sessions for each unique set of conditions. Moreover, the nature of the inputs confines the choice of neural network architecture to types such as multilayer perceptron networks, which may present challenges in training due to their complexity. \par

In the context of agricultural modeling, a novel application of the PINNs methodology was introduced for demonstrating unknown parameter fields and constitutive relationships. For example, Tartakovsky \textit{et al.} \cite{tartakovsky2020physics} applies PINNs to subsurface flow problems, which are relevant to agricultural systems (e.g., soil moisture dynamics). The authors show that PINNs can effectively learn unknown parameter fields (e.g., hydraulic conductivity) and constitutive relationships (e.g., soil-water retention curves) from sparse observational data, highlighting their potential for agricultural modeling. This approach was exemplified through the modeling of steady-state groundwater flow within an aquifer characterized by an indeterminate conductivity field, alongside unsaturated flow in the vadose zone with pressure-dependent conductivity that remained undefined. \par

\subsection{Agricultural applications of Hybrid PBM-DL modeling}
\label{sec:32}
Based on our proposed two categories in Section \ref{subsec:hybridarchitecture} (i.e. DL-informed PBM and PBM-informed DL), the hybrid PDM-DL models present a versatile approach for addressing a wide array of agricultural challenges. These model applications span from deducing physical parameters and revising structural model assumptions to predicting temporal forcing terms within natural systems. It's crucial to distinguish the hybrid PDM-DL model from prior iterations found in physics process that lack complete differentiability. This distinction underscores the unique methodology and application spectrum of the PDM-DL model, leveraging the synergistic integration of physics-based understanding and data-driven insights. For example, Ibrahim \textit{et al} \cite{ibrahim2022expert} presented a hybrid learning approach that combines mechanistic models of pest population dynamics with machine learning techniques to predict pest outbreaks. Singh \textit{et al} \cite{singh2024novel} proposed a hybrid model for crop yield prediction which capture both the underlying biological processes and the complex patterns in historical data, leading to improved prediction accuracy. \par

\begin{figure}[]   
    \centering  
    \includegraphics[width=3.5in]{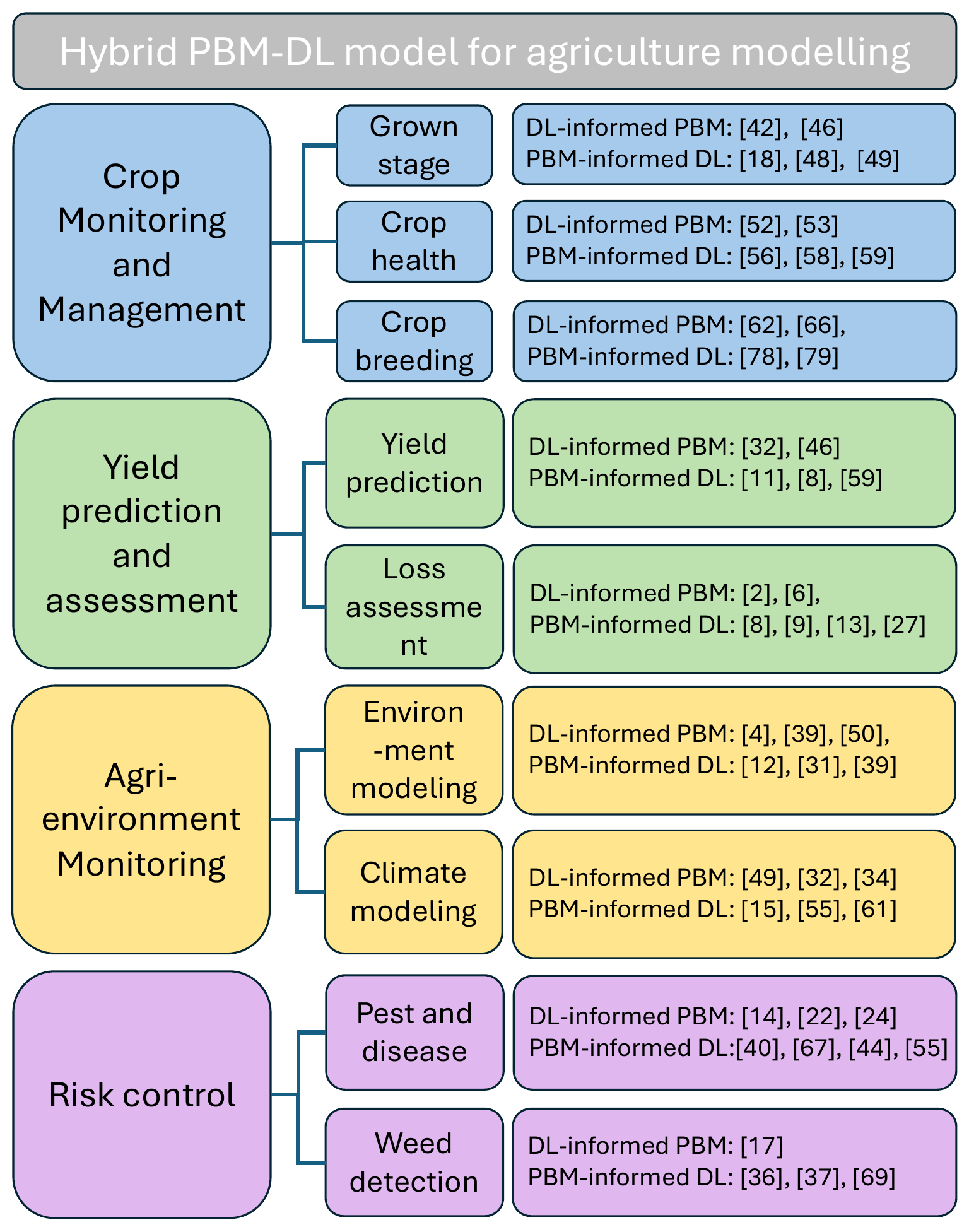}  
    \caption{The agricultural applications of Hybrid PBM-DL modeling.}  
    \label{fig:5}  
\end{figure}

\begin{figure}[]   
    \centering  
    \includegraphics[width=6.5in]{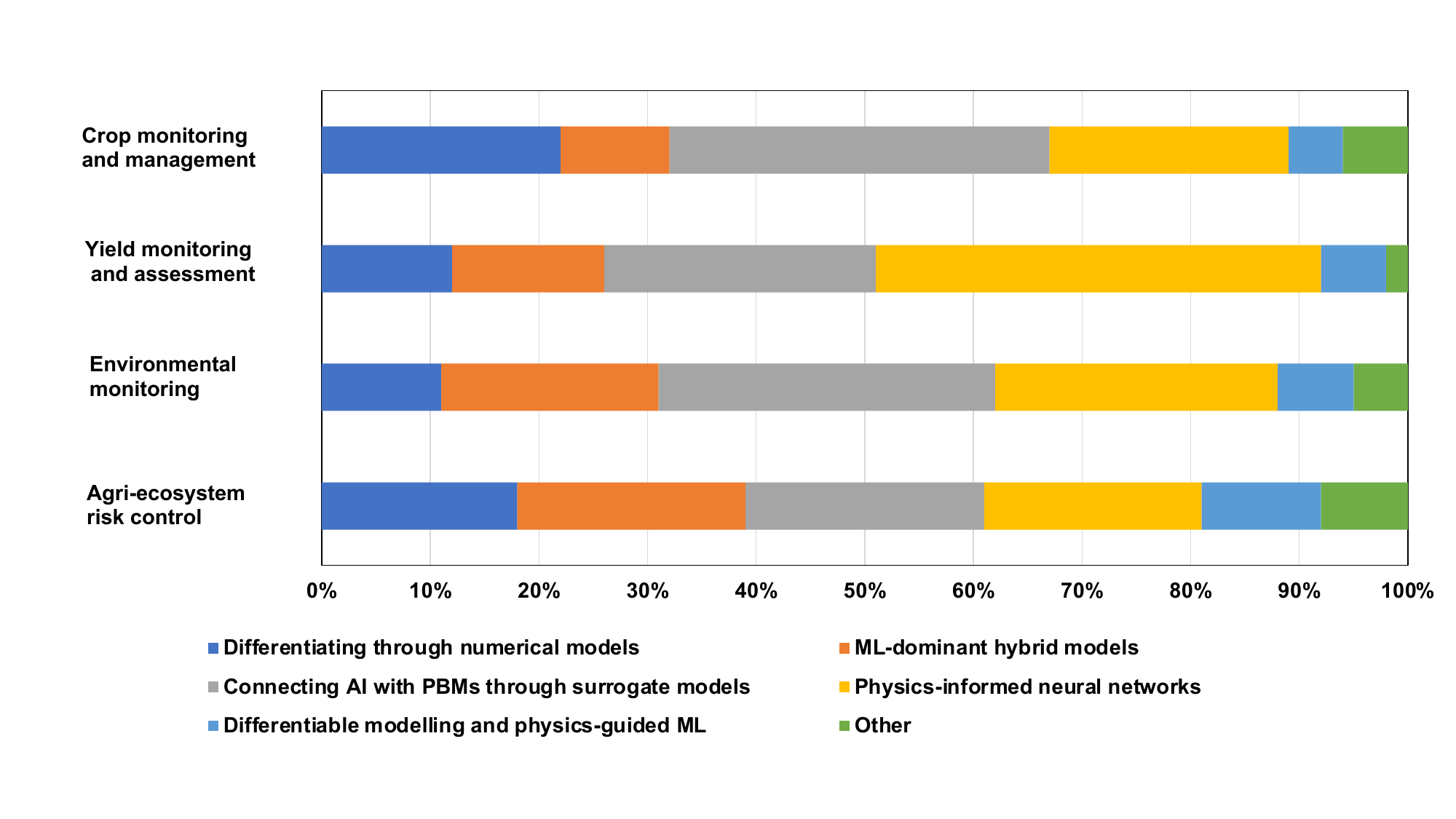}  
    \caption{The proportions of different hybrid methodologies used in each research category.}  
    \label{fig:12}  
\end{figure}

In this section, we will review and summarize existing related applications based on our proposed two categories of hybrid models, all of which involve the use of DL-informed PBM and PBM-informed DL modeling in addressing agricultural problems. To better organize our review, we categorize these papers according to their applications in agricultural and provide subdivisions for some common applications, as illustrated in Fig. \ref{fig:5}. We further explore the relationship between research categories and methodology, as shown in Fig. \ref{fig:12} It is important to note that some applications may overlap with each other, but our categorization attempts to align with the core problems addressed by each paper. \par

\subsubsection{Crop monitoring and management}

Hybrid modeling helps explain realistic crop conditions and support various crop monitoring and management applications. Both DL-informed PBM and PBM-informed DL methods can address wide-ranging tasks in crop monitoring and management, including:
\begin{itemize}
    \item \textbf{Crop growth stage monitoring}: tracking phenophases using hybrid data assimilation, 3-D imaging, or phenotypic trait extraction. 
    \item \textbf{Crop health management}: incorporating sensor-derived indicators (e.g., canopy temperature, pest damage signals) into PBM equations, or constraining DL models with physical insights on disease progression.
    \item \textbf{Stress level detection}: integrating RS-based vegetation indices into PBMs or imposing physical thresholds on DL outputs for abiotic stress characterization.
\end{itemize}

While DL-informed PBMs capitalize on neural networks to refine or directly drive the PBM’s inputs, PBM-informed DL focuses on weaving physical rules into the DL architecture. The detailed review for the existing application cases are discussed below:

\textbf{1) DL-informed PBM applications to crop monitoring and management} \par
Over the past decade, DL-informed PBM approaches have proven beneficial by integrating remote sensing (RS) data streams---an important technique for monitoring crop ecosystem dynamics in the presence of limited or noisy in-field observations. \par

Specifically, for crop growth stage monitoring, DL-informed modeling approach provides an opportunity of multi-source data assimilation into the PBM approach, for instance, the model proposed by Ma \textit{et al.} \cite{ma2013assimilation} merged RS products (e.g., leaf area index (LAI) or vegetation indices (VIs)) with the PBM. Works like \cite{nguyen2019mathematical, shawon2020assessment} demonstrated how incorporating RS can fill spatiotemporal gaps, while \cite{shawon2020two, shin2021simulation} presented RS-integrated crop models (RSCM) that rely on hybrid schemes for simulating staple crops (barley, paddy rice, soybean, wheat). In these instances, the neural network component processed RS features (such as multi-spectral or LiDAR data), refining or adjusting PBM parameters before the PBM predicts crop growth status. This synergy leverages both the physics-based structure of a PBM and the adaptiveness of DL to complex, high-dimensional inputs. \par

In addition, for crop health management scenarios, DL-informed modeling modules can help correct the PBM’s tendencies to over- or underestimate certain crop traits when direct measurements are sparse \cite{ojeda2017evaluation}. For stress detection characterization, a DL model can process 2-D or 3-D imagery of canopies or organs to extract critical parameters (e.g., biomass, leaf coverage), the PBM then takes these refined parameters as inputs, effectively aligning simulated growth curves to observed conditions in near-real time \cite{alabdrabalnabi2022machine}. By relying on the neural network for higher-level feature extraction from images, the DL-informed PBM architecture more accurately reproduces the sequence of events in plant growth, aiding agronomic decisions such as fertilization, harvest timing, and pest control strategies \cite{shawon2020two}. \par

\textbf{2) PBM-informed DL applications to crop monitoring and management} \par
In contrast, PBM-informed DL places the PBM’s physical principles and constraints at the core of the learning process. Rather than simply supplementing a PBM with neural networks, this strategy embeds physical laws, growth equations, or ecological constraints directly into the DL architecture, training procedure, or loss function. The result is a deep model that remains faithful to known agronomic dynamics, yet flexible enough to learn from data-rich remote or proximal sensing.\par

For the application to growth stage monitoring and phenology, a plant’s dynamic development or “growth stage” has long been central to critical agronomic decisions, from breed selection to fertilization and harvest scheduling. In PBM-informed DL approaches, well-established physiological or phenological concepts (e.g., the Biologische Bundesanstalt, Bundessortenamt and Chemical industry, or BBCH, growth scale \cite{koch2007guidelines}) guide the neural network so that it remains consistent with known plant developmental transitions. For example, a growth-stage detection network might incorporate PBM-based temperature or photoperiod functions as prior knowledge, ensuring that the predicted phenological stage aligns with physically realistic growth timelines \cite{zhao2019phenological, das2017automated}. Research has also explored quantitative phenotyping with measurements of leaves, stems, flowers, or fruits \cite{campillo2010study, chacon2013quantitative, cortes2017model}. In PBM-informed DL workflows, structural or physiological features derived from images are constrained by equations describing how plant organs expand, senesce, or accumulate biomass over time. Consequently, the model more effectively disentangles actual organ growth from artifacts such as lighting or perspective distortions.\par

For applications to crop health management, plant images often suffer from ambiguities in organ expansion or movement when taken under varying lighting conditions or sun angles. PBM-informed DL systems address this by enforcing physical consistency in the neural network outputs. For instance, they can incorporate mechanistic equations relating leaf elongation to daily thermal time accumulation. This is especially useful in 3-D imaging setups that employ LiDAR or multi-camera reconstructions to produce point clouds or volumetric data \cite{cortes2017model}. By combining physical growth laws with advanced DL-based feature extraction, PBM-informed DL methods generate precise estimates of geometric traits, even when dealing with partial occlusions or variable imaging geometries.\par

\subsubsection{Yield monitoring and assessment}
\label{sec:yield_monitoring_assessment}

Yield prediction directly supports food security, providing timely information on food availability, enabling farmers to plan harvests, make informed decisions, and negotiate prices \cite{abdel2024proposed, singh2024novel}. Recent approaches have linked crop yields to remote sensing indices, weather, soil, and management factors but can face challenges in limited data availability and domain shifts. Hybrid PBM-DL solutions aim to overcome these challenges by leveraging the physical constraints of PBMs and the representation power of DL.

\textbf{1) DL-informed PBM in yield prediction} \par
In a DL-informed PBM framework, the PBM’s simulation of crop yield (e.g., through mechanistic growth equations) is refined or informed by DL outputs \cite{muruganantham2022systematic}. This configuration is especially useful when the PBM lacks key parameter estimates or when data are noisy. By integrating a trained neural network to provide dynamic corrections or parameter updates, the PBM can retain its physically interpretable structure while benefiting from the DL model’s ability to capture complex, non-linear interactions. For instance, BI \textit{et al.} \cite{bi2023transformer} proposed a transformer based neural network to estimate site-specific yield sensitivity factors based on historical yield data and local management inputs, then feed these refined factors into the PBM’s yield simulation routine.

\textbf{2) PBM-informed DL and transfer learning} \par
By contrast, in a PBM-informed DL strategy, the DL architecture or training procedure is guided by the PBM’s mechanistic principles. Wang \textit{et al.} \cite{wang2021transfer} employed feature-based TL with an LSTM model initially trained on Argentina’s soybean data, adapting it to Brazil’s conditions to improve prediction accuracy without extensive local data. Similarly, a PBM--DL hybrid can encode known plant physiology, enabling the model to map learned growth patterns from one region to another by aligning them with the target’s environmental attributes. In scenarios with very limited in-field yield data, domain adaptation methods, such as unsupervised domain adaptation, can be incorporated. Ma \textit{et al.} \cite{ma2021adaptive} proposed an adaptive DANN model (ADANN) for corn yield prediction, dynamically balancing prediction loss and domain loss to estimate yields under minimal labeled data. Integrating Bayesian inference further improved robustness in Bayesian adversarial domain adaptation (BDANN) \cite{wang2023bp}. \par

PBM-Informed DL methods can also benefit from multi-task learning (MTL), where multiple crops or temporal forecasts are learned simultaneously. Khaki \textit{et al.} \cite{khaki2021simultaneous} applied MTL across corn and soybean yields using MODIS spectral data, achieving accurate predictions even months before harvest. Similarly, YieldNet \cite{yildirim2022using} used a CNN-based architecture to share core features across tasks, reducing training time and improving generalization. By embedding PBM-based constraints, these hybrid models can better address domain shift during TL, such as via multi-source maximum predictor discrepancy (MMPD), selectively aligning predictions with the most relevant source domain \cite{ma2023multisource}. \par

\subsubsection{Environmental monitoring}
\label{sec:environmental_monitoring}

Agriculture and ecosystem health are critical indicators of environmental stability, yet traditional monitoring methods (e.g., field surveys, camera trapping, visual inspections) often involve high costs, time commitments, and limited spatial coverage \cite{wang2020detecting}. By merging domain knowledge with DL techniques, hybrid PBM--DL approaches provide systematic ways to address these complexities in applications such as species composition estimation, deforestation detection, and invasive species mapping. \par

\textbf{1) DL-informed PBM for ecological assessments} \par
In environmental contexts, satellite observations or UAV imagery can yield high-dimensional data (e.g., hyperspectral signatures) that inform PBMs through domain of DL. Alohali \textit{et al.} \cite{alohali2023anomaly}, for instance, employed feature-based transfer learning with hyperspectral imagery to estimate phytoplankton species composition. Although their focus was primarily on DL adaptation and domain alignment, this methodology can readily be integrated with PBMs that describe phytoplankton population dynamics, enabling improved parameter estimation and forecast accuracy under complex ocean conditions. Similarly, generative models like CycleGAN can reduce domain discrepancies across multiple spatial or environmental conditions \cite{vega2024convolutional}, aligning remote-sensing observations with the PBM’s input space for robust deforestation monitoring or habitat assessment.

\textbf{2) PBM-informed DL for invasive species and ecosystem health} \par
Alternatively, a PBM-informed DL approach embeds ecological or agricultural principles directly into the neural network’s loss function or architecture, thereby constraining the DL system to respect known environmental processes. This is valuable for invasive species detection, where labeled data are scarce or inconsistent across regions. Chaudhuri and Mishra \cite{chaudhuri2023detection} fine-tuned a pre-trained ResNet-152 model with UAV imagery to detect purple loosestrife along the Mississippi River; a PBM-informed technique could incorporate population dynamics or dispersal constraints for more consistent predictions under varying hydrological conditions. Tufail \textit{et al.} \cite{tufail2024classification} similarly utilized an Inception-ResNet model to identify \emph{Ziziphus lotus} in aerial images, achieving high accuracy suitable for large-scale monitoring. By integrating PBM-derived biological constraints, PBM-informed DL can prevent physically implausible outputs (e.g., unrealistically rapid spread rates) and guide model predictions in data-limited scenarios. \par

PBM-Informed DL systems also extend to generative unsupervised domain adaptation (UDA) methods, where models like CycleGAN synthesize training images under diverse environmental conditions to improve model robustness. For instance, Zhao \textit{et al.} \cite{zhao2024satellite} demonstrated how synthesizing field images spanning various seasons, lighting conditions, and weather events enhances real-world performance in crop monitoring. In a PBM-informed DL context, these generated images could be constrained by ecological principles (e.g., phenological milestones), ensuring that the synthetic samples reflect plausible environmental states. Consequently, the model becomes more adaptable to unforeseen conditions, improving both detection accuracy and resilience in operational monitoring tasks. \par

\subsubsection{agricultural risk control}
\label{sec:agri_risk_control}

Hybrid PBM--DL models offer a powerful means to describe and predict plant stress over large spatial domains, leveraging complementary strengths of PBMs and DL. Recent advances in Internet of Things (IoT) and airborne platforms allow for more comprehensive agro-ecological modeling \cite{weiss2020remote}, while integrative methodologies are still being developed to capture plant stress indicators. Although early hybrid PBM--DL efforts relied on parametric regressions relating IoT measurements to plant functional traits \cite{okyere2024hyperspectral}, more sophisticated schemes now incorporate radiative transfer, photosynthesis, and energy-balance models to better represent cause-and-effect in stress conditions \cite{wolanin2019estimating, verhoef2018hyperspectral}. \par

\textbf{1) DL-Informed PBM for stress prediction} \par
In DL-informed PBM approaches, the PBM serves as the foundational engine for simulating plant physiology under stress (e.g., water deficit or pathogen infection), while a neural network refines or supplies additional parameters critical for accurate modeling. For instance, Sishodia \textit{et al.} \cite{sishodia2020applications} highlighted how advanced physical models should complement DL methods to reduce the complexity of PBMs yet maintain computational efficiency. A relevant example is found in multi-sensor data assimilation setups, where hyperspectral or thermal imagery captures real-time plant stress indicators (e.g., canopy temperature). The DL module can then process these high-dimensional data streams, providing refined inputs (e.g., updated leaf biochemical traits) to the PBM, which performs stress simulation. This workflow taps into the interpretability of PBMs while harnessing the data-driven capabilities of DL to handle noise and multi-sensor complexities.

\textbf{2) PBM-informed DL for early disease detection} \par
PBM-informed DL weaves physical or physiological constraints directly into the deep learning pipeline, ensuring stress predictions remain grounded in known agro-ecological processes. Zarco-Tejada \textit{et al.} \cite{zarco2018previsual} exemplify this in studying Xylella fastidiosa-induced stress in trees, using a three-dimensional radiative transfer model (RTM) to estimate fluorescence efficiency and derive thermal stress indicators. These physically based outputs were then combined with CNN-based analyses of narrow-band spectral indices (e.g., chlorophyll, carotenoids, and xanthophylls) to classify early-stage disease incidence and severity at landscape scales. Here, the RTM (a type of PBM) actively constrains and shapes the DL task by translating raw spectral signatures into biologically meaningful features. Likewise, Hernández-Cumplido \textit{et al.} \cite{hernandez2019early} illustrate how incorporating physiological indicators, such as canopy temperature and photosynthesis rates, can guide DL algorithms to identify stress onset more reliably. These examples underscore the promise of PBM-informed DL models in improving agricultural risk control. By synthesizing the detailed mechanistic underpinnings of PBMs with the predictive power and flexibility of DL, researchers and practitioners can detect and manage plant stress more effectively across extensive farmland. Whether through DL-informed approaches that augment PBMs with rich observational data or PBM-informed methods that embed physical principles into DL architectures, hybrid frameworks can adapt to evolving environmental conditions and scale to region-wide monitoring. This enables timely interventions and more resilient agro-ecosystems in the face of climate variability, disease outbreaks, and other agronomic threats.

\section{Evaluation of PBM, DL and Hybrid models: A Case Study}
\label{sec:4}

This section presents an illustrative case study to demonstrate how hybrid models integrate the strengths of process-based models (PBMs) and deep learning (DL). Rather than conducting an exhaustive comparison, this example provides a simplified representation of how these models function and interact, offering insights into their fundamental processes and applications.

Hybrid modeling approaches aim to combine the interpretability of physics-driven methods with the adaptability of data-driven learning, resulting in more robust and accurate predictions than standalone PBM or DL models. To illustrate this, we examine their performance under three key challenges:

\begin{itemize}
\item Handling noisy data, which is common in real-world agricultural and environmental applications.
\item Learning from limited datasets, as collecting large, labeled agricultural datasets can be expensive or impractical.
\item Generalizing across different spatial locations, since models are often required to perform well in new, unseen environments.
\end{itemize}

To showcase the fundamental workflow of hybrid modeling, we select a single representative model from each category and apply them to a crop dry biomass prediction task, a crucial problem in agricultural modeling.

\begin{itemize}
\item PBM Model: We use SIMPLACE, a well-established agricultural system model, to compute key crop indices such as radiation, leaf area index (LAI), and photosynthetically active radiation (PAR) using mechanistic equations.
\item DL Model: We employ a Long Short-Term Memory (LSTM) network, a deep learning architecture designed to capture temporal dependencies in agricultural data.
\item DL-Informed PBM: In this approach, LSTM is integrated with SIMPLACE, following the principle of differentiating through numerical models (Section \ref{subsub:DL-Informed}). Here, LSTM helps refine uncertain PBM parameters, improving prediction accuracy.
\item PBM-Informed DL: This method incorporates SIMPLACE’s physics-based constraints into the LSTM training process, following the Physics-Informed Neural Networks (PINNs) framework (Section \ref{subsub:PBMS-InformedDL}). This ensures that DL predictions adhere to established agricultural principles and remain physically meaningful.
\end{itemize}
This case study serves as a concise demonstration of how hybrid PBM-DL models operate, illustrating their ability to leverage mechanistic knowledge while benefiting from data-driven adaptation. The primary goal is to provide a clear, high-level understanding of the workflow and interaction between PBMs and DL models, offering insights into their potential advantages in agricultural applications.

\subsection{Data collection}
\begin{figure}[]   
    \centering  
    \includegraphics[width=5.5in]{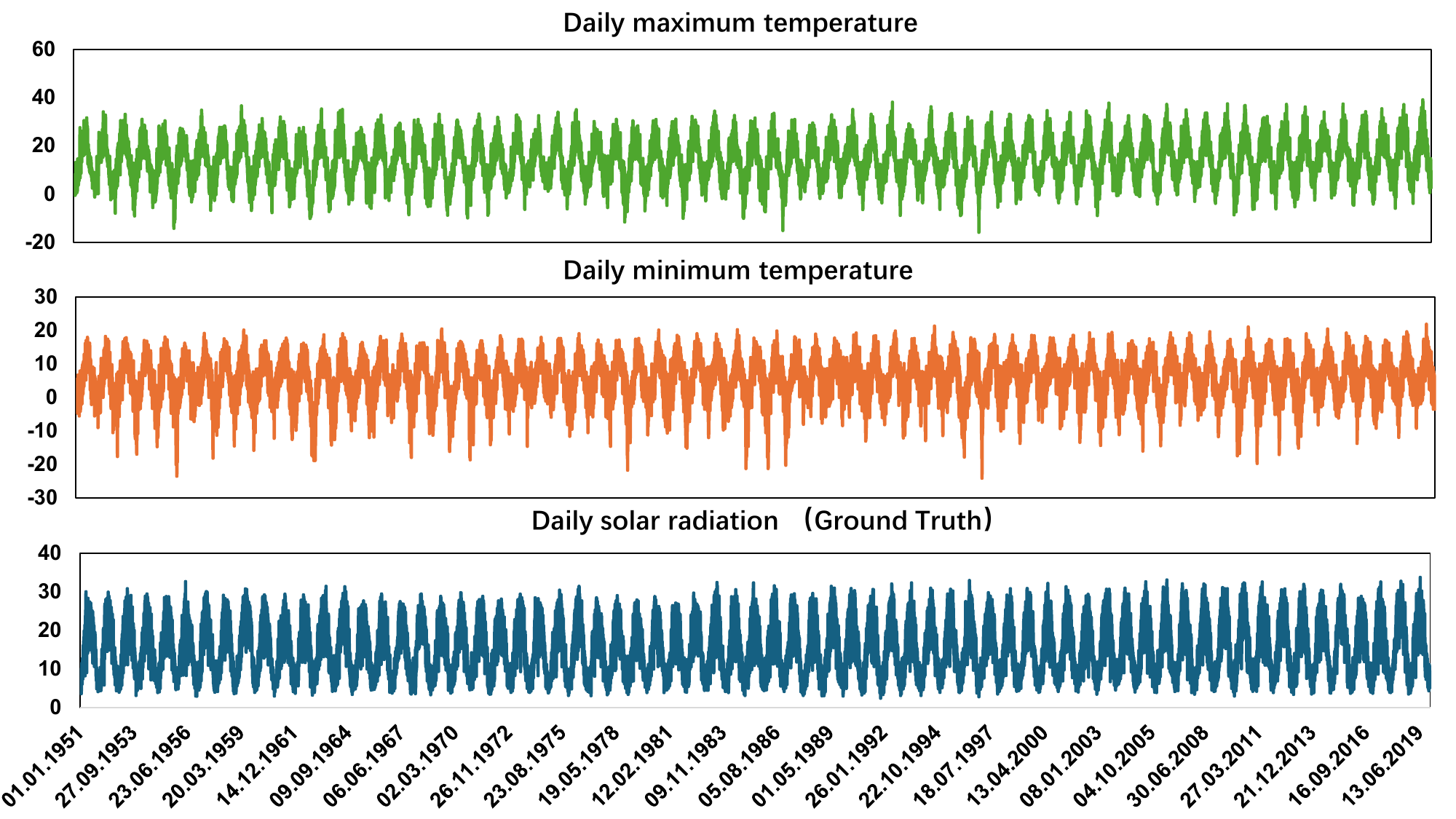}  
    \caption{The data in average used for the comparative modeling experiment with PBM, DL, and hybrid models}
    \label{fig:6}  
\end{figure}

The case studies aim to model the crop dry biomass in three observation sites in Germany. Each experimental site covers an area of 75 $km^2$ (25 $km^2$ for each). Fig. \ref{fig:6} shows an example daily temperature data cross 68-year. The first 48 years (1951-1999) data are used as the training dataset, and the last 20 years (1999 - 2019) data are used for model evaluation on crop dry biomass estimation.  \par

% \begin{figure}[]   
%     \centering  
%     \includegraphics[width=5.5in]{9_yield_GT.pdf}  
%     \caption{The ground truth biomass in average used for model evaluation}
%     \label{fig:6_2}  
% \end{figure} 

\subsection{Experimental Design and Goals}
The study includes three main experimental setups to compare PBM, DL, and hybrid models: model performance under noisy data, learning from limited datasets, and generalizing across different spatial locations.\par

\textbf{1) Model performance under noisy data}\par
For the model performance under noisy data evaluation, the first 48 years (1951-1999) of data contaminated by three Gaussian noise levels were designated for model calibration (i.e., the 'seen' data during model development), while the last 20 years (1999-2019) were reserved for model validation (i.e., the 'unseen' data). For the training process of DL model and hybrid model, the calibration period was further divided into training (randomly selected 40 years) and testing (rest 10 years) periods, the latter used for early stopping to prevent overfitting. For the evaluation of few-shot learning performance, we trained the DL and hybrid models in three training scales: randomly selected 7 years (about $10\%$), 3 years (about $5\%$), and 1 year (about $1\%$) of data. For the evaluation of the model generality, the pre-trained models were used for different areas. \par

\textbf{2) Learning from limited datasets}\par
In data-scarce conditions, we randomly select 7 years ($10\%$), 3 years ($5\%$), or 1 year ($1.4\%$) of the entire calibration dataset to train the models. This setup evaluates how effectively each approach generalizes with limited samples. \par

\textbf{3) Generalizing across different spatial locations}\par
We test each model’s ability to transfer knowledge from two observation sites to an unseen location. A cross-validation approach is adopted: two sites serve as the training/testing dataset, while the remaining site is used solely for validation. This step assesses whether each model can accurately predict outcomes in new spatial contexts.\par

\subsection{Results}

\subsubsection{Evaluation of model performance under noisy data}

The standard approach to evaluating a model’s performance in estimating crop dry biomass is to test its ability to reproduce historical biomass records that were not used during model calibration. However, this type of validation is inherently partial, as it only partially captures how well the model generalizes to new conditions (unseen environmental conditions or spatial locations). In this study, we frame the data-noise calibration challenge as an optimization task aimed at denoising contaminated crop biomass data by adjusting model parameters. Specifically, we treat the original station-based biomass observations as noise-free “ground truth” and introduce three levels of Gaussian noise (i.e. 1$\times$, 2$\times$, 3$\times$) to simulate real-world contamination. We then calibrate the DL and hybrid models using backpropagation (BP) with an early-stopping strategy. During each training epoch, the models update their parameters using the training subset of data, while the testing subset helps detect potential overfitting.\par

\begin{figure}[]   
    \centering  
    \includegraphics[width=5.5in]{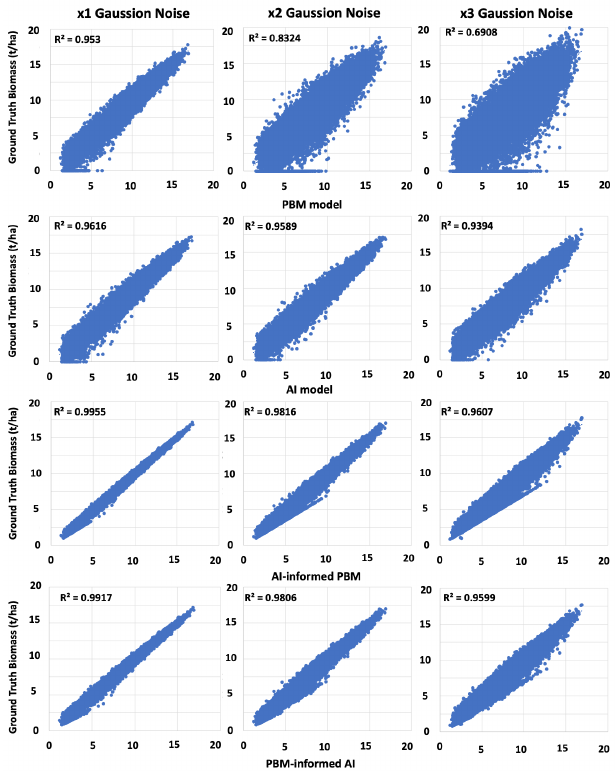}  
    \caption{The correlation comparison of the ground truth dry biomass and biomass predicted by PBM, DL, and hybrid models (DL-informed and PBM-informed) from the 1x, 2x, and 3x Gaussian contaminated weather inputs.}  
    \label{fig:7}  
\end{figure}

\begin{figure}[]   
    \centering  
    \includegraphics[width=5.5in]{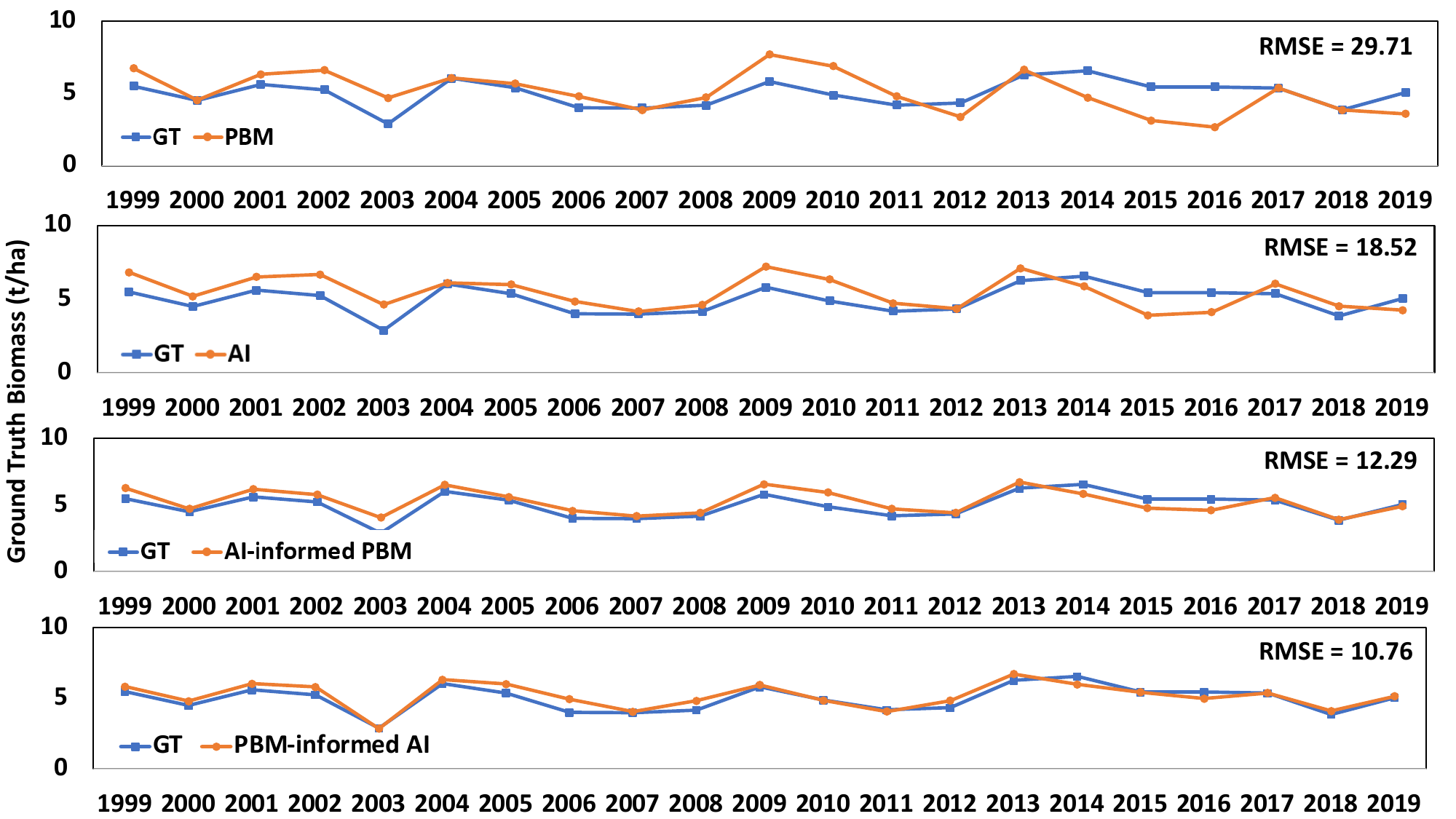}  
    \caption{The model evaluation by PBM, DL, and hybrid models (DL-informed and PBM-informed) from the 3x Gaussian contaminated weather inputs.}  
    \label{fig:7_2}  
\end{figure}

Figure \ref{fig:7} illustrates the comparison of the PBM, DL model, and hybrid models (DL-informed and PBM-informed) in predicting crop dry biomass from contaminated data, benchmarked against ground truth data. As anticipated, the hybrid models demonstrated enhanced robustness in predictions derived from noisy data, achieving higher $R^2$ scores that outperformed both the vanilla PBM and DL models. Furthermore, the DL-informed hybrid model achieved better results, with higher $R^2$ scores, compared to the PBM-informed hybrid model.  As shown in Fig. \ref{fig:7_2}, the hybrid model yields better dry biomass predictions, as evidenced by lower RMSE values. These results highlight the superiority of hybrid models over traditional PBM and DL models in handling contaminated biomass data, leading to improved goodness-of-fit. \par
%As the results, the dry biomass prediction, shown in Fig. \ref{fig:7_2}, illustrates the hybrid model achieves better results in terms of RMSE.\par. The superior performance of the DL-informed hybrid model can be attributed to the multi-start Newton-type optimization algorithm used for its calibration. This allows the DL-informed PBM model to better fit the ground truth distribution by denoising contaminated data, leading to more accurate predictions. These results underscore the significant superiority of hybrid models over traditional PBM and DL models in terms of goodness-of-fit for contaminated streamflow predictions. \par

\subsubsection{Evaluation on learning from limited datasets}

Learning from limited datasets is formulated as an optimization task aimed at effectively learning from a limited number of training examples.  To simulate learning scenarios with limited training data, we use weather data subsets comprising $10\%$, $5\%$, and $1.4\%$ of the full dataset. During each epoch, the training data are used to update the network parameters, while the validation data help identify potential overfitting.

%To simulate learning scenarios from limited training dataset, we use subsets of the weather data comprising $10\%$, $5\%$, and $1.4\%$ of the full dataset. 
\begin{figure}[]   
    \centering  
    \includegraphics[width=5.5in]{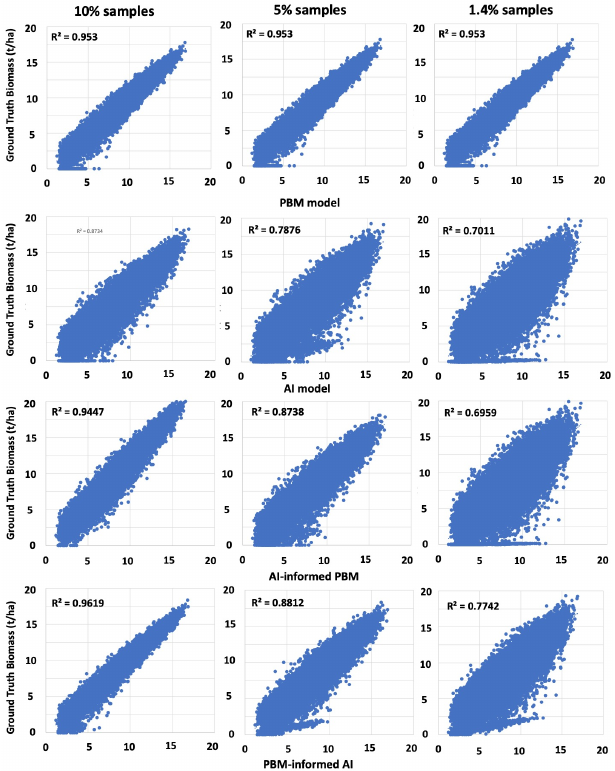}  
    \caption{The correlation comparison of the ground truth and predicted dry biomass by PBM, DL, and hybrid models (DL-informed and PBM-informed) with $10\%$, $5\%$, and $1.4\%$ subset of the raw dataset.}  
    \label{fig:8}  
\end{figure}

\begin{figure}[]   
    \centering  
    \includegraphics[width=5.5in]{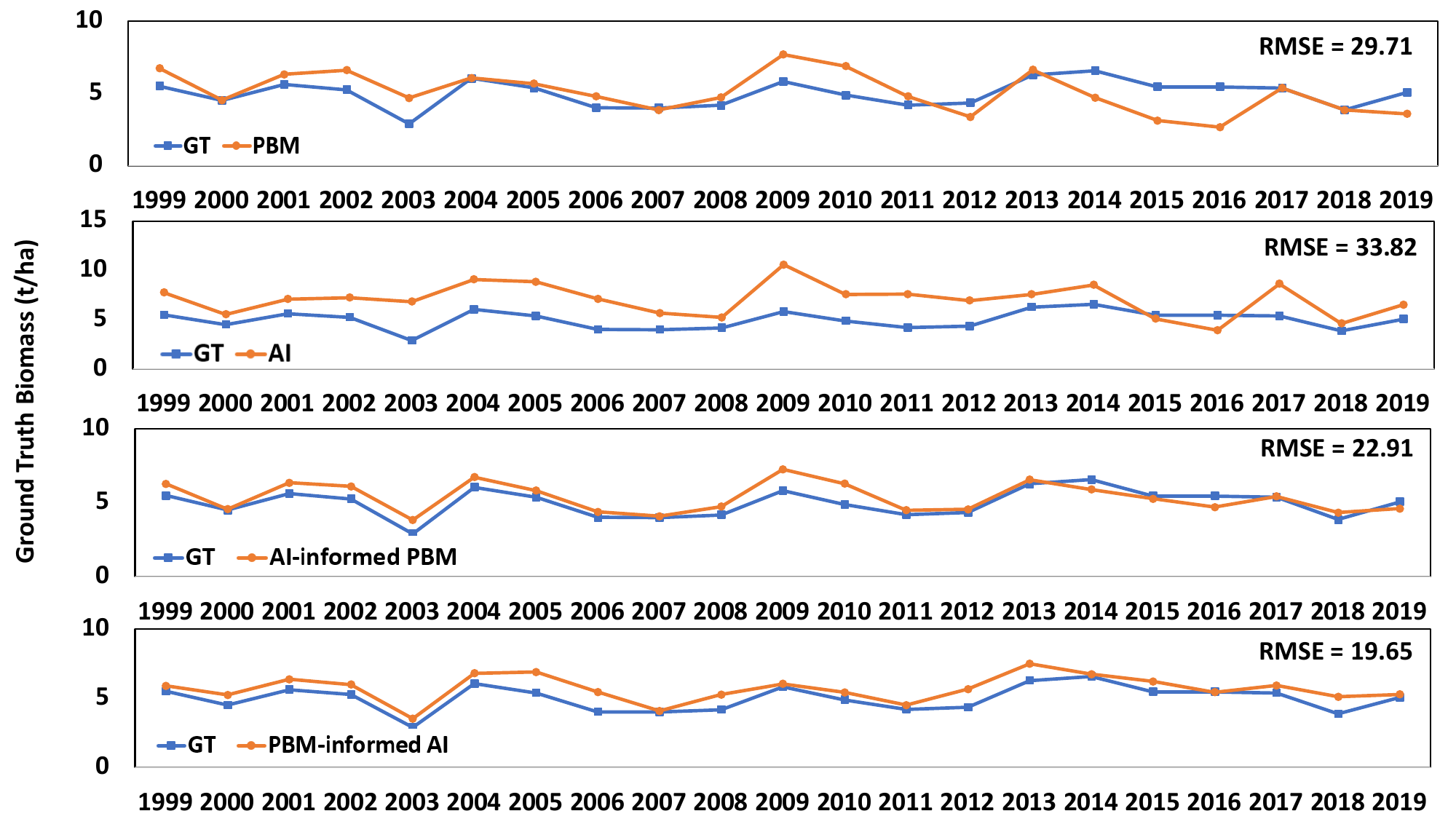}  
    \caption{The model evaluation on dry biomass prediction by PBM, DL, and hybrid models (DL-informed and PBM-informed) from $5\%$ subset of the raw dataset.}  
    \label{fig:8_2}  
\end{figure}

Fig \ref{fig:8} illustrates the comparison of the baseline PBM, DL model, and hybrid models (DL-informed and PBM-informed). As expected, the hybrid models demonstrated robust predictions derived from scarce data, achieving accuracy scores comparable to the PBM model and outperforming the vanilla DL model. Furthermore, The PBM-informed hybrid model achieved higher accuracy scores than the DL-informed hybrid model. The yield prediction from the few-shot learning scenarios are shown in Fig. \ref{fig:8_2}, illustrating that the hybrid model achieves better prediction in terms of RMSE. This suggests that the PBM-informed hybrid model better fits the ground truth data distribution and effectively leverages the limited examples for more realistic predictions.
%Suggesting that the PBM-informed hybrid model to better fit the ground truth data distribution from the limited data, effectively leveraging the limited examples for more realistic predictions. These results underscore the significant superiority of hybrid models over traditional DL models in terms of accuracy and generalization in few-shot learning tasks.

\subsubsection{Evaluation on model generalizing across different spatial locations}

\begin{figure}[]   
    \centering  
    \includegraphics[width=5.5in]{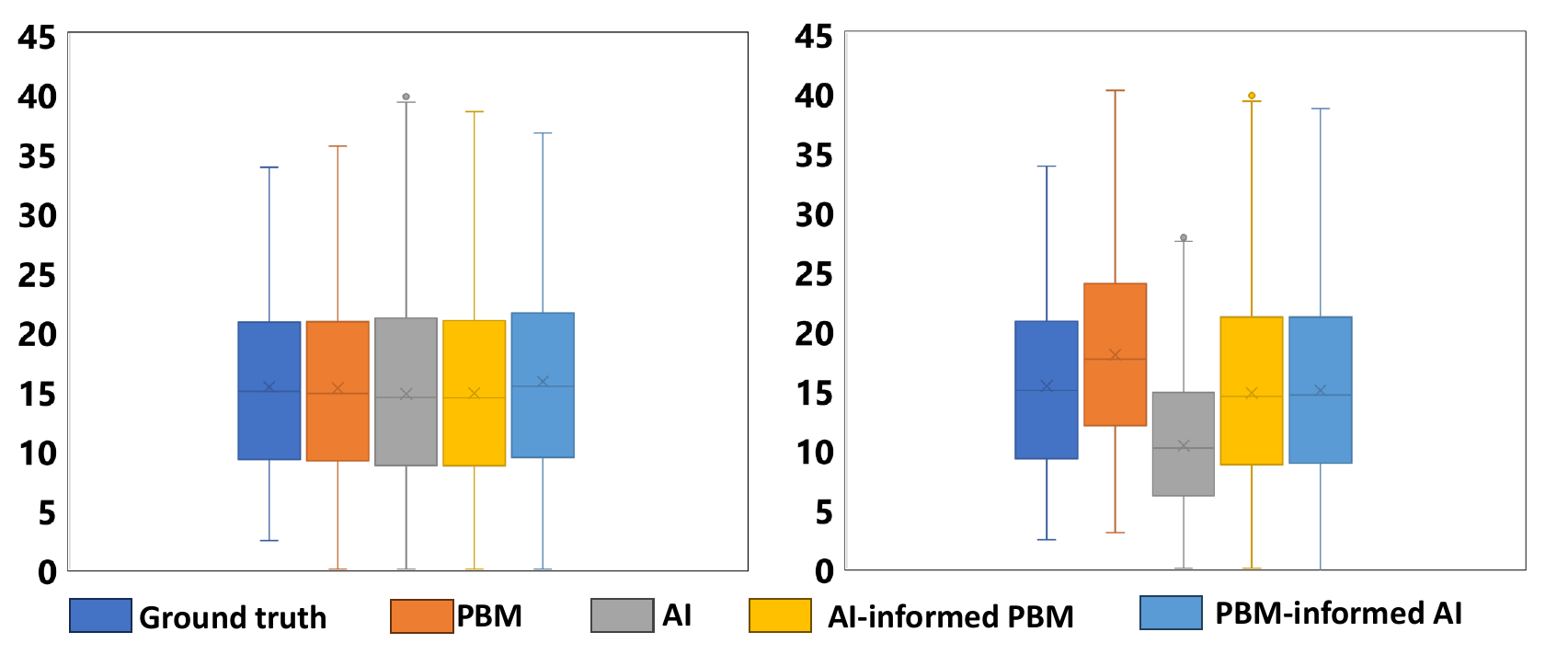}  
    \caption{A box plot that compares the yield prediction results on training/testing dataset (left) and the evaluation dataset (right).}  
    \label{fig:9}  
\end{figure}

Model performance in spatial generalization refers to a model's ability to apply patterns or relationships learned from certain locations or regions to predict outcomes or perform tasks in new, unseen locations. This assessment evaluates how well the model generalizes across different spatial contexts, rather than being limited to the specific locations on which it was trained. To rigorously evaluate the spatial generalization capabilities of PBM, DL, and hybrid models, we employ a cross-validation approach. This method randomly selects two observation sites for the training/testing dataset and uses the remaining site as the validation dataset. Figure \ref{fig:9} presents the results of dry biomass prediction on both the training/testing and validation datasets. Our results indicate that the hybrid models consistently produce accurate predictions in validation locations. 

\subsection{Experimental Summary} 
\label{sec:conclusion}

This case study demonstrates the effectiveness of hybrid modeling approaches that integrate PBMs with DL techniques for agricultural biomass prediction tasks. Hybrid models consistently outperformed standalone PBM and DL approaches across three key challenges: coping with noisy data, learning from limited datasets, and generalizing to unseen spatial locations. In data-scarce scenarios, hybrid models again outperformed the DL baseline, with the PBM-informed hybrid model demonstrating particularly strong performance, likely due to its ability to encode physical constraints that compensate for the lack of training data. Finally, in the context of spatial generalization, both hybrid approaches maintained reliable performance when applied to unseen locations, highlighting their adaptability and transferability in real-world agricultural modeling tasks. \par

Overall, these findings illustrate the value of hybrid PBM-DL models as a promising paradigm for agricultural modeling, offering a balance between mechanistic interpretability and data-driven flexibility. The hybrid framework not only enhances predictive performance but also enables broader applicability under challenging modeling conditions, making it a compelling direction for future research and deployment in precision agriculture.

\section{Conclusions and Future Challenges} 
\label{sec:conclusions}
This study presented a systematic review of process-based models (PBMs), deep learning (DL) methods, and hybrid PBM-DL frameworks in agri-ecosystem modeling. We summarized the strengths and limitations of each approach, emphasizing the interpretability and biophysical grounding of PBMs, the flexibility and pattern-recognition capacity of DL, and the emerging promise of hybrid models that integrate the two.
We categorized hybrid approaches into two main groups: DL-informed PBMs, where neural networks enhance mechanistic components, and PBM-informed DL, where physical constraints guide deep learning predictions. Through a representative case study on crop dry biomass prediction, we demonstrated that hybrid models consistently outperformed standalone PBMs and DL models—especially in scenarios with noisy data, limited training samples, and spatially unseen conditions. These results underscore the potential of hybrid models to deliver both robustness and generalization under real-world agricultural constraints.

\subsection{Recommendations for Advancing Hybrid Modeling in Agriculture}
To further advance hybrid PBM-DL integration, we recommend the following priorities:
\begin{itemize} 
\item \textit{Enhance Model Transparency}: Incorporate interpretable DL components (e.g., attention mechanisms) and diagnostic tools (e.g., ablation or sensitivity analyses) to clarify how physical priors influence predictions.
\item \textit{Invest in HPC and Algorithmic Efficiency}: Develop memory-efficient training routines and solvers capable of handling embedded differential equations, long time series, and large-scale simulation tasks.
\item \textit{Standardize Data Protocols}: Establish open, well-documented datasets and benchmarking frameworks across diverse crop types and regions to ensure reproducibility and model comparability.
\item \textit{Expand Uncertainty Quantification}: Use advanced probabilistic and Bayesian methods to account for uncertainty, improving confidence in decision-making under uncertainty.
\item \textit{Foster Cross-Disciplinary Collaboration}: Engage agronomists, ecologists, machine learning researchers, and software engineers to co-develop hybrid frameworks rooted in both scientific rigor and practical utility.
\end{itemize}

\subsection{Open Challenges and Research Opportunities}
Despite promising results, several challenges remain:
\begin{itemize} 
\item \textit{Balancing Interpretability and Complexity}: While PBMs offer domain transparency, integrating them with DL models can introduce abstraction layers that obscure model behavior. Future research should prioritize robust validation protocols and explainability methods to strike a balance between transparency and performance.
\item \textit{Scaling Computationally}: Hybrid models that incorporate ordinary or partial differential equations are computationally intensive, particularly on GPUs where memory limits can pose bottlenecks. Advances in high-performance computing, efficient auto-differentiation, and modular model architectures will be key to scaling these models to broader agricultural applications.
\item \textit{Integrating Heterogeneous and Sparse Data}: Combining diverse data sources—such as field measurements, remote sensing, and sensor networks—requires standardized data protocols and effective data fusion strategies. Collaboration across disciplines will be essential to improve data quality, availability, and integration.
\end{itemize}

Overall, our findings reinforce the potential of hybrid PBM-DL models as a next-generation modeling paradigm for sustainable agriculture. By uniting mechanistic insights with data-driven learning, these models provide interpretable, scalable, and robust tools for understanding and managing complex agricultural systems. Future research should continue to enhance methodological efficiency, data integration, and uncertainty quantification—advancing both scientific understanding and real-world impact in agri-environmental decision-making.

\section*{Acknowledgments}
This research is supported by BBSRC(BB/Y513763/1, BB/R019983/1, BB/S020969/), EPSRC(EP/X013707/1). This research is also partly supported by the Deutsche Forschungsgemeinschaft (DFG, German Research Foundation) under the DETECT - Collaborative Research Center (SFB 1502/1-2022 - Projektnummer: 450058266) as well as under the Germany’s Excellence Strategy – EXC 2070 – 390732324. We are also grateful for funding by the German Federal Ministry for Education and Science under the project number 01LL2204C (COINS).

%Bibliography
\bibliographystyle{unsrt}  
\bibliography{references}

\begin{thebibliography}{100}

\bibitem{gerland2023s}
Patrick Gerland.
\newblock What's beneath the future: World population prospects.
\newblock In {\em Semaine Data-SHS}, 2023.

\bibitem{laniak2013integrated}
Gerard~F Laniak, Gabriel Olchin, Jonathan Goodall, Alexey Voinov, Mary Hill, Pierre Glynn, Gene Whelan, Gary Geller, Nigel Quinn, Michiel Blind, et~al.
\newblock Integrated environmental modeling: a vision and roadmap for the future.
\newblock {\em Environmental modelling \& software}, 39:3--23, 2013.

\bibitem{change2001climate}
Intergovernmental Panel On~Climate Change.
\newblock Climate change 2007: Impacts, adaptation and vulnerability.
\newblock {\em Genebra, Su{\'\i}{\c{c}}a}, 2001.

\bibitem{shaikh2022towards}
Tawseef~Ayoub Shaikh, Tabasum Rasool, and Faisal~Rasheed Lone.
\newblock Towards leveraging the role of machine learning and artificial intelligence in precision agriculture and smart farming.
\newblock {\em Computers and Electronics in Agriculture}, 198:107119, 2022.

\bibitem{nakayama2022impact}
Tadanobu Nakayama.
\newblock Impact of anthropogenic disturbances on carbon cycle changes in terrestrial-aquatic-estuarine continuum by using an advanced process-based model.
\newblock {\em Hydrological Processes}, 36(2):e14471, 2022.

\bibitem{macpherson2020linking}
Joseph MacPherson, Carsten Paul, and Katharina Helming.
\newblock Linking ecosystem services and the sdgs to farm-level assessment tools and models.
\newblock {\em Sustainability}, 12(16):6617, 2020.

\bibitem{couedel2024long}
Antoine Cou{\"e}del, Gatien~N Falconnier, Myriam Adam, R{\'e}mi Cardinael, Kenneth Boote, Eric Justes, Ward~N Smith, Anthony~M Whitbread, Fran{\c{c}}ois Affholder, Juraj Balkovic, et~al.
\newblock Long-term soil organic carbon and crop yield feedbacks differ between 16 soil-crop models in sub-saharan africa.
\newblock {\em European Journal of Agronomy}, 155:127109, 2024.

\bibitem{galmarini2024assessing}
S~Galmarini, E~Solazzo, R~Ferrise, A~Kumar Srivastava, M~Ahmed, S~Asseng, AJ~Cannon, F~Dentener, G~De~Sanctis, T~Gaiser, et~al.
\newblock Assessing the impact on crop modelling of multi-and uni-variate climate model bias adjustments.
\newblock {\em Agricultural Systems}, 215:103846, 2024.

\bibitem{srivastava2023dynamic}
Amit~Kumar Srivastava, Jaber Rahimi, Karam Alsafadi, Murilo Vianna, Andreas Enders, Wenzhi Zheng, Alparslan Demircan, Mame Diarra~Bousso Dieng, Seyni Salack, Babacar Faye, et~al.
\newblock Dynamic modelling of mixed crop-livestock systems: A case study of climate change impacts in sub-saharan africa.
\newblock 2023.

\bibitem{sillero2021want}
Neftal{\'\i} Sillero, Salvador Arenas-Castro, Urtzi Enriquez-Urzelai, C{\^a}ndida~Gomes Vale, Diana Sousa-Guedes, Fernando Mart{\'\i}nez-Freir{\'\i}a, Raimundo Real, and A~M{\'a}rcia Barbosa.
\newblock Want to model a species niche? a step-by-step guideline on correlative ecological niche modelling.
\newblock {\em Ecological Modelling}, 456:109671, 2021.

\bibitem{zeng2022optical}
Yelu Zeng, Dalei Hao, Alfredo Huete, Benjamin Dechant, Joe Berry, Jing~M Chen, Joanna Joiner, Christian Frankenberg, Ben Bond-Lamberty, Youngryel Ryu, et~al.
\newblock Optical vegetation indices for monitoring terrestrial ecosystems globally.
\newblock {\em Nature Reviews Earth \& Environment}, 3(7):477--493, 2022.

\bibitem{geary2020guide}
William~L Geary, Michael Bode, Tim~S Doherty, Elizabeth~A Fulton, Dale~G Nimmo, Ayesha~IT Tulloch, Vivitskaia~JD Tulloch, and Euan~G Ritchie.
\newblock A guide to ecosystem models and their environmental applications.
\newblock {\em Nature Ecology \& Evolution}, 4(11):1459--1471, 2020.

\bibitem{he2023predicting}
Nianpeng He, Pu~Yan, Congcong Liu, Li~Xu, Mingxu Li, Koenraad Van~Meerbeek, Guangsheng Zhou, Guoyi Zhou, Shirong Liu, Xuhui Zhou, et~al.
\newblock Predicting ecosystem productivity based on plant community traits.
\newblock {\em Trends in Plant Science}, 28(1):43--53, 2023.

\bibitem{masson2021climate}
Val{\'e}rie Masson-Delmotte, Panmao Zhai, Anna Pirani, Sarah~L Connors, Clotilde P{\'e}an, Sophie Berger, Nada Caud, Y~Chen, L~Goldfarb, MI~Gomis, et~al.
\newblock Climate change 2021: the physical science basis.
\newblock {\em Contribution of working group I to the sixth assessment report of the intergovernmental panel on climate change}, 2(1):2391, 2021.

\bibitem{battiston2021physics}
Federico Battiston, Enrico Amico, Alain Barrat, Ginestra Bianconi, Guilherme Ferraz~de Arruda, Benedetta Franceschiello, Iacopo Iacopini, Sonia K{\'e}fi, Vito Latora, Yamir Moreno, et~al.
\newblock The physics of higher-order interactions in complex systems.
\newblock {\em Nature Physics}, 17(10):1093--1098, 2021.

\bibitem{razavi2021deep}
Saman Razavi.
\newblock Deep learning, explained: Fundamentals, explainability, and bridgeability to process-based modelling.
\newblock {\em Environmental Modelling \& Software}, 144:105159, 2021.

\bibitem{bhusal2022application}
Amrit Bhusal, Utsav Parajuli, Sushmita Regmi, and Ajay Kalra.
\newblock Application of machine learning and process-based models for rainfall-runoff simulation in dupage river basin, illinois.
\newblock {\em Hydrology}, 9(7):117, 2022.

\bibitem{jeong2020process}
Dongsu Jeong, Dohyun Kim, Taihun Choi, and Yoonho Seo.
\newblock A process-based modeling method for describing production processes of ship block assembly planning.
\newblock {\em Processes}, 8(7):880, 2020.

\bibitem{wagena2020comparison}
Moges~B Wagena, Dustin Goering, Amy~S Collick, Emily Bock, Daniel~R Fuka, Anthony Buda, and Zachary~M Easton.
\newblock Comparison of short-term streamflow forecasting using stochastic time series, neural networks, process-based, and bayesian models.
\newblock {\em Environmental Modelling \& Software}, 126:104669, 2020.

\bibitem{sinha2020estimation}
Sanjiv~K Sinha, Hitendra Padalia, Anindita Dasgupta, Jochem Verrelst, and Juan~Pablo Rivera.
\newblock Estimation of leaf area index using prosail based lut inversion, mlra-gpr and empirical models: Case study of tropical deciduous forest plantation, north india.
\newblock {\em International Journal of Applied Earth Observation and Geoinformation}, 86:102027, 2020.

\bibitem{chen2023evaluating}
Guangjie Chen, Tingfang Meng, Wenjie Wu, Bingcheng Si, Min Li, Boyang Liu, Shufang Wu, Hao Feng, and Kadambot~HM Siddique.
\newblock Evaluating potential groundwater recharge in the unsteady state for deep-rooted afforestation in deep loess deposits.
\newblock {\em Science of The Total Environment}, 858:159837, 2023.

\bibitem{kang2021semantic}
Jia Kang, Liantao Liu, Fucheng Zhang, Chen Shen, Nan Wang, and Limin Shao.
\newblock Semantic segmentation model of cotton roots in-situ image based on attention mechanism.
\newblock {\em Computers and electronics in agriculture}, 189:106370, 2021.

\bibitem{enders2023simplace}
Andreas Enders, Murilo Vianna, Thomas Gaiser, Gunther Krauss, Heidi Webber, Amit~Kumar Srivastava, Sabine~Julia Seidel, Andreas Tewes, Ehsan~Eyshi Rezaei, and Frank Ewert.
\newblock Simplace—a versatile modelling and simulation framework for sustainable crops and agroecosystems.
\newblock {\em in silico Plants}, 5(1):diad006, 2023.

\bibitem{van1989wofost}
CA~van Van~Diepen, J~van Wolf, H~Van~Keulen, and C~Rappoldt.
\newblock Wofost: a simulation model of crop production.
\newblock {\em Soil use and management}, 5(1):16--24, 1989.

\bibitem{mccown1995apsim}
RL~McCown, Graeme~L Hammer, John~NG Hargreaves, D~Holzworth, and Neil~I Huth.
\newblock Apsim: an agricultural production system simulation model for operational research.
\newblock {\em Mathematics and computers in simulation}, 39(3-4):225--231, 1995.

\bibitem{hoogenboom2004decision}
G~Hoogenboom, JW~Jones, PW~Wilkens, CH~Porter, WD~Batchelor, LA~Hunt, KJ~Boote, U~Singh, O~Uryasev, WT~Bowen, et~al.
\newblock Decision support system for agrotechnology transfer version 4.0.
\newblock {\em University of Hawaii, Honolulu, HI (CD-ROM)}, 2004.

\bibitem{fraga2015modeling}
Helder Fraga, Ricardo Costa, Jos{\'e} Moutinho-Pereira, Carlos~M Correia, Lia-T{\^a}nia Dinis, Igor Gon{\c{c}}alves, Jos{\'e} Silvestre, Jos{\'e} Eiras-Dias, Aureliano~C Malheiro, and Jo{\~a}o~A Santos.
\newblock Modeling phenology, water status, and yield components of three portuguese grapevines using the stics crop model.
\newblock {\em American Journal of Enology and Viticulture}, 66(4):482--491, 2015.

\bibitem{steduto2009aquacrop}
Pasquale Steduto, Theodore~C Hsiao, Dirk Raes, and Elias Fereres.
\newblock Aquacrop—the fao crop model to simulate yield response to water: I. concepts and underlying principles.
\newblock {\em Agronomy Journal}, 101(3):426--437, 2009.

\bibitem{webber2018physical}
Heidi Webber, Jeffrey~W White, Bruce~A Kimball, Frank Ewert, Senthold Asseng, Ehsan~Eyshi Rezaei, Paul~J Pinter~Jr, Jerry~L Hatfield, Matthew~P Reynolds, Behnam Ababaei, et~al.
\newblock Physical robustness of canopy temperature models for crop heat stress simulation across environments and production conditions.
\newblock {\em Field Crops Research}, 216:75--88, 2018.

\bibitem{chen2021no}
Qiuying Chen, Shengrui Wang, Zhaokui Ni, Ying Guo, Xiaofei Liu, Guoqiang Wang, and Hong Li.
\newblock No-linear dynamics of lake ecosystem in responding to changes of nutrient regimes and climate factors: Case study on dianchi and erhai lakes, china.
\newblock {\em Science of the Total Environment}, 781:146761, 2021.

\bibitem{kouadio2021performance}
Louis Kouadio, Philippe Tixier, Vivekananda Byrareddy, Torben Marcussen, Shahbaz Mushtaq, Bruno Rapidel, and Roger Stone.
\newblock Performance of a process-based model for predicting robusta coffee yield at the regional scale in vietnam.
\newblock {\em Ecological Modelling}, 443:109469, 2021.

\bibitem{de201925}
Allard De~Wit, Hendrik Boogaard, Davide Fumagalli, Sander Janssen, Rob Knapen, Daniel van Kraalingen, Iwan Supit, Raymond van~der Wijngaart, and Kees van Diepen.
\newblock 25 years of the wofost cropping systems model.
\newblock {\em Agricultural systems}, 168:154--167, 2019.

\bibitem{su2013lambert}
Hai-Xia Su, Zhao-Hui Zhang, Xiao-Yan Zhao, Zhi Li, Fang Yan, Han Zhang, et~al.
\newblock The lambert-beer’s law characterization of formal analysis in terahertz spectrum quantitative testing.
\newblock {\em Spectroscopy and Spectral Analysis}, 33(12):3180--3186, 2013.

\bibitem{ojeda2017evaluation}
Jonathan~J Ojeda, Jeffrey~J Volenec, Sylvie~M Brouder, Octavio~P Caviglia, and M{\'o}nica~G Agnusdei.
\newblock Evaluation of agricultural production systems simulator as yield predictor of panicum virgatum and miscanthus x giganteus in several us environments.
\newblock {\em Gcb Bioenergy}, 9(4):796--816, 2017.

\bibitem{holzworth2014apsim}
Dean~P Holzworth, Neil~I Huth, Peter~G deVoil, Eric~J Zurcher, Neville~I Herrmann, Greg McLean, Karine Chenu, Erik~J van Oosterom, Val Snow, Chris Murphy, et~al.
\newblock Apsim--evolution towards a new generation of agricultural systems simulation.
\newblock {\em Environmental Modelling \& Software}, 62:327--350, 2014.

\bibitem{dury2012models}
J{\'e}r{\^o}me Dury, No{\'e}mie Schaller, Fr{\'e}d{\'e}rick Garcia, Arnaud Reynaud, and Jacques~Eric Bergez.
\newblock Models to support cropping plan and crop rotation decisions. a review.
\newblock {\em Agronomy for sustainable development}, 32:567--580, 2012.

\bibitem{jones2003dssat}
James~W Jones, Gerrit Hoogenboom, Cheryl~H Porter, Ken~J Boote, William~D Batchelor, LA~Hunt, Paul~W Wilkens, Upendra Singh, Arjan~J Gijsman, and Joe~T Ritchie.
\newblock The dssat cropping system model.
\newblock {\em European journal of agronomy}, 18(3-4):235--265, 2003.

\bibitem{yan2020simulating}
Wenting Yan, Wenting Jiang, Xiaori Han, Wei Hua, Jinfeng Yang, and Peiyu Luo.
\newblock Simulating and predicting crop yield and soil fertility under climate change with fertilizer management in northeast china based on the decision support system for agrotechnology transfer model.
\newblock {\em Sustainability}, 12(6):2194, 2020.

\bibitem{pathak2017application}
Himanshu Pathak.
\newblock Application of decision support system for agrotechnology transfer (dssat--a crop simulation model) to assess the impacts of mean temperature rise on wheat yield.
\newblock {\em Climate Action: Mitigation and Adaptation in a Post Paris World}, page 107, 2017.

\bibitem{brisson2004crop}
Nadine Brisson.
\newblock Crop model stics (simulateur multidisciplinaire pour les cultures standard).
\newblock {\em Agronomie}, 24(6--7):293--293, 2004.

\bibitem{brisson2003overview}
Nadine Brisson, Christian Gary, Eric Justes, Romain Roche, Bruno Mary, Dominique Ripoche, Daniel Zimmer, Jorge Sierra, Patrick Bertuzzi, Philippe Burger, et~al.
\newblock An overview of the crop model stics.
\newblock {\em European Journal of agronomy}, 18(3-4):309--332, 2003.

\bibitem{queyrel2016pesticide}
Wilfried Queyrel, Florence Habets, H{\'e}l{\`e}ne Blanchoud, Dominique Ripoche, and Marie Launay.
\newblock Pesticide fate modeling in soils with the crop model stics: Feasibility for assessment of agricultural practices.
\newblock {\em Science of the Total Environment}, 542:787--802, 2016.

\bibitem{jego2015impact}
Guillaume J{\'e}go, Elizabeth Pattey, S~Morteza Mesbah, Jiangui Liu, and Isabelle Duchesne.
\newblock Impact of the spatial resolution of climatic data and soil physical properties on regional corn yield predictions using the stics crop model.
\newblock {\em International Journal of Applied Earth Observation and Geoinformation}, 41:11--22, 2015.

\bibitem{abedinpour2012performance}
M~Abedinpour, A~Sarangi, TBS Rajput, Man Singh, H~Pathak, and T~Ahmad.
\newblock Performance evaluation of aquacrop model for maize crop in a semi-arid environment.
\newblock {\em Agricultural Water Management}, 110:55--66, 2012.

\bibitem{steduto2009concepts}
Pasquale Steduto, Dirk Raes, Theodore~C Hsiao, Elias Fereres, Lee~K Heng, Terry~A Howell, Steven~R Evett, Basilio~A Rojas-Lara, Hamid~J Farahani, Gabriella Izzi, et~al.
\newblock Concepts and applications of aquacrop: The fao crop water productivity model.
\newblock In {\em Crop modeling and decision support}, pages 175--191. Springer, 2009.

\bibitem{yin2022observational}
Xiaomeng Yin and Guoyong Leng.
\newblock Observational constraint of process crop models suggests higher risks for global maize yield under climate change.
\newblock {\em Environmental Research Letters}, 17(7):074023, 2022.

\bibitem{roukh2020big}
Amine Roukh, Fabrice~Nolack Fote, Sidi~Ahmed Mahmoudi, and Sa{\"\i}d Mahmoudi.
\newblock Big data processing architecture for smart farming.
\newblock {\em Procedia Computer Science}, 177:78--85, 2020.

\bibitem{mathew2021deep}
Amitha Mathew, P~Amudha, and S~Sivakumari.
\newblock Deep learning techniques: an overview.
\newblock {\em Advanced Machine Learning Technologies and Applications: Proceedings of AMLTA 2020}, pages 599--608, 2021.

\bibitem{muruganantham2022systematic}
Priyanga Muruganantham, Santoso Wibowo, Srimannarayana Grandhi, Nahidul~Hoque Samrat, and Nahina Islam.
\newblock A systematic literature review on crop yield prediction with deep learning and remote sensing.
\newblock {\em Remote Sensing}, 14(9):1990, 2022.

\bibitem{li2022using}
Weide Li, Xi~Gao, Zihan Hao, and Rong Sun.
\newblock Using deep learning for precipitation forecasting based on spatio-temporal information: a case study.
\newblock {\em Climate Dynamics}, 58(1):443--457, 2022.

\bibitem{uddin2021review}
Md~Galal Uddin, Stephen Nash, and Agnieszka~I Olbert.
\newblock A review of water quality index models and their use for assessing surface water quality.
\newblock {\em Ecological Indicators}, 122:107218, 2021.

\bibitem{alabdrabalnabi2022machine}
Aessa Alabdrabalnabi, Ribhu Gautam, and S~Mani Sarathy.
\newblock Machine learning to predict biochar and bio-oil yields from co-pyrolysis of biomass and plastics.
\newblock {\em Fuel}, 328:125303, 2022.

\bibitem{lin2021hybrid}
Yongen Lin, Dagang Wang, Guiling Wang, Jianxiu Qiu, Kaihao Long, Yi~Du, Hehai Xie, Zhongwang Wei, Wei Shangguan, and Yongjiu Dai.
\newblock A hybrid deep learning algorithm and its application to streamflow prediction.
\newblock {\em Journal of Hydrology}, 601:126636, 2021.

\bibitem{yuan2020deep}
Qiangqiang Yuan, Huanfeng Shen, Tongwen Li, Zhiwei Li, Shuwen Li, Yun Jiang, Hongzhang Xu, Weiwei Tan, Qianqian Yang, Jiwen Wang, et~al.
\newblock Deep learning in environmental remote sensing: Achievements and challenges.
\newblock {\em Remote Sensing of Environment}, 241:111716, 2020.

\bibitem{khan2023comprehensive}
Junaid Khan, Eunkyu Lee, Awatef~Salem Balobaid, and Kyungsup Kim.
\newblock A comprehensive review of conventional, machine leaning, and deep learning models for groundwater level (gwl) forecasting.
\newblock {\em Applied Sciences}, 13(4):2743, 2023.

\bibitem{taud2018multilayer}
Hind Taud and Jean-Franccois Mas.
\newblock Multilayer perceptron (mlp).
\newblock {\em Geomatic approaches for modeling land change scenarios}, pages 451--455, 2018.

\bibitem{venkatraman2024channel}
Shravan Venkatraman et~al.
\newblock A channel attention-driven hybrid cnn framework for paddy leaf disease detection.
\newblock {\em arXiv preprint arXiv:2407.11753}, 2024.

\bibitem{park2023development}
Soo-Hwan Park, Bo-Young Lee, Min-Jee Kim, Wangyu Sang, Myung~Chul Seo, Jae-Kyeong Baek, Jae~E Yang, and Changyeun Mo.
\newblock Development of a soil moisture prediction model based on recurrent neural network long short-term memory (rnn-lstm) in soybean cultivation.
\newblock {\em Sensors}, 23(4):1976, 2023.

\bibitem{ur2024generative}
Zahid ur~Rahman, Mohd Shahrimie~Mohd Asaari, Haidi Ibrahim, Intan Sorfina~Zainal Abidin, and Mohamad~Khairi Ishak.
\newblock Generative adversarial networks (gans) for image augmentation in farming: A review.
\newblock {\em IEEE Access}, 2024.

\bibitem{shi2021biologically}
Yue Shi, Liangxiu Han, Wenjiang Huang, Sheng Chang, Yingying Dong, Darren Dancey, and Lianghao Han.
\newblock A biologically interpretable two-stage deep neural network (bit-dnn) for vegetation recognition from hyperspectral imagery.
\newblock {\em IEEE Transactions on Geoscience and Remote Sensing}, 60:1--20, 2021.

\bibitem{haghighat2022physics}
Ehsan Haghighat, Danial Amini, and Ruben Juanes.
\newblock Physics-informed neural network simulation of multiphase poroelasticity using stress-split sequential training.
\newblock {\em Computer Methods in Applied Mechanics and Engineering}, 397:115141, 2022.

\bibitem{cai2021physics}
Shengze Cai, Zhicheng Wang, Sifan Wang, Paris Perdikaris, and George~Em Karniadakis.
\newblock Physics-informed neural networks for heat transfer problems.
\newblock {\em Journal of Heat Transfer}, 143(6):060801, 2021.

\bibitem{bazrafshan2022predicting}
Ommolbanin Bazrafshan, Mohammad Ehteram, Sarmad~Dashti Latif, Yuk~Feng Huang, Fang~Yenn Teo, Ali~Najah Ahmed, and Ahmed El-Shafie.
\newblock Predicting crop yields using a new robust bayesian averaging model based on multiple hybrid anfis and mlp models.
\newblock {\em Ain Shams Engineering Journal}, 13(5):101724, 2022.

\bibitem{moseley2023finite}
Ben Moseley, Andrew Markham, and Tarje Nissen-Meyer.
\newblock Finite basis physics-informed neural networks (fbpinns): a scalable domain decomposition approach for solving differential equations.
\newblock {\em Advances in Computational Mathematics}, 49(4):62, 2023.

\bibitem{lu2021learning}
Lu~Lu, Pengzhan Jin, Guofei Pang, Zhongqiang Zhang, and George~Em Karniadakis.
\newblock Learning nonlinear operators via deeponet based on the universal approximation theorem of operators.
\newblock {\em Nature machine intelligence}, 3(3):218--229, 2021.

\bibitem{padhi2024paddy}
Jagamahon Padhi, Laxminarayana Korada, Ashis Dash, Prabira~Kumar Sethy, Santi~Kumari Behera, and Aziz Nanthaamornphong.
\newblock Paddy leaf disease classification using efficientnet b4 with compound scaling and swish activation: A deep learning approach.
\newblock {\em IEEE Access}, 2024.

\bibitem{zhang2022robustness}
Zhaodi Zhang, Jing Liu, Guanjun Liu, Jiacun Wang, and John Zhang.
\newblock Robustness verification of swish neural networks embedded in autonomous driving systems.
\newblock {\em IEEE Transactions on Computational Social Systems}, 2022.

\bibitem{shahhosseini2021corn}
Mohsen Shahhosseini, Guiping Hu, Saeed Khaki, and Sotirios~V Archontoulis.
\newblock Corn yield prediction with ensemble cnn-dnn.
\newblock {\em Frontiers in plant science}, 12:709008, 2021.

\bibitem{latif2022deep}
Ghazanfar Latif, Sherif~E Abdelhamid, Roxane~Elias Mallouhy, Jaafar Alghazo, and Zafar~Abbas Kazimi.
\newblock Deep learning utilization in agriculture: Detection of rice plant diseases using an improved cnn model.
\newblock {\em Plants}, 11(17):2230, 2022.

\bibitem{abdullahi2017convolution}
Halimatu~Sadiyah Abdullahi, R~Sheriff, and Fatima Mahieddine.
\newblock Convolution neural network in precision agriculture for plant image recognition and classification.
\newblock In {\em 2017 Seventh International Conference on Innovative Computing Technology (INTECH)}, volume~10, pages 256--272. Ieee New York, 2017.

\bibitem{zhao2023physics}
Xiaoyu Zhao, Zhiqiang Gong, Yunyang Zhang, Wen Yao, and Xiaoqian Chen.
\newblock Physics-informed convolutional neural networks for temperature field prediction of heat source layout without labeled data.
\newblock {\em Engineering Applications of Artificial Intelligence}, 117:105516, 2023.

\bibitem{faroughi2024physics}
Salah~A Faroughi, Nikhil~M Pawar, C{\'e}lio Fernandes, Maziar Raissi, Subasish Das, Nima~K Kalantari, and Seyed Kourosh~Mahjour.
\newblock Physics-guided, physics-informed, and physics-encoded neural networks and operators in scientific computing: Fluid and solid mechanics.
\newblock {\em Journal of Computing and Information Science in Engineering}, 24(4), 2024.

\bibitem{ndikumana2018deep}
Emile Ndikumana, Dinh Ho~Tong~Minh, Nicolas Baghdadi, Dominique Courault, and Laure Hossard.
\newblock Deep recurrent neural network for agricultural classification using multitemporal sar sentinel-1 for camargue, france.
\newblock {\em Remote Sensing}, 10(8):1217, 2018.

\bibitem{ren2021research}
Birong Ren, Xiangyu Xu, and Hongshen Yu.
\newblock Research of lstm-rnn model and its application evaluation on agricultural products circulation.
\newblock In {\em 2021 IEEE 3rd Eurasia Conference on IOT, Communication and Engineering (ECICE)}, pages 467--471. IEEE, 2021.

\bibitem{bhimavarapu2023improved}
Usharani Bhimavarapu, Gopi Battineni, and Nalini Chintalapudi.
\newblock Improved optimization algorithm in lstm to predict crop yield.
\newblock {\em Computers}, 12(1):10, 2023.

\bibitem{wang2022winter}
Jian Wang, Haiping Si, Zhao Gao, and Lei Shi.
\newblock Winter wheat yield prediction using an lstm model from modis lai products.
\newblock {\em Agriculture}, 12(10):1707, 2022.

\bibitem{shi2023physics}
Yue Shi, Shuhao Ma, Yihui Zhao, Chaoyang Shi, and Zhiqiang Zhang.
\newblock A physics-informed low-shot adversarial learning for semg-based estimation of muscle force and joint kinematics.
\newblock {\em IEEE Journal of Biomedical and Health Informatics}, 2023.

\bibitem{shi2022latent}
Yue Shi, Liangxiu Han, Lianghao Han, Sheng Chang, Tongle Hu, and Darren Dancey.
\newblock A latent encoder coupled generative adversarial network (le-gan) for efficient hyperspectral image super-resolution.
\newblock {\em IEEE Transactions on Geoscience and Remote Sensing}, 60:1--19, 2022.

\bibitem{prasad2022two}
Aaditya Prasad, Nikhil Mehta, Matthew Horak, and Wan~D Bae.
\newblock A two-step machine learning approach for crop disease detection using gan and uav technology.
\newblock {\em Remote Sensing}, 14(19):4765, 2022.

\bibitem{zhang2022improving}
Jingqi Zhang, Huiren Tian, Pengxin Wang, Kevin Tansey, Shuyu Zhang, and Hongmei Li.
\newblock Improving wheat yield estimates using data augmentation models and remotely sensed biophysical indices within deep neural networks in the guanzhong plain, pr china.
\newblock {\em Computers and Electronics in Agriculture}, 192:106616, 2022.

\bibitem{bartz2014evolutionary}
Thomas Bartz-Beielstein, J{\"u}rgen Branke, J{\"o}rn Mehnen, and Olaf Mersmann.
\newblock Evolutionary algorithms.
\newblock {\em Wiley Interdisciplinary Reviews: Data Mining and Knowledge Discovery}, 4(3):178--195, 2014.

\bibitem{holland1992genetic}
John~H Holland.
\newblock Genetic algorithms.
\newblock {\em Scientific american}, 267(1):66--73, 1992.

\bibitem{wang2018particle}
Dongshu Wang, Dapei Tan, and Lei Liu.
\newblock Particle swarm optimization algorithm: an overview.
\newblock {\em Soft computing}, 22:387--408, 2018.

\bibitem{sahoo2019long}
Bibhuti~Bhusan Sahoo, Ramakar Jha, Anshuman Singh, and Deepak Kumar.
\newblock Long short-term memory (lstm) recurrent neural network for low-flow hydrological time series forecasting.
\newblock {\em Acta Geophysica}, 67(5):1471--1481, 2019.

\bibitem{liu2024hybrid}
Ce~Liu, Shengli Du, Aimin Wei, Zhihui Cheng, Huanwen Meng, and Yike Han.
\newblock Hybrid prediction in horticulture crop breeding: Progress and challenges.
\newblock {\em Plants}, 13(19):2790, 2024.

\bibitem{abdel2024proposed}
Mahmoud Abdel-salam, Neeraj Kumar, and Shubham Mahajan.
\newblock A proposed framework for crop yield prediction using hybrid feature selection approach and optimized machine learning.
\newblock {\em Neural Computing and Applications}, pages 1--28, 2024.

\bibitem{brown2022designing}
Molly~E Brown, Stephen Mugo, Sebastian Petersen, and Dominik Klauser.
\newblock Designing a pest and disease outbreak warning system for farmers, agronomists and agricultural input distributors in east africa.
\newblock {\em Insects}, 13(3):232, 2022.

\bibitem{zhou2023application}
Feng Zhou, Yangbo Chen, and Jun Liu.
\newblock Application of a new hybrid deep learning model that considers temporal and feature dependencies in rainfall--runoff simulation.
\newblock {\em Remote Sensing}, 15(5):1395, 2023.

\bibitem{wang2021modelling}
Ying-Ping Wang and Daniel~S Goll.
\newblock Modelling of land nutrient cycles: recent progress and future development.
\newblock {\em Faculty reviews}, 10, 2021.

\bibitem{baydin2018automatic}
Atilim~Gunes Baydin, Barak~A Pearlmutter, Alexey~Andreyevich Radul, and Jeffrey~Mark Siskind.
\newblock Automatic differentiation in machine learning: a survey.
\newblock {\em Journal of machine learning research}, 18(153):1--43, 2018.

\bibitem{feng2019incorporating}
Puyu Feng, Bin Wang, De~Li~Liu, Cathy Waters, and Qiang Yu.
\newblock Incorporating machine learning with biophysical model can improve the evaluation of climate extremes impacts on wheat yield in south-eastern australia.
\newblock {\em Agricultural and Forest Meteorology}, 275:100--113, 2019.

\bibitem{choudhary2023non}
Manoj Choudhary, Sruthi Sentil, Jeffrey~B Jones, and Mathews~L Paret.
\newblock Non-coding deep learning models for tomato biotic and abiotic stress classification using microscopic images.
\newblock {\em Frontiers in Plant Science}, 14:1292643, 2023.

\bibitem{perdomo2022literature}
Maria~Elena Perdomo, Manuel Cardona, Denisse~Melisa Castro, and Wendy~Melany Mejia.
\newblock Literature review on artificial intelligence implementation in the honduran agricultural sector.
\newblock In {\em 2022 IEEE International Conference on Machine Learning and Applied Network Technologies (ICMLANT)}, pages 1--5. IEEE, 2022.

\bibitem{terasaki2023priority}
Drew~E Terasaki~Hart, Samantha Yeo, Maya Almaraz, Damien Beillouin, R{\'e}mi Cardinael, Edenise Garcia, Sonja Kay, Sarah~Taylor Lovell, Todd~S Rosenstock, Starry Sprenkle-Hyppolite, et~al.
\newblock Priority science can accelerate agroforestry as a natural climate solution.
\newblock {\em Nature Climate Change}, 13(11):1179--1190, 2023.

\bibitem{baker2023artemis}
James Baker.
\newblock Artemis: Using gans with multiple discriminators to generate art.
\newblock {\em arXiv preprint arXiv:2311.08278}, 2023.

\bibitem{seibert2021retrospective}
Jan Seibert and Sten Bergstr{\"o}m.
\newblock A retrospective on hydrological modelling based on half a century with the hbv model.
\newblock {\em Hydrology and Earth System Sciences Discussions}, 2021:1--28, 2021.

\bibitem{singh2021soil}
Nitin~K Singh, Ryan~E Emanuel, Brian~L McGlynn, and Chelcy~F Miniat.
\newblock Soil moisture responses to rainfall: Implications for runoff generation.
\newblock {\em Water Resources Research}, 57(9):e2020WR028827, 2021.

\bibitem{li2023enhancing}
Bu~Li, Ting Sun, Fuqiang Tian, and Guangheng Ni.
\newblock Enhancing process-based hydrological models with embedded neural networks: A hybrid approach.
\newblock {\em Journal of Hydrology}, 625:130107, 2023.

\bibitem{grigorian2024hybrid}
Gevik Grigorian, Sandip~V George, Sam Lishak, Rebecca~J Shipley, and Simon Arridge.
\newblock A hybrid neural ordinary differential equation model of the cardiovascular system.
\newblock {\em Journal of the Royal Society Interface}, 21(212):20230710, 2024.

\bibitem{huang2022novel}
Shuzhe Huang, Xiang Zhang, Nengcheng Chen, Hongliang Ma, Peng Fu, Jianzhi Dong, Xihui Gu, Won-Ho Nam, Lei Xu, Gerhard Rab, et~al.
\newblock A novel fusion method for generating surface soil moisture data with high accuracy, high spatial resolution, and high spatio-temporal continuity.
\newblock {\em Water Resources Research}, 58(5):e2021WR030827, 2022.

\bibitem{reichstein2019deep}
Markus Reichstein, Gustau Camps-Valls, Bjorn Stevens, Martin Jung, Joachim Denzler, Nuno Carvalhais, and F~Prabhat.
\newblock Deep learning and process understanding for data-driven earth system science.
\newblock {\em Nature}, 566(7743):195--204, 2019.

\bibitem{schuur2015climate}
Edward~AG Schuur, A~David McGuire, Christina Sch{\"a}del, Guido Grosse, Jennifer~W Harden, Daniel~J Hayes, Gustaf Hugelius, Charles~D Koven, Peter Kuhry, David~M Lawrence, et~al.
\newblock Climate change and the permafrost carbon feedback.
\newblock {\em Nature}, 520(7546):171--179, 2015.

\bibitem{van2024forward}
Birthe van~den Berg, Tom Schrijvers, James McKinna, and Alexander Vandenbroucke.
\newblock Forward-or reverse-mode automatic differentiation: What's the difference?
\newblock {\em Science of Computer Programming}, 231:103010, 2024.

\bibitem{ma2023multisource}
Yuchi Ma, Zhengwei Yang, and Zhou Zhang.
\newblock Multisource maximum predictor discrepancy for unsupervised domain adaptation on corn yield prediction.
\newblock {\em IEEE Transactions on Geoscience and Remote Sensing}, 61:1--15, 2023.

\bibitem{raissi2017physics}
Maziar Raissi, Paris Perdikaris, and George~Em Karniadakis.
\newblock Physics informed deep learning (part i): Data-driven solutions of nonlinear partial differential equations.
\newblock {\em arXiv preprint arXiv:1711.10561}, 2017.

\bibitem{shahhosseini2021coupling}
Mohsen Shahhosseini, Guiping Hu, Isaiah Huber, and Sotirios~V Archontoulis.
\newblock Coupling machine learning and crop modeling improves crop yield prediction in the us corn belt.
\newblock {\em Scientific reports}, 11(1):1606, 2021.

\bibitem{kheir2023integrating}
Ahmed~MS Kheir, Siyabusa Mkuhlani, Jane~W Mugo, Abdelrazek Elnashar, Vinay Nangia, Medha Devare, and Ajit Govind.
\newblock Integrating apsim model with machine learning to predict wheat yield spatial distribution.
\newblock {\em Agronomy Journal}, 115(6):3188--3196, 2023.

\bibitem{kotwal2024applying}
Sanaya Kotwal.
\newblock Applying remote sensing, google earth engine, and machine learning to predict the carbon sequestration potential of sundarbans mangroves.
\newblock {\em Asian Journal of Environment \& Ecology}, 23(8):140--150, 2024.

\bibitem{airaldi2023learning}
Filippo Airaldi, Bart De~Schutter, and Azita Dabiri.
\newblock Learning safety in model-based reinforcement learning using mpc and gaussian processes.
\newblock {\em IFAC-PapersOnLine}, 56(2):5759--5764, 2023.

\bibitem{wu2022optimizing}
Jing Wu, Ran Tao, Pan Zhao, Nicolas~F Martin, and Naira Hovakimyan.
\newblock Optimizing nitrogen management with deep reinforcement learning and crop simulations.
\newblock In {\em Proceedings of the IEEE/CVF conference on computer vision and pattern recognition}, pages 1712--1720, 2022.

\bibitem{malezieux2009mixing}
Eric Mal{\'e}zieux, Yves Crozat, Christian Dupraz, Marilyne Laurans, David Makowski, Harry Ozier-Lafontaine, Bruno Rapidel, St{\'e}phane De~Tourdonnet, and Muriel Valantin-Morison.
\newblock Mixing plant species in cropping systems: concepts, tools and models: a review.
\newblock {\em Sustainable agriculture}, pages 329--353, 2009.

\bibitem{czembor2022simulating}
Elzbieta Czembor, Zygmunt Kaczmarek, Wies{\l}aw Pilarczyk, Dariusz Ma{\'n}kowski, and Jerzy~H Czembor.
\newblock Simulating spring barley yield under moderate input management system in poland.
\newblock {\em Agriculture}, 12(8):1091, 2022.

\bibitem{tsai2021calibration}
Wen-Ping Tsai, Dapeng Feng, Ming Pan, Hylke Beck, Kathryn Lawson, Yuan Yang, Jiangtao Liu, and Chaopeng Shen.
\newblock From calibration to parameter learning: Harnessing the scaling effects of big data in geoscientific modeling.
\newblock {\em Nature communications}, 12(1):5988, 2021.

\bibitem{xue2022ensemble}
Yu~Xue, Yiling Tong, and Ferrante Neri.
\newblock An ensemble of differential evolution and adam for training feed-forward neural networks.
\newblock {\em Information Sciences}, 608:453--471, 2022.

\bibitem{innes2019differentiable}
Mike Innes, Alan Edelman, Keno Fischer, Chris Rackauckas, Elliot Saba, Viral~B Shah, and Will Tebbutt.
\newblock A differentiable programming system to bridge machine learning and scientific computing.
\newblock {\em arXiv preprint arXiv:1907.07587}, 2019.

\bibitem{tartakovsky2020physics}
Alexandre~M Tartakovsky, C~Ortiz Marrero, Paris Perdikaris, Guzel~D Tartakovsky, and David Barajas-Solano.
\newblock Physics-informed deep neural networks for learning parameters and constitutive relationships in subsurface flow problems.
\newblock {\em Water Resources Research}, 56(5):e2019WR026731, 2020.

\bibitem{ibrahim2022expert}
Eric~A Ibrahim, Daisy Salifu, Samuel Mwalili, Thomas Dubois, Richard Collins, and Henri~EZ Tonnang.
\newblock An expert system for insect pest population dynamics prediction.
\newblock {\em Computers and Electronics in Agriculture}, 198:107124, 2022.

\bibitem{singh2024novel}
Aditi Singh, Aryan Awasthi, Uday Badola, Ranjeet Bidwe, and Sashikala Mishra.
\newblock A novel hybrid approach to crop yield prediction: Combining deep learning efficiency with statistical precision.
\newblock {\em International Journal of Computing and Digital Systems}, 16(1):1--16, 2024.

\bibitem{ma2013assimilation}
Guannan Ma, Jianxi Huang, Wenbin Wu, Jinlong Fan, Jinqiu Zou, and Sijie Wu.
\newblock Assimilation of modis-lai into the wofost model for forecasting regional winter wheat yield.
\newblock {\em Mathematical and Computer Modelling}, 58(3-4):634--643, 2013.

\bibitem{nguyen2019mathematical}
Van~Cuong Nguyen, Seungtaek Jeong, Jonghan Ko, Chi~Tim Ng, and Jongmin Yeom.
\newblock Mathematical integration of remotely-sensed information into a crop modelling process for mapping crop productivity.
\newblock {\em Remote Sensing}, 11(18):2131, 2019.

\bibitem{shawon2020assessment}
Ashifur~Rahman Shawon, Jonghan Ko, Bokeun Ha, Seungtaek Jeong, Dong~Kwan Kim, and Han-Yong Kim.
\newblock Assessment of a proximal sensing-integrated crop model for simulation of soybean growth and yield.
\newblock {\em Remote Sensing}, 12(3):410, 2020.

\bibitem{shawon2020two}
Ashifur~Rahman Shawon, Jonghan Ko, Seungtaek Jeong, Taehwan Shin, Kyung~Do Lee, and Sang~In Shim.
\newblock Two-dimensional simulation of barley growth and yield using a model integrated with remote-controlled aerial imagery.
\newblock {\em Remote Sensing}, 12(22):3766, 2020.

\bibitem{shin2021simulation}
Taehwan Shin, Jonghan Ko, Seungtaek Jeong, Ashifur~Rahman Shawon, Kyung~Do Lee, and Sang~In Shim.
\newblock Simulation of wheat productivity using a model integrated with proximal and remotely controlled aerial sensing information.
\newblock {\em Frontiers in Plant Science}, 12:649660, 2021.

\bibitem{koch2007guidelines}
Elisabeth Koch, Ekko Bruns, Frank-M Chmielewski, Claudio Defila, Wolfgang Lipa, and Annette Menzel.
\newblock Guidelines for plant phenological observations.
\newblock {\em World Climate Data and Monitoring Programme}, 2007.

\bibitem{zhao2019phenological}
Guochun Zhao, Yuhan Gao, Shilun Gao, Yuanyuan Xu, Jiming Liu, Caowen Sun, Yuan Gao, Shiqi Liu, Zhong Chen, and Liming Jia.
\newblock The phenological growth stages of sapindus mukorossi according to bbch scale.
\newblock {\em Forests}, 10(6):462, 2019.

\bibitem{das2017automated}
Sruti Das~Choudhury, Saptarsi Goswami, Srinidhi Bashyam, Ashok Samal, and Tala Awada.
\newblock Automated stem angle determination for temporal plant phenotyping analysis.
\newblock In {\em Proceedings of the IEEE International Conference on Computer Vision Workshops}, pages 2022--2029, 2017.

\bibitem{campillo2010study}
Carlos Campillo, MI~Garcia, C~Daza, and MH~Prieto.
\newblock Study of a non-destructive method for estimating the leaf area index in vegetable crops using digital images.
\newblock {\em HortScience}, 45(10):1459--1463, 2010.

\bibitem{chacon2013quantitative}
Borja Chac{\'o}n, Roberto Ballester, Virginia Birlanga, Anne-Ga{\"e}lle Rolland-Lagan, and Jos{\'e}~Manuel P{\'e}rez-P{\'e}rez.
\newblock A quantitative framework for flower phenotyping in cultivated carnation (dianthus caryophyllus l.).
\newblock {\em PloS one}, 8(12):e82165, 2013.

\bibitem{cortes2017model}
Diego Fernando~Marmolejo Cortes, Renato~Santa Catarina, Gislanne Brito de~Ara{\'u}jo Barros, Fernanda Abreu~Santana Ar{\^e}des, Silvaldo Felipe~da Silveira, Geraldo~Ant{\^o}nio Ferreguetti, Helaine Christine~Cancela Ramos, Alexandre~Pio Viana, and Messias~Gonzaga Pereira.
\newblock Model-assisted phenotyping by digital images in papaya breeding program.
\newblock {\em Scientia Agricola}, 74(4):294--302, 2017.

\bibitem{bi2023transformer}
Luning Bi, Owen Wally, Guiping Hu, Albert~U Tenuta, and Daren~S Mueller.
\newblock A transformer-based approach for early prediction of soybean yield using time-series images.
\newblock {\em Frontiers in Plant Science}, 14:1173036, 2023.

\bibitem{wang2021transfer}
Yuping Wang and Weidong Li.
\newblock Transfer-based deep neural network for fault diagnosis of new energy vehicles.
\newblock {\em Frontiers in Energy Research}, 9:796528, 2021.

\bibitem{ma2021adaptive}
Yuchi Ma, Zhou Zhang, Hsiuhan~Lexie Yang, and Zhengwei Yang.
\newblock An adaptive adversarial domain adaptation approach for corn yield prediction.
\newblock {\em Computers and Electronics in Agriculture}, 187:106314, 2021.

\bibitem{wang2023bp}
Shanshan Wang, Lei Zhang, Pichao Wang, MengZhu Wang, and Xingyi Zhang.
\newblock Bp-triplet net for unsupervised domain adaptation: A bayesian perspective.
\newblock {\em Pattern Recognition}, 133:108993, 2023.

\bibitem{khaki2021simultaneous}
Saeed Khaki, Hieu Pham, and Lizhi Wang.
\newblock Simultaneous corn and soybean yield prediction from remote sensing data using deep transfer learning.
\newblock {\em Scientific Reports}, 11(1):11132, 2021.

\bibitem{yildirim2022using}
Tugba Yildirim, Daniel~N Moriasi, Patrick~J Starks, and Debaditya Chakraborty.
\newblock Using artificial neural network (ann) for short-range prediction of cotton yield in data-scarce regions.
\newblock {\em Agronomy}, 12(4):828, 2022.

\bibitem{wang2020detecting}
Ran Wang, John~A Gamon, Craig~A Emmerton, Kyle~R Springer, Rong Yu, and Gabriel Hmimina.
\newblock Detecting intra-and inter-annual variability in gross primary productivity of a north american grassland using modis maiac data.
\newblock {\em Agricultural and Forest Meteorology}, 281:107859, 2020.

\bibitem{alohali2023anomaly}
Manal~Abdullah Alohali, Mohammed Aljebreen, Nadhem Nemri, Randa Allafi, Mesfer~Al Duhayyim, Mohamed Ibrahim~Alsaid, Amani~A Alneil, and Azza~Elneil Osman.
\newblock Anomaly detection in pedestrian walkways for intelligent transportation system using federated learning and harris hawks optimizer on remote sensing images.
\newblock {\em Remote Sensing}, 15(12):3092, 2023.

\bibitem{vega2024convolutional}
Pedro Juan~Soto Vega, Panagiotis Papadakis, Marjolaine Matabos, Lo{\"\i}c Van~Audenhaege, Annah Ramiere, Joz{\'e}e Sarrazin, and Gilson Alexandre Ostwald~Pedro da~Costa.
\newblock Convolutional neural networks for hydrothermal vents substratum classification: An introspective study.
\newblock {\em Ecological Informatics}, 80:102535, 2024.

\bibitem{chaudhuri2023detection}
Gargi Chaudhuri and Niti~B Mishra.
\newblock Detection of aquatic invasive plants in wetlands of the upper mississippi river from uav imagery using transfer learning.
\newblock {\em Remote Sensing}, 15(3):734, 2023.

\bibitem{tufail2024classification}
Hina Tufail, Abdul Ahad, Mustahsan~Hammad Naqvi, Rahman Maqsood, and Ivan~Miguel Pires.
\newblock Classification of vascular dementia on magnetic resonance imaging using deep learning architectures.
\newblock {\em Intelligent Systems with Applications}, 22:200388, 2024.

\bibitem{zhao2024satellite}
Yaqi Zhao, Xianqiang He, Shuping Pan, Yan Bai, Difeng Wang, Teng Li, Fang Gong, and Xuan Zhang.
\newblock Satellite retrievals of water quality for diverse inland waters from sentinel-2 images: An example from zhejiang province, china.
\newblock {\em International Journal of Applied Earth Observation and Geoinformation}, 132:104048, 2024.

\bibitem{weiss2020remote}
Marie Weiss, Fr{\'e}d{\'e}ric Jacob, and Grgory Duveiller.
\newblock Remote sensing for agricultural applications: A meta-review.
\newblock {\em Remote sensing of environment}, 236:111402, 2020.

\bibitem{okyere2024hyperspectral}
Frank~Gyan Okyere, Daniel~Kingsley Cudjoe, Nicolas Virlet, March Castle, Andrew~Bernard Riche, Latifa Greche, Fady Mohareb, Daniel Simms, Manal Mhada, and Malcolm~John Hawkesford.
\newblock Hyperspectral imaging for phenotyping plant drought stress and nitrogen interactions using multivariate modeling and machine learning techniques in wheat.
\newblock {\em Remote Sensing}, 16(18):3446, 2024.

\bibitem{wolanin2019estimating}
Aleksandra Wolanin, Gustau Camps-Valls, Luis G{\'o}mez-Chova, Gonzalo Mateo-Garc{\'\i}a, Christiaan van~der Tol, Yongguang Zhang, and Luis Guanter.
\newblock Estimating crop primary productivity with sentinel-2 and landsat 8 using machine learning methods trained with radiative transfer simulations.
\newblock {\em Remote sensing of environment}, 225:441--457, 2019.

\bibitem{verhoef2018hyperspectral}
Wouter Verhoef, Christiaan Van Der~Tol, and Elizabeth~M Middleton.
\newblock Hyperspectral radiative transfer modeling to explore the combined retrieval of biophysical parameters and canopy fluorescence from flex--sentinel-3 tandem mission multi-sensor data.
\newblock {\em Remote sensing of environment}, 204:942--963, 2018.

\bibitem{sishodia2020applications}
Rajendra~P Sishodia, Ram~L Ray, and Sudhir~K Singh.
\newblock Applications of remote sensing in precision agriculture: A review.
\newblock {\em Remote sensing}, 12(19):3136, 2020.

\bibitem{zarco2018previsual}
Pablo~J Zarco-Tejada, C~Camino, PSA Beck, R~Calderon, Alberto Hornero, Roc{\'\i}o Hern{\'a}ndez-Clemente, T~Kattenborn, M~Montes-Borrego, L~Susca, M~Morelli, et~al.
\newblock Previsual symptoms of xylella fastidiosa infection revealed in spectral plant-trait alterations.
\newblock {\em Nature Plants}, 4(7):432--439, 2018.

\bibitem{hernandez2019early}
Roc{\'\i}o Hern{\'a}ndez-Clemente, Alberto Hornero, Matti Mottus, Josep Pe{\~n}uelas, Victoria Gonz{\'a}lez-Dugo, Juan~Carlos Jim{\'e}nez, Lola Su{\'a}rez, Luis Alonso, and Pablo~J Zarco-Tejada.
\newblock Early diagnosis of vegetation health from high-resolution hyperspectral and thermal imagery: Lessons learned from empirical relationships and radiative transfer modelling.
\newblock {\em Current forestry reports}, 5:169--183, 2019.

\end{thebibliography}

\end{document}